\documentclass{article}

\usepackage{amsmath,amsfonts,bm}

\def\eqref#1{equation~\ref{#1}}

\def\1{\bm{1}}

\DeclareMathAlphabet{\mathsfit}{\encodingdefault}{\sfdefault}{m}{sl}
\SetMathAlphabet{\mathsfit}{bold}{\encodingdefault}{\sfdefault}{bx}{n}

\usepackage{microtype}
\usepackage{graphicx}
\usepackage{subfigure}
\usepackage{booktabs} 

\usepackage{hyperref}

\usepackage{amsfonts}       
\usepackage{nicefrac}       
\usepackage{xcolor}         
\usepackage{multirow}
\usepackage{bbm}
\usepackage{natbib}
\usepackage{tikz}
\usepackage{amssymb}
\usetikzlibrary{arrows}
\usepackage{adjustbox}
\usepackage{latexsym}
\usepackage{epsfig}
\usepackage{fancybox}
\usepackage{pstricks}
\usepackage{caption}
\usepackage{fancyhdr}
\usepackage{color}
\usepackage{wrapfig,lipsum}
\usepackage{cancel}
\usepackage{kotex}
\usepackage{framed}

\def\cL{{\cal L}}
\def\cN{{\cal N}}

\def\cU{{\cal U}}

\def\qed{\space$\square$ \par \vspace{.15in}}

\newcommand{\bz}{{\bf z}}

\newcommand{\bx}{{\bf x}}

\newcommand{\bw}{{\bf w}}

\newcommand{\bu}{{\bf u}}

\newcommand{\bc}{\begin{center}}
\newcommand{\ec}{\end{center}}
\newcommand{\be}{\begin{equation}}
\newcommand{\ee}{\end{equation}}
\newcommand{\ba}{\begin{array}}
\newcommand{\ea}{\end{array}}
\newcommand{\bean}{\begin{eqnarray*}}
\newcommand{\eean}{\end{eqnarray*}}
\newcommand{\bea}{\begin{eqnarray}}
\newcommand{\eea}{\end{eqnarray}}
\newcommand{\ben}{\begin{enumerate}}
\newcommand{\een}{\end{enumerate}}
\newcommand{\bed}{\begin{itemize}}
\newcommand{\eed}{\end{itemize}}
\definecolor{mmd}{HTML}{D2302C}
\definecolor{laftr}{HTML}{A04EF6}
\definecolor{sipm}{HTML}{F95700}
\definecolor{ftm}{HTML}{343148}

\usepackage[accepted]{icml2024}

\usepackage{amsmath}
\usepackage{amssymb}
\usepackage{mathtools}
\usepackage{amsthm}

\usepackage[capitalize,noabbrev]{cleveref}

\definecolor{purple}{RGB}{112,48,160}
\definecolor{green}{RGB}{56,87,35}
\definecolor{red}{RGB}{255,0,0}

\theoremstyle{plain}
\newtheorem{theorem}{Theorem}[section]
\newtheorem{proposition}[theorem]{Proposition}

\theoremstyle{definition}

\theoremstyle{remark}
\newtheorem{remark}[theorem]{Remark}

\usepackage[textsize=tiny]{todonotes}

\icmltitlerunning{ODIM: Outlier Detection via Likelihood of Under-Fitted Generative Models}

\begin{document}

\twocolumn[
\icmltitle{ODIM: Outlier Detection via Likelihood of Under-Fitted Generative Models}



\icmlsetsymbol{equal}{*}

\begin{icmlauthorlist}
\icmlauthor{Dongha Kim*}{sungshin}
\icmlauthor{Jaesung Hwang*}{skt}
\icmlauthor{Jongjin Lee}{sr}
\icmlauthor{Kunwoong Kim}{snu}
\icmlauthor{Yongdai Kim}{snu}
\end{icmlauthorlist}

\icmlaffiliation{sungshin}{Department of Statistics and Data Science Center, Sungshin Women's University, Seoul, Republic of Korea}
\icmlaffiliation{skt}{SK Telecom, Seoul, Republic of Korea}
\icmlaffiliation{sr}{Samsung Research, Seoul, Republic of Korea}
\icmlaffiliation{snu}{Department of Statistics, Seoul National University, Seoul, Republic of Korea}

\icmlcorrespondingauthor{Yongdai Kim}{ydkim0903@gmail.com}

\icmlkeywords{Machine Learning, ICML}

\vskip 0.3in
]



\printAffiliationsAndNotice{\icmlEqualContribution} 

\begin{abstract}
The unsupervised outlier detection (UOD) problem refers to a task to identify inliers given training data which contain outliers as well as inliers, without any labeled information about inliers and outliers. 
It has been widely recognized that using fully-trained likelihood-based deep generative models (DGMs) often results in poor performance in distinguishing inliers from outliers. 
In this study, we claim that \textit{the likelihood itself could serve as powerful evidence for identifying inliers in UOD tasks, provided that DGMs are carefully under-fitted.}
Our approach begins with a novel observation called the \textit{inlier-memorization (IM) effect}--when training a deep generative model with data including outliers, the model initially memorizes inliers before outliers. 
Based on this finding, we develop a new method called the \textit{outlier detection via the IM effect (ODIM)}. 
Remarkably, the ODIM requires only a few updates, making it computationally efficient--\textit{at least tens of times faster} than other deep-learning-based algorithms. 
Also, the ODIM filters out outliers excellently, regardless of the data type, including tabular, image, and text data. 
To validate the superiority and efficiency of our method, we provide extensive empirical analyses on close to 60 datasets.
\end{abstract}

\section{Introduction}
\label{sec:1}
\paragraph{Outlier detection}
Outlier (also anomaly) is an observation that differs significantly from other observations, and outlier detection (OD) is the task of identifying outliers in a given dataset.
OD has wide applications such as fraud detection, fault detection, and defect detection in images. 
OD is also used as a pre-processing step in supervised learning to filter out anomalous training samples, which may degrade the performance of a predictive model. 

OD problems can be categorized into three areas in general:
1) Supervised outlier detection (SOD) requires label information about whether each training sample is inlier (also normal) or outlier and solves the two-class classification task. 
2) Semi-supervised outlier detection (SSOD) refers to methods that assume 
all training data being inliers and construct patterns or models based only on the inliers.
3) Unsupervised outlier detection (UOD) deals with the most realistic situations where training data include some outliers but no label information about anomalousness is available. 
Most anomaly detection tasks in practice are involved in UOD since the information of outliers in massive data is hardly known in advance.

\begin{figure}[t]
\begin{center}
\centerline{
\includegraphics[width=0.49\textwidth]{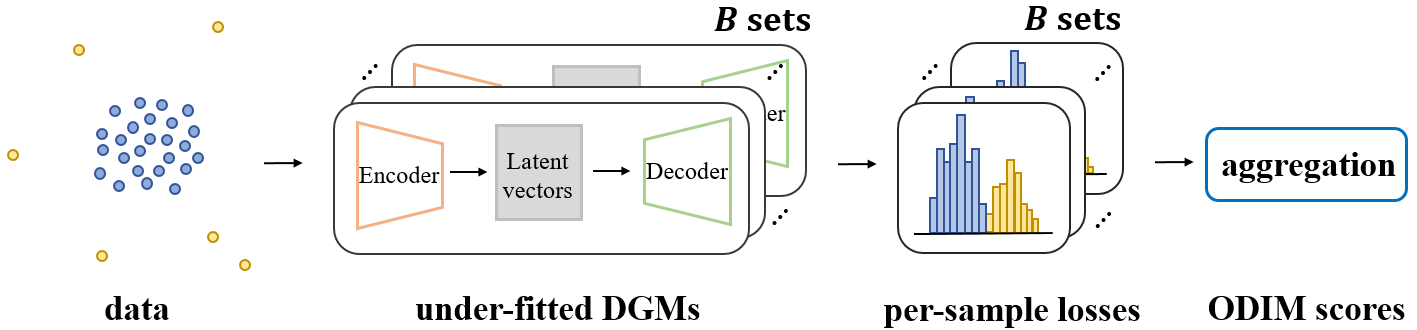}
}
\caption{An illustration of the ODIM method. 
}
\label{fig:odim_method}
\end{center}
\end{figure}

\paragraph{Likelihood-based approaches in OD}
To detect outliers from data, a fundamental approach might involve using a deep generative model (DGM) and regarding each sample as either an inlier or not based on its likelihood value. 
However, it is widely recognized that \textit{the likelihood value
itself using fully trained DGMs is by no means a reasonable indicator to identify outliers}, particularly in out-of-distribution (OOD) tasks (or SSOD tasks) where none of the outliers are present in training data. 
The likelihood values of outliers with a fully trained DGM are often higher than those of inliers \citep{DBLP:conf/iclr/NalisnickMTGL19,DBLP:journals/corr/abs-1906-02994,DBLP:journals/entropy/LanD21}. 


\paragraph{Overview of our method}
In this study, we claim that \textit{likelihood could serve as an effective score in UOD tasks, provided we employ carefully trained under-fitting DGMs.}

Our algorithm is motivated by the so called {\it memorization effect} that is observed in noisy label problems \citep{arpit2017closer,jiang2018mentornet}. The goal of noisy label problems is to learn an accurate classifier when some of the class labels in the training data are contaminated.

When standard supervised learning algorithms are applied to such mislabeled data, an interesting phenomenon known as the memorization effect is observed. 
In this effect, correctly labeled data are learned earlier than mislabeled ones in the training phase of deep neural networks. 
The memorization effect makes it possible to detect mislabeled data by comparing per-sample losses in the early training phase. 

Building upon the effect, our primary goal is to apply this concept to the field of UOD. 
We start with finding a new and interesting observation that \textit{the memorization effect is also observed in learning DGMs}.
That is, when we train a deep generative model with training data that include outliers, the inliers' loss values reduce prior to those of outliers at early updates. 
We call this observation the \textit{inlier-memorization (IM) effect}. 
The IM effect occurs because, in the early training phase, decreasing the loss values of inliers rather than outliers is a more beneficial direction to reduce the overall loss, which will be discussed in Section \ref{sec:3}.

Based on the IM effect, we propose a simple yet powerful UOD solver called the \textit{outlier detection via the IM effect} (ODIM). 
We train a DGM with a log-likelihood-based approach such as the VAE \citep{kingma2013auto} or IWAE \citep{DBLP:journals/corr/BurdaGS15} for a few updates, and we regard data with large loss values compared to the per-sample loss distribution as outliers. 
Figure \ref{fig:odim_method} provides a visual representation of our method. 

A previous study has explored the use of the memorization effect for UOD problems \citep{NEURIPS2019_6c4bb406}. In their work, the authors found that the memorization effect can also be observed in the self-supervised learning framework when using artificially and carefully designed pseudo-labels. 
They successfully applied this approach to accurately detect outliers in image domains. However, it is challenging to extend their method to other data domains such as tabular data and sequential data due to the unavailability of a suitable pseudo-labeling strategy for these domains.

On the other hand, the ODIM is \textit{domain-agnostic} and thus can be applied to various data domains, including tabular, image, and sequential data, as it does not rely on specific pseudo-labeling strategies. 
By analyzing nearly 60 datasets from various domains, we demonstrate that the ODIM consistently yields state-of-the-art or competitive results in identifying outliers across a wide range of data types.

Additionally, the ODIM offers significant \textit{computational efficiency}. 
In fact, the ODIM requires only a few training updates, often less than a single epoch, in the training phase to detect outliers. 
Thus, the ODIM is at least tens of times faster than other recent deep-learning-based UOD solvers, such as \citet{ruff2018deep}, requiring a larger number of training updates, for example, 200 epochs.


The remainder of this paper is organized as follows. 
Section \ref{sec:2} provides a brief review of related research on OD problems. 
Detailed descriptions of the ODIM algorithm with discussions of the IM effect are given in Section \ref{sec:3}. 
Results of various experiments including performance tests and ablation studies follow in Section \ref{sec:4}. 
Finally, further discussions and concluding remarks are respectively presented in Section \ref{sec:5} and \ref{sec:6}.
The key contributions of this work are:

\bed
\item We find a new phenomenon called the IM effect that DGMs memorize inliers prior to outliers at early training updates. 
\item We develop a simple and powerful likelihood-based UOD solver called the ODIM based on the IM effect and other improving techniques. 
\item We empirically demonstrate the superiority and efficiency of our method by analyzing extensive benchmark datasets. 
\item We conduct additional experiments to cover a couple of extensions where our method can be applied.
\eed

\section{Related works}
\label{sec:2}

We review algorithms for both SSOD and UOD since the former algorithms are often used in UOD tasks as well.
\vspace{-.3cm}
\paragraph{Semi-supervised outlier detection} 
A popular technique for SSOD is the one class classification approach which
transforms data into a feature space and distinguishes outliers from inliers by their radii from the center on the feature space.
The OCSVM \citep{ocsvm} and SVDD \citep{tax2004support}
are two representative algorithms, which use kernel techniques to construct the feature space.

Succeeding their ideas, plenty of SSOD algorithms using deep neural networks have been developed.
The DeepSVDD \citep{deepsvdd} extends the SVDD by utilizing a deep autoencoder (AE) for learning a feature map, and the DeepSAD \cite{deepsad} modifies the DeepSVDD to incorporate labeled outliers to training data.
Modifications of the DeepSVDD have been developed by \citet{dagmm, mahmood2021multiscale, 7410534}. 
In addition to AE, deep generative models are also popularly used for SSOD \citep{ryu-etal-2018-domain,https://doi.org/10.48550/arxiv.1906.02994, jiang2022revisiting}.

There are methods for SSOD other than the one class classification approach.
The SimCLR \citep{simclr} and BERT \citep{devlin-etal-2019-bert} utilize  self-supervised learning to obtain a desirable feature map, 
and various algorithms based on this idea have been developed \citep{geom, goad, csi, ssd}.
When some labels (not related to inliers or outliers) are available, several studies have found that feature maps for classification of those labels can improve outlier detection
\citep{hendrycks17baseline, liang2018enhancing, gomes2022igeood}.

There have been attempts to use the likelihood to detect outliers.
As mentioned in Section \ref{sec:1}, the likelihood itself performs poorly in the SSOD, thus,
certain transformations or indirect uses of the likelihood have been studied \citep{DBLP:conf/iclr/NalisnickMTGL19,DBLP:journals/corr/abs-1906-02994,DBLP:journals/entropy/LanD21}.

\paragraph{Unsupervised outlier detection}
As for traditional approaches,
the LOF \citep{10.1145/335191.335388}  compares  the density of a given datum compared to the densities of its neighborhoods,  
and the IF \citep{4781136} utilizes the fact that outliers can be separated out by random
trees with relatively small sizes.
The UOCL \citep{6909884} solves UOD problems by employing pseudo soft labels and training them jointly with the one-class classification model.

There are various methods to solve UOD problems with deep learning models.
The RDA \citep{rda} combines the robust PCA and AE to detect outliers.
The DSEBM \citep{10.5555/3045390.3045507} utilizes the energy-based model for density estimation and uses the energy score or reconstruction error to identify outliers.
The RSRAE \citep{lai2020robust} devises a new hidden layer called RSR, inserting it between encoder and decoder of a deep AE to separate inliers and outliers effectively.
The $E^{3}$-Outlier \citep{NEURIPS2019_6c4bb406} trains a deep neural network
by self-supervised learning and identifies outliers based on how fast the loss decreases as the training proceeds.
Recently, diffusion models \citep{DBLP:conf/nips/HoJA20} have also been leveraged to detect outliers in both SSOD and UOD tasks \citep{DBLP:journals/corr/abs-2305-18593}.

\section{Proposed method}
\label{sec:3}
\subsection{Notations and definitions}
For a given input vector $\bx\in\mathbb{R}^D$, we denote its anomalousness by $y^o\in\{0,1\}$, that is, $y^o=0$ if $\bx$ is an inlier and $y^o=1$ otherwise. 
Note that only $\bx$ is observable but $y^o$ is not under the UOD regime. 
Let $\mathcal{U}^{tr}=\{\bx_1,\ldots,\bx_n\}$ be unlabeled training data. 
Our goal is to detect outlier samples, i.e. $\bx$ with $y^o=1$, from $\mathcal{U}^{tr}$ as accurately as possible. 

Let $p(\bx|\bz;\theta)$ and $q(\bz|\bx;\phi)$ be given decoder and encoder parameterized by $\theta$ and $\phi$, respectively, where $\bz\in\mathbb{R}^d$ (generally assuming $d < D$) is a latent vector. 
We construct the distribution of the input random vector $\boldsymbol{X}\in\mathbb{R}^D$ as follows:
\begin{align*}
\boldsymbol{X}\sim p(\bx|\boldsymbol{Z};\theta),
\end{align*}
where $\boldsymbol{Z}\sim\mathcal{N}({0}_d,{I}_d)$ denotes a latent random vector.

For a given $p\in\mathbb{N}$, we denote the $l_p$-norm of a vector $\mathbf{a}$ by $\|\mathbf{a}\|_p$. 
For two real-valued functions defined on $\mathbb{R}_{>0}$, $f(t)$ and $g(t)$, $f(t)$ is said to be $\Theta(g(t))$ if there exist positive constants $C_1,C_2$, and $T$ such that $C_1\cdot g(t)\le f(t) \le C_2\cdot g(t)$ holds for all $t\ge T$.

\begin{figure*}[t]
\begin{center}
\centerline{
\includegraphics[width=0.162\linewidth]{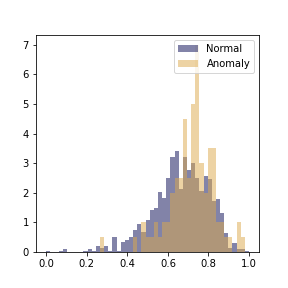}
\includegraphics[width=0.162\linewidth]{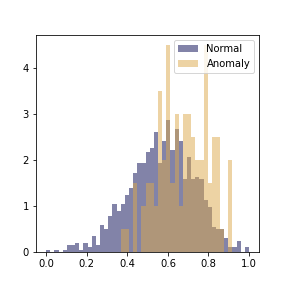}
\includegraphics[width=0.162\linewidth]{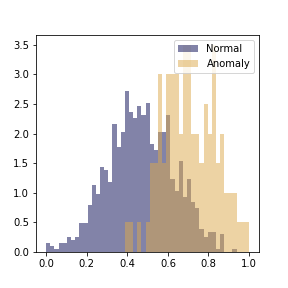}
\includegraphics[width=0.162\linewidth]{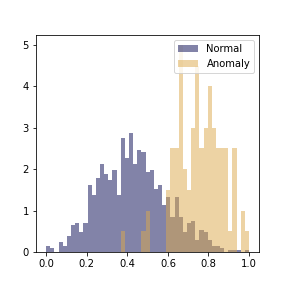}
\includegraphics[width=0.162\linewidth]{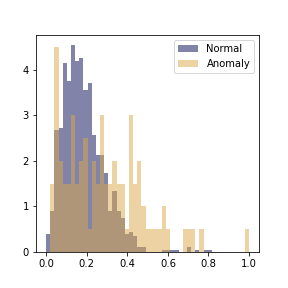}
\includegraphics[height=0.147\textwidth]{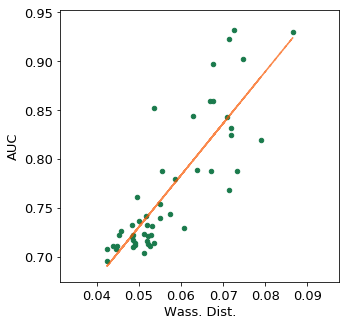}
}
\caption{
({\bf 1st to 5th}) The distributions of the per-sample (normalized) VAE loss values of {\texttt{Cardio}} after 10, 20, 30, 40, and 500 training updates, respectively.
For each panel, we depict the histograms of inliers and outliers separately. 
({\bf Last}) The positive relationship between the Wasserstein distance and identifying performance (AUC) on \texttt{Cardio}. 
}
\label{fig:inlier_memorization}
\end{center}
\end{figure*}

\subsection{Motivation: inlier-memorization effect}
\label{sec:3.1}

Suppose that we are training a likelihood-based DGM with a certain learning framework where the training data contain both inliers and outliers.
To illustrate the IM effect, we analyze \texttt{Cardio} dataset and train a DGM using the VAE method \citep{kingma2013auto}. 

To prepare the data, we normalize each variable of \texttt{Cardio} to a range between 0 and 1.
The encoder and decoder architectures are 2-layered deep neural networks (DNNs) with $d=5$ and 50 hidden nodes for each hidden layer.  
Our focus is to analyze the distribution of per-sample VAE loss as updates proceed. 
We train the decoder and encoder by minimizing the VAE loss function for up to 500 updates.


The panels in Figure \ref{fig:inlier_memorization}, excluding the last one, display the empirical distribution of per-sample loss values for the training data at different updates. 
We can observe that the discrepancy in loss distributions between inliers and outliers becomes clearer as updates progress in the early training phase. 
However, as the DGM is sufficiently trained, the two distributions become overlapped, making it almost impossible to distinguish them based solely on their loss values.
We call this phenomenon the \textit{inlier-memorization (IM) effect}. 


The IM effect is not surprising conceptually. 
When the per-sample loss function is continuous, reducing the loss in dense regions is beneficial for overall loss reduction (e.g. the negative log-likelihood). 
As inliers tend to be located in \textit{dense} regions and outliers in \textit{sparse} regions, reasonable learning algorithms prioritize dense regions in the early training phase, leading to the IM effect. 
It is important to note that the IM effect is observed only in the early training phase, as the learned model memorizes both inliers and outliers later.

\subsection{Theoretical analysis}
\label{sec:theorem}

We provide a theoretical explanation of the occurence of the IM effect
with a simple example where we train a linear factor model using the VAE.
That is, $p(\bx|\bz;\theta)$ is the density function of $W\bz+b+\epsilon$, where $W\in\mathbb{R}^{D\times d},b\in\mathbb{R}^{D}$ are the loading matrix and bias vector and $\epsilon\sim N(0_D,\sigma^2I_D)$ is a noise random vector. 
And we set $q(\bz|\bx;\phi)$ as the density function of $U\bx+v+\tau$, where $U\in\mathbb{R}^{d\times D}$, $v\in\mathbb{R}^{d}$, and $\tau\sim\cN(0_d, \eta^2I_d)$. 
For simplicity, we fix $\sigma$ and $\eta$ and only train $W,b,U,$ and $v$.
That is, the learnable parameters $\theta$ and $\phi$ become $(W,b)$ and $(U,v),$ respectively.
Note that the objective function of the VAE for a given input vector $\bx$ is given as
\begin{align*}
\label{vae}
L^{\text{VAE}}(\theta,\phi;\bx):=\underset{{\bz\sim q(\bz|\bx;\phi)}}{\mathbb{E}}\left[ \log \left( \frac{p(\bx|\bz;\theta)p(\bz)}{q(\bz|\bx;\phi)} \right) \right],    
\end{align*}
where $p(\bz)$ is the density function of $\mathcal{N}(0_d,1_d)$. 
We assume that each element in $W,b,U,$ and $v$ is randomly initialized by the i.i.d. uniform distribution on $[-1,1]$.
Then we have the following result whose proof is given in Appendix A. 

{\proposition\footnote{Since the generative model $p(\bx;\theta)$ is related only with the parameter $\theta$, we only consider the gradient with respect to $\theta$.}{
\label{prop:1}
For an input vector $\bx$, the following holds:
\begin{equation*}
    \begin{split}
        \mathbb{E}_{\theta,\phi}  \left\|\frac{\partial}{\partial \theta}L^{\text{VAE}}(\theta,\phi;\bx)\right\|_2^2=\Theta\left(\|\bx\|_1^4\right).
    \end{split}
\end{equation*}
}}

Proposition \ref{prop:1} indicates that in the early phases of learning, the magnitude of the gradient of the VAE is proportional to the $l_1$-norm of the input vector on average. This implies that when the norms of inliers and outliers are similar, the initial update direction of $\theta$ is influenced by the inliers, as they are much more prevalent than outliers in the training dataset.
Consequently, during the initial training phase, the generative model is trained towards memorizing the inliers before the outliers, resulting in the IM effect.

Of course, Proposition \ref{prop:1} may not hold \textit{after} initial updates. 
However, we empirically found that this tendency persists for a while, and the loss distributions between inliers and outliers becomes more distinguishable (See Figure \ref{fig:inlier_memorization}).

\begin{remark}
As mentioned in the overview of our method in Section \ref{sec:1}, the behavior of parameter update in early training steps is frequently utilized, particularly in addressing noisy label problems \citep{arpit2017closer,jiang2018mentornet}, yet it lacks rigorous theoretical validation. 
To the best of our knowledge, our study offers the first theoretical insights about parameter updates during the initial learning step, particularly in relation to the norms of data. 
Exploring this aspect further could be a promising research avenue.
\end{remark}

\begin{remark}
Besides the ELBO, such as VAE, there is another widely-used likelihood-based approach, normalizing flows \citep{kobyzev2020normalizing}. 
Among the normalizing flows methods, we consider GLOW \citep{DBLP:conf/nips/KingmaD18} to investigate whether the IM effect also occurs in other likelihood-based DGMs. 
We observe that the IM effect clearly appears during the GLOW training. 
This finding supports the claim that \textit{the IM effect is a universal phenomenon in likelihood-based models.} 
Detailed descriptions of this experiment are provided in Appendix B.
\end{remark}

\begin{figure}[t]
\begin{center}
\includegraphics[width=0.21\textwidth]{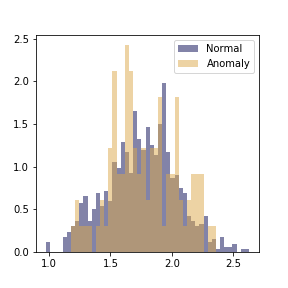}
\includegraphics[width=0.21\textwidth]{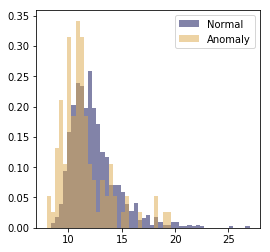}
\includegraphics[width=0.21\textwidth]{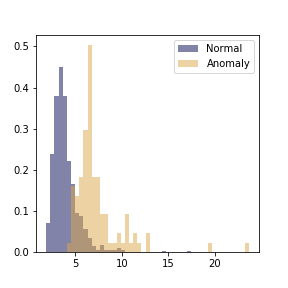}
\includegraphics[width=0.21\textwidth]{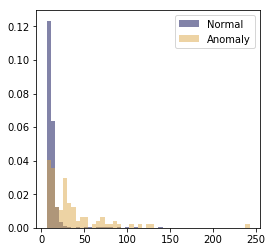}
\caption{Distributions of per-sample (\textbf{Left}) input $l_1$-norm values and (\textbf{Right})  gradient $l_2$-norm values of VAE loss on \texttt{Cardio}. 
We consider two pre-processing schemes to normalize each feature: 1) (\textbf{Upper}) min-max scaling, and 2) (\textbf{Lower}) standardization.
}
\label{fig:grad_norm_restuls}
\end{center}
\end{figure}

\subsection{Choice of pre-processing technique}

Proposition \ref{prop:1} suggests that ODIM performs well when the norms of inliers and outliers are similar. 
However, it should be noted that the norm of a datum depends on the choice of a pre-processing. 
Therefore, careful selection of pre-processing is essential for the success of ODIM. 
A pre-processing which makes the norms of inliers and outleris be similar would be a good choice.

As potential pre-processing candidates, we compare two widely used techniques, 1) min-max scaling and 2) standardization. 
The min-max scaling transforms each variable to have a distribution ranged between zero and one, and the standardization makes each variable to have zero mean and unit variance. 

We explore how these techniques affect the performance of the ODIM.
Intuitively, we can infer that, since the majority of data are inliers, the process of shifting the entire data towards the origin in the standardization technique would likely result in small norms for the inliers, potentially weakening the IM effect. 
This conjecture can be supported by the following simple proposition, whose proof is provided in Appendix A. 
\begin{proposition}
Let $\mathbf{X}^{in}$ and $\mathbf{X}^{out}$ be inlier and outlier random vectors with zero mean, i.e., $\mathbb{E}(X^{in})=\mathbb{E}(X^{out})=0$. 
Suppose that their respective supports are $Supp(X^{in})=A^{in}$ and $Supp(X^{out})=A^{out}$, where $A^{in}$ is a bounded convex set and $A^{out}$ is a set wrapping $A^{in}$, i.e., $A^{in}\cap A^{out}=\emptyset$ and $\text{conv}(A^{out})\supsetneq A^{in}$. 
Define $\mathbf{X}_{mm}^{in}$ and $\mathbf{X}_{mm}^{out}$ as pre-processed inlier and outlier random vectors using
the min-max scaling. 
Similarly, we define $\mathbf{X}_{st}^{in}$ and $\mathbf{X}_{st}^{out}$ obtained by the standardization.
Then, we have 
$\mathbb{E}\|\mathbf{X}_{mm}^{in}\|_1=\mathbb{E}\|\mathbf{X}_{mm}^{out}\|_1$, while $\mathbb{E}\|\mathbf{X}_{st}^{in}\|_1<\mathbb{E}\|\mathbf{X}_{st}^{out}\|_1$. 
\end{proposition}

We note that the condition that the supports of inliers and outliers do not overlap in the above proposition is not strictly necessary for the proof.  
But we maintain this, as the definition of an outlier is an observation \textit{significantly different from inliers}.


Figure \ref{fig:grad_norm_restuls} visually validates our theoretical result. 
We again note the implication of Proposition \ref{prop:1} that if the norm of a particular sample is large, its gradient would have a more significant impact on the parameter update.
Thus, when the standardization is applied, the input norms of inliers are larger than those of outliers, which leads to potentially biased parameter update towards memorizaing outliers, even if inliers are more prevalent than outliers.
This phenomenon hinders a DGM from training inliers at early learning updates, and thus, diminishes the strength of the IM effect. 

From these findings, we use the min-max scaling throughout our experiments, instead of the standardization. We empirically observe that the ODIM with the min-max scaled data has better results on most of the tabular datasets analyzed in the experiment.
More results about comparing the min-max and standardization are provided in the ablation studies and Appendix C.

While the min-max scaling is better than the standardization, we do not claim that it is the optimal choice for the IM effect. We leave the optimal choice of pre-processing as a future research topic.

\subsection{Algorithm description}
\label{sec:step1}

The IM effect can be used for outlier detection by utilizing the per-sample loss value
of a deep generative model during  early training phases. 
In this section, we propose a new UOD solver called the \textit{outlier detection via IM effect (ODIM)}. 
The ODIM consists of three steps: 1) applying the min-max scaling to the data, 2) training a DGM for a specified number of updates, and 3) identifying a sample as an outlier if its corresponding loss value is relatively large. 
To implement this approach, additional considerations are required, which are explained below.


\paragraph{Choice of the learning algorithm for a DGM} 
Selecting a learning algorithm for a DGM carefully is crucial to make the IM effect appear more clearly. 
There exist numerous algorithms to train likelihood-based DGMs, which can be roughly divided into two approaches: 1) the ELBO approach, which calculates the lower bound of log-likelihood, \citep{kingma2013auto,DBLP:journals/corr/BurdaGS15,DBLP:conf/nips/HoJA20} and 2) normalizing flows approach, which calculates the exact log-likelihood \citep{DBLP:journals/corr/DinhKB14,DBLP:conf/iclr/DinhSB17,DBLP:conf/nips/KingmaD18}. 
In this study, we conclude to employ the importance weighted autoencoder (IWAE, \citet{DBLP:journals/corr/BurdaGS15}), one of the ELBO approach.  
Detailed discussions can be found in Appendix B.

The objective function of the IWAE is given as:
\begin{equation*}\label{eq:KKL}
    \begin{split}
        &{L}^{\text{IWAE}}(\theta,\phi;\bx) \\
        &:=  -\underset{\bz_1,...,\bz_K\sim q(\bz|\bx;\phi)}{\mathbb{E}}\left[ \log \left( \frac{1}{K} \sum_{k=1}^K \frac{p(\bx|\bz_k;\theta)p(\bz_k)}{q(\bz_k|\bx;\phi)} \right) \right],
    \end{split}    
\end{equation*}

where $p(\bz)$ is the density of the standard multivariate Gaussian distribution and 
$K$ is the number of samples. 
Note that the IWAE reduces to the VAE when $K=1$. 
We train the encoder and decoder simultaneously by minimizing the negative empirical expectation, given as 
\begin{equation}\label{eq:iwae_obj}
    \begin{split}
\mathbb{E}_{\bx\sim\cU^{tr}}{L}^{\text{IWAE}}(\theta,\phi;\bx)
    \end{split}    
\end{equation}
with respect to $\theta$ and $\phi$ using an SGD-based optimizer.

\paragraph{Selection of the optimal number of updates}
We empirically found out that the IM effect often emerges very early in the training phase, often even within a single epoch, and its magnitude (i.e., the difference of the loss distributions between inliers and outliers) is highly sensitive to the number of model updates. 
And as the model memorizes outliers as well as inliers gradually, the model becomes no longer capable of distinguishing inliers from outliers (See the 5th panel in Figure \ref{fig:inlier_memorization}). 
Thus, it would be a key for the success of the ODIM algorithm to choose the optimal number of updates data adaptively.

We devise a heuristic but powerful strategy to decide the number of updates where the IM effect is maximized.
At each update of the model, we evaluate the \textit{degree of bimodality} of 
the per-sample loss distribution and select the optimal number of updates where  
the degree of bimodality is maximized.

We observe that the two per-sample loss distributions, one for inliers and the other for outliers, seemingly follow  Gaussian distributions (See the first four panels in Figure \ref{fig:inlier_memorization}).
Therefore, to quantify the degree of bimodality, we fit a two-component Gaussian mixture model (GMM-2), denoted as ${\pi}_1 \mathcal{N}({\mu}_1,{\sigma}_1^2)+{\pi}_2\mathcal{N}({\mu}_2,{\sigma}_2^2)$, to the (normalized) per-sample loss values using the currently estimated generative model. 
We then measure the discrepancy between the two normal distributions in the fitted GMM-2 using the Wasserstein distance. 
This discrepancy measure serves as an indicator of the degree of bimodality.

The rightmost panel in Figure \ref{fig:inlier_memorization} illustrates the values of AUC on the training data of \texttt{Cardio} at the first $10\times m$ updates for $m=1,\ldots,50$ and their corresponding Wasserstein distances. We can clearly see that the Wasserstein distance is a useful measure for selecting the optimal number of updates. 

In practice, we calculate the Wasserstein distance at every $N_\text{u}$ update and stop the update process if the largest Wasserstein distance has not been improved for $N_\text{pat}$ consecutive times. 
And the optimal number of updates is determined as the one that maximizes the Wasserstein distance.
We set $N_{\text{u}}$ and $N_{\text{pat}}$ to 10 in all numerical experiments, unless otherwise specified.


\paragraph{Incorporating multiple ODIM scores}
Our method includes several random components, such as parameter initialization and mini-batch arrangement, resulting in stochastic outcomes.
To stabilize and enhance our method, we employ an ensemble strategy.
Multiple models with different initial values are trained in parallel to obtain the multiple best models by use of the ODIM.


Let $(\theta^{*(b)},\phi^{*(b)}),b=1,\ldots,B$ be $B$ pairs of estimated parameters, each of which is trained independently using the IWAE and our early stopping rule, where $B$ is the number of ensembled models.  
Then, for a given datum ${\bf x}$, the formulation of the ensembled ODIM score becomes $\sum_{b=1}^B {L}^{\text{IWAE}}(\theta^{*(b)},\phi^{*(b)};{\bf x})/B$. 
We consider the input $\bx$ as an inlier when its score is low and vise versa. 
In our experiments, the number of multiple models is fixed at 10, i.e., $B=10$, unless otherwise specified. 
It is worth noting that the ODIM algorithm runs quickly, so implementing an ensemble is still computationally efficient.
We provide the ODIM's pseudo algorithm in Algorithm \ref{alg:proposed}. 



\begin{algorithm}[t]
 \caption{ \textbf{ODIM}  
        \\
        In practice, we set $(K,N_\text{u},N_\text{pat})=(50,10,10)$.}
 \label{alg:proposed}
 \textbf{Input}: Training dataset $\mathcal{U}^{tr}=\{\bx_1,...,\bx_n\}$

    \begin{algorithmic}[1]
        \REQUIRE: Decoder and encoder: $p(\bx|\bz;\theta)$ and $q(\bz|\bx;\phi)$, GMM-2 model: $\pi_1\mathcal{N}(\mu_1,\sigma_1^2)+\pi_2\mathcal{N}(\mu_2,\sigma_2^2)$, Mini-batch size: $n_{\text{mb}}$, Optimizer: $\mathcal{O}$, 
        Number of samples in IWAE: $K$, Update unit number: $N_{\text{u}}$, Maximum patience: $N_\text{pat}$
        

    \FOR{b in $(1:B)$}
        \STATE Initialize $(\theta^{(b)},\phi^{(b)})$ and set $d_{\text{WD}}^{max}$ to 0.
        \WHILE{$n_{\text{pat}}<N_{\text{pat}}$}
            \FOR{k in $(1:N_\text{u})$}
            	 \STATE Drawn $n_\text{mb}$ samples, $\{\bx_i\}_{i=1}^{n_{\text{mb}}}$, from $\mathcal{U}^{tr}$.
                  \STATE Apply the min-max scaling to $\{\bx_i\}_{i=1}^{n_{\text{mb}}}$.
    	       \STATE Update $(\theta^{(b)},\phi^{(b)})$ using the IWAE with $\{\bx_i\}_{i=1}^{n_{\text{mb}}}$ and $\mathcal{O}$.
            	\STATE $\{\tilde{l}_i\}_{i=1}^{n_{\text{mb}}}$ $\leftarrow$ $\text{MinMax}(\{{L}^{\text{IWAE}}(\bx_i)\}_{i=1}^{n_{\text{mb}}})$. 
            	\STATE Fit the parameters in GMM-2 using $\{\tilde{l}_i\}_{i=1}^{n_{\text{mb}}}$ and calculate the WD distance $d_{\text{WD}}$.
                \IF{$d_{\text{WD}}>D_{\text{WD}}^{\text{max}}$}
                    \STATE $d_{\text{WD}}^{\text{max}}$ $\leftarrow$ $d_{\text{WD}}$ 
    	           \STATE $(\theta^{*(b)},\phi^{*(b)})$ $\leftarrow$ $(\theta^{(b)},\phi^{(b)})$ 
    	           \STATE $n_{\text{pat}}$ $\leftarrow$ $0$
                \ELSE
                    \STATE $n_{\text{pat}}$ $\leftarrow$ $n_{\text{pat}}+1$
                \ENDIF
            \ENDFOR
        \ENDWHILE
    \ENDFOR
        \\
    Calculate ODIM scores: 
    $$l_i^{*}\leftarrow \frac{1}{B}\sum_{b=1}^B {L}^{\text{IWAE}}(\theta^{*(b)},\phi^{*(b)};{\bf x}_i), i=1,\ldots,n$$ 
    \end{algorithmic}
\textbf{Output}: ODIM scores $\{l_i^{*}\}_{i=1}^n$
\end{algorithm}

\section{Numerical experiments}
\label{sec:4}

We demonstrate the superiority of our proposed method through extensive experiments. We analyze a wide range of datasets across tabular, image, and text types. 
Across all data types, we show that ODIM outperforms other competitors, including state-of-the-art methods, in terms of outlier detection performance and computational cost. Additionally, we discuss an extension of ODIM in two situations: 1) when a small amount of anomalous information is available and 2) when we consider a differential privacy regime.

For all the experiments, we report the averaged results based on five implementations with randomly initialized parameters. 
We utilize the \texttt{Pytorch} framework to run our algorithm using a single NVIDIA TITAN XP GPU. 
The implementation code for our method is publicly available at \url{https://github.com/jshwang0311/ODIM}.
  
\paragraph{Dataset description}
As aforementioned above, we analyze 57 benchmark datasets for OD covering tabular, images, and texts, all of which are sourced from \texttt{ADBench}\footnote{\url{https://github.com/Minqi824/ADBench}} \citep{DBLP:conf/nips/HanHHJ022}. 

We consider 36 tabular datasets that are frequently analyzed in the OD literature.
These datasets originate from diverse domains, such as healthcare, finance, and astronomy.

For images, we analyze six datasets from \texttt{ADBench}: \texttt{MNIST}, \texttt{MNIST-C}, \texttt{FMNIST}, \texttt{CIFAR10}, \texttt{SVHN}, and \texttt{MVTec-AD}. 
We utilize the feature vectors extracted by the ViT model \citep{DBLP:conf/iclr/DosovitskiyB0WZ21}, which are available in \texttt{ADBench}. 

We additionally include five more benchmark datasets commonly employed in natural language processing (NLP) domains: \texttt{Amazon}, \texttt{20news}, \texttt{Agnews}, \texttt{Imdb}, and \texttt{Yelp}. 
For these datasets, we employ our method using the embedding features provided by either BERT \citep{devlin-etal-2019-bert} or RoBERTa \citep{DBLP:journals/corr/abs-1907-11692}, also available in \texttt{ADBench}.


For each dataset, we perform the min-max scaling. 
We refer to Appendix C and \citet{han2022adbench} for the detailed descriptions of all the datasets.

\paragraph{Baseline}
For baselines to be compared with the ODIM, we refer to \citet{DBLP:journals/corr/abs-2305-18593} and the baselines they considered.
To be more detailed, we first consider all the UOD solvers, 16 in total, including deep-learning-based ones, described in \texttt{ADBench}: PCA \citep{shyu2003novel}, OCSVM, \citep{ocsvm}, LOF \citep{breunig2000LOF}, CBLOF \citep{he2003discovering}, COF \citep{tang2002enhancing}, HBOS \citep{goldstein2012histogram}, kNN \citep{ramaswamy2000efficient}, SOD \citep{kriegel2009outlier}, COPOD \citep{li2020copod}, ECOD \citep{li2022ecod}, IF \citep{liu2008isolation}, LODA \citep{pevny2016loda}, FeatureBagging \citep{lazarevic2005feature}, MCD \citep{fauconnier2009outliers}, DeepSVDD \citep{ruff2018deep}, and DAGMM \citep{DBLP:conf/iclr/ZongSMCLCC18}. 

Furthermore, we consider four recent UOD methods outside of \texttt{ADBench}: DROCC \citep{DBLP:conf/icml/GoyalRJS020}, GOAD \citep{goad}, ICL \citep{DBLP:conf/iclr/ShenkarW22}, and DTE \citep{DBLP:journals/corr/abs-2305-18593}.


We exclude likelihood-based methods specifically developed for SSOD, such as \citet{DBLP:conf/iclr/NalisnickMTGL19}, as we have observed their poor performance in UOD tasks (Please read Appendix D for empirical evidence.).


\paragraph{Architecture \& learning schedule}
We use two hidden layered DNN architectures for building the encoder and decoder and set $K$, the number of samples drawn from the encoder used for constructing the IWAE objective function, to 50. 
We minimize the IWAE objective function in (\ref{eq:iwae_obj}) with the Adam optimizer \citep{kingma2014adam} with a mini-batch size of 128 and a learning rate of 5e-4.  
To run the ODIM, we fix the two hyper-parameters, $N_{\text{u}}$ and $N_{\text{pat}}$, to 10. 
For ensemble learning, we train 10 pairs of encoder and decoder, each of which is trained independently. 

\subsection{Performance for outlier identification}

We begin by comparing ODIM with baseline methods to assess its performance in identifying outliers within a training dataset.  
To do this, we examine the area under receiver operating characteristic (AUC), a standard measure that most other studies have used. 
Additionally, to provide solid evidence of our method's superior performance, we evaluate the area under precision-recall (PR), which summarizes the precision-recall curve.

As we compare tens of baselines with ODIM, we provide selected results in our main manuscript, with detailed results of all methods for each dataset in Appendix C. 
We note that all the baseline results are referenced from Appendix in \citet{DBLP:journals/corr/abs-2305-18593}. 




\paragraph{Results for tabular data}
Table \ref{tab:tabular_avg_auc_ap} provides a summary of the averaged AUC and PR scores for 46 tabular datasets. 
The ODIM achieves the highest averaged scores in terms of both AUC and PR, indicating its superior performance in outlier detection for tabular data. 
These superior results imply that the ODIM can be readily used as a reliable method for outlier detection in tabular data.

\begin{table}[t]
\caption{Averaged AUC and PR scores over 46 tabular datasets.}
\centering
\resizebox{0.485\textwidth}{!}{
\begin{tabular}{l|cccccccc}
\toprule
{\Large \textbf{Method}} & {\Large\textbf{OCSVM}} & {\Large\textbf{COPOD}} & {\Large\textbf{ECOD}} & {\Large\textbf{DeepSVDD}} & {\Large\textbf{ICL}} & {\Large\textbf{DDPM}} & {\Large\textbf{DTE}} & {\Large\textbf{ODIM}}  \\ 
\midrule
{\Large \textbf{AUC}}    & {\Large 0.740}          & {\Large 0.730    }    & {\Large 0.729   }    & {\Large 0.543    }      & {\Large 0.652    }     & {\Large 0.712      }       & {\Large 0.730  }        & {\Large\textbf{0.757}}
\\
{\Large \textbf{PR}}     & {\Large 0.360      }    & {\Large 0.339  }      & {\Large 0.349   }    & {\Large 0.182  }        & {\Large 0.201    }     & {\Large 0.332     }        & {\Large 0.321  }        & {\Large\textbf{0.366}} \\ 
\bottomrule
\end{tabular}
}
\label{tab:tabular_avg_auc_ap}
\end{table}

\begin{table}[t]
\caption{Averaged AUC and PR scores over 6 image datasets.}
\centering
\resizebox{0.485\textwidth}{!}{
\begin{tabular}{l|cccccccc}
\toprule
{\Large \textbf{Method}} & {\Large\textbf{OCSVM}} & {\Large\textbf{COPOD}} & {\Large\textbf{ECOD}} & {\Large\textbf{DeepSVDD}} & {\Large\textbf{ICL}} & {\Large\textbf{DDPM}} & {\Large\textbf{DTE}} & {\Large\textbf{ODIM}}  \\ 
\midrule
{\Large \textbf{AUC}}    & {\Large 0.744}          & {\Large 0.508    }    & {\Large 0.511   }    & {\Large 0.580    }      & {\Large 0.655    }     & {\Large 0.738      }       & {\Large 0.757  }        & {\Large\textbf{0.813}}
\\
{\Large \textbf{PR}}     & {\Large 0.271      }    & {\Large 0.090  }      & {\Large 0.091   }    & {\Large 0.176  }        & {\Large 0.172    }     & {\Large 0.267     }        & {\Large 0.282  }        & {\Large\textbf{0.429}} \\ 
\bottomrule
\end{tabular}
}
\label{tab:image_avg_auc_ap}
\end{table}


\begin{table}[t]
\caption{Averaged AUC and PR scores over 5 text datasets.}
\centering
\resizebox{0.485\textwidth}{!}{
\begin{tabular}{l|cccccccc}
\toprule
{\Large \textbf{Method}} & {\Large\textbf{OCSVM}} & {\Large\textbf{COPOD}} & {\Large\textbf{ECOD}} & {\Large\textbf{DeepSVDD}} & {\Large\textbf{ICL}} & {\Large\textbf{DDPM}} & {\Large\textbf{DTE}} & {\Large\textbf{ODIM}}  \\ 
\midrule
{\Large \textbf{AUC}}    & {\Large 0.566}          & {\Large 0.554    }    & {\Large 0.537   }    & {\Large 0.504    }      & {\Large 0.546    }     & {\Large 0.548      }       & {\Large 0.598  }        & {\Large\textbf{0.659}}
\\
{\Large \textbf{PR}}     & {\Large 0.062   }    & {\Large 0.060  }      & {\Large 0.057   }    & {\Large 0.054  }        & {\Large 0.058    }     & {\Large 0.059     }        & {\Large 0.070  }        & {\Large\textbf{0.097}} \\ 
\bottomrule
\end{tabular}
}
\label{tab:text_avg_auc_ap}
\end{table}

\paragraph{Results for image data}
Table \ref{tab:image_avg_auc_ap} showcases the averaged AUC and PR scores for 6 image datasets. 
The results clearly show that ODIM provides outstanding results with large margins compared to other competitors in most cases. 


\paragraph{Results for text data}
The performance results for 5 text datasets are also summarized in Table \ref{tab:text_avg_auc_ap}. 
Once again, the ODIM stands out as the best-performing method across text datasets on average. 
It is worth highlighting that the ODIM has demonstrated superior and consistent performance across different data types throughout the aforementioned empirical experiments, making it an effective method for outlier detection tasks. 
Therefore, the ODIM can be used as an off-the-shelf tool for outlier detection. 

\begin{figure*}[t]
\begin{center}
\includegraphics[width=0.245\textwidth]{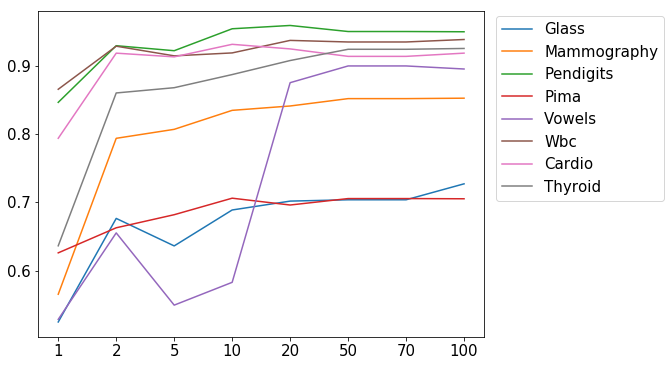}
\includegraphics[width=0.245\textwidth]{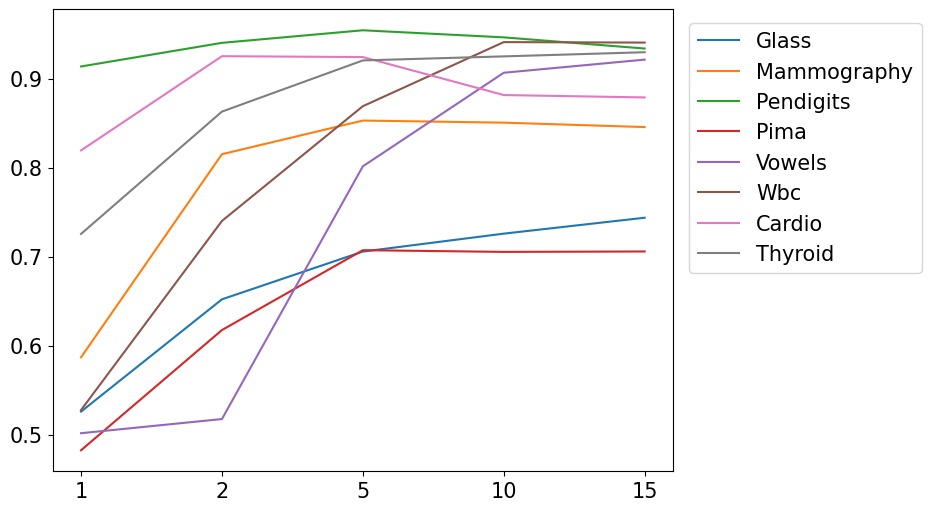}
\includegraphics[width=0.245\textwidth]{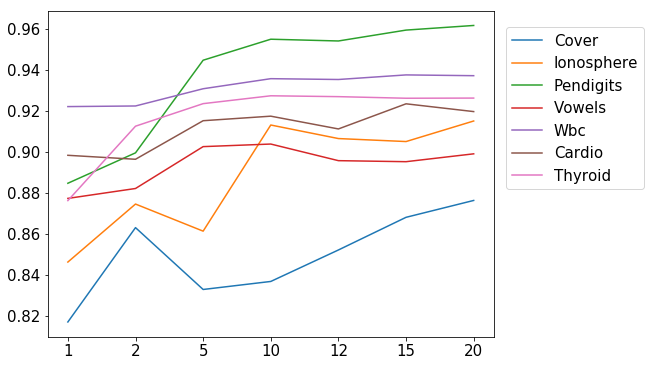}
\includegraphics[width=0.215\textwidth]{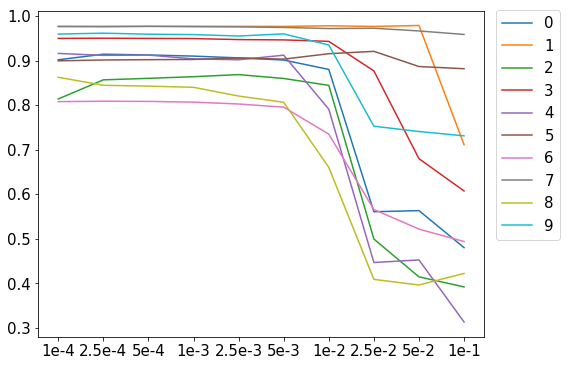}
\caption{({\bf{From left to right}}) {\bf {1)}} AUC results on tabular datasets with various values of $K$.
{\bf {2)}} AUC results on tabular datasets with various values of $N_{\text{pat}}$.
{\bf {3)}} AUC results on tabular datasets with various values of $B$. 
{\bf {4)}} AUC results on \texttt{FMNIST} for each class with various learning rates. We vary the learning rate from 1e-4 to 1e-1. 
}
\label{fig:ablation_nens_lr_auc_result}
\end{center}
\end{figure*}

\paragraph{Implementation time comparison}

Recall our claim mentioned in Section \ref{sec:1} that our method is computationally efficient compared to other deep-learning-based baselines and even other conventional competitors. 
To validate this claim, we conduct a comparative analysis of the running times on multiple datasets, as summarized in Table \ref{tab:running_time}. 

As expected, the two deep learning methods are considerably slower compared to other methods.
On the other hand, the ODIM demonstrates remarkable efficiency. 
This significant improvement in computational efficiency makes ODIM a highly practical and scalable method for OD tasks.

\begin{table}[t]
\caption{Running time comparison for the ODIM and other competitors. All records are measured in seconds.}
\centering
\resizebox{0.45\textwidth}{!}{
\begin{tabular}{l|rrrrr}
\toprule
{\Large\textbf{Method}} & \multicolumn{1}{c}{\Large\textbf{OCSVM}} & \multicolumn{1}{c}{\Large\textbf{LoF}} & \multicolumn{1}{c}{\Large\textbf{IF}} & \multicolumn{1}{c}{\Large\textbf{DeepSVDD}} & \multicolumn{1}{c}{\Large\textbf{ODIM}} \\ 
\midrule
{\Large\texttt{Cover}}         & {\Large  2164.270    }                             & {\Large  24.959    }                       & {\Large  6.192  }                         & {\Large  3463.821    }                                                  & {\Large  7.735 }                            \\
{\Large\texttt{Mammography}}   & {\Large  2.482  }                                  & {\Large  0.221    }                        & {\Large  0.408  }                         & {\Large  135.958  }                                                    & {\Large  4.057 }                           \\
{\Large\texttt{Pendigits}}     & {\Large  1.247   }                                 & {\Large  1.899      }                      & {\Large  0.350  }                         & {\Large  83.944  }                                                   & {\Large  5.595  }                           \\
{\Large\texttt{Satellite}}     & {\Large  1.152   }                                 & {\Large  1.735      }                      & {\Large  0.369  }                         & {\Large  78.876   }                                                   & {\Large  4.949   }                          \\
{\Large\texttt{Shuttle}}       & {\Large  63.146    }                               & {\Large  4.426      }                      & {\Large  0.987  }                         & {\Large  594.589   }                                                & {\Large  6.306  }                           \\
{\Large\texttt{FMNIST}}        & {\Large  52.114    }                               & {\Large  15.446        }                   & {\Large  4.298  }                         & {\Large  744.652  }                                               & {\Large  12.293 }                           \\
{\Large\texttt{WM-811K}}        & {\Large  11498.385  }                              & {\Large  910.600          }                & {\Large  89.499 }                         & {\Large  4561.028  }                              & {\Large  9.498  }                           \\ 
{\Large\texttt{Agnews}}        & {\Large  25.877  }                              & {\Large  3.348              }            & {\Large  3.156  }                        & {\Large  138.114      }                          & {\Large  7.506     }                        \\ 
\bottomrule
\end{tabular}
}
\label{tab:running_time}
\end{table}

\subsection{Ablation study}
\label{sec:ablation_study}

We conduct additional experiments to cover how the ODIM behaves with respect to the choice of the hyper-parameters, whose results are given in Figure \ref{fig:ablation_nens_lr_auc_result}.  
The followings are the summary of our ablation studies, whose detailed explanations are in Appendix C. 

\begin{enumerate}
    \item Increasing the value of $K$ leads to better performance, and the improvement saturates when $K\ge 50$. 
    \item Large value of $N_{\text{pat}}$ is beneficial, though the extent of improvement diminishes when $N_{\text{pat}}\ge 10$.
    \item Using a larger value of $B$ generally improves the outlier identification performance.
    \item The ODIM is stable with respect to reasonable learning rates, highlighting the ease of applicability in practice. 
\end{enumerate}

Furthermore, we apply two pre-processing techniques, min-max scaling and standardization, and compare their corresponding results for the ODIMs on 30 tabular datasets. 
We observe that the standardization yields the averaged AUC score of 0.760, which is worse than that when the min-max scaling is used, 0.790. 
The result for each dataset can be found in Appendix C.

\section{Further discussions}
\label{sec:5}

\paragraph{ODIM with partially labeled outliers}
Our method can be extended to a scenario where partially labeled data are additionally given. 
That is, we assume that besides $\cU^{tr}$, a few labeled outlier dataset  $\cL^{tr}=\{(\bx_1^l,1),\ldots,(\bx_m^l,1)\}$ is also available. 

We simply adopt the idea of \citet{daniel2019deep}, which encourages the log-likelihood of known outliers to decrease with the variational \textit{upper} bound, called $\chi$ upper bound (CUBO). 
And we modify the loss function of the ODIM by adding the expected CUBO on $\cL^{tr}$ from the original IWAE. 

Table \ref{app_tab:auc_result_label} summarizes the averaged training AUC and PR results of the modified method across various tabular datasets for different proportions of labeled outliers. 
Please refer to Appendix D for detailed results of each dataset. 
It is clearly seen that using label information helps to enhance identifying performance by a large margin. 
Rigorous descriptions of this modification and further discussions are provided in Appendix D. 

\begin{table}[t]
\caption{Averaged results of training AUC (and PR) scores with various values of $l$. 
We consider $l$, $l=0.0$, $0.3$, $0.5$.
}
\centering
\resizebox{0.37\textwidth}{!}{
    \begin{tabular}{l|ccc}
    \hline
        $l$ & $0.0$ & $0.3$ & $0.5$ \\ \hline
        AUC (PR) & 0.885 (0.647) & 0.947 (0.871) & 0.958 (0.891) \\ \hline

\end{tabular}
}
\label{app_tab:auc_result_label}
\end{table}

\begin{table}[t]
\caption{Averaged results of training AUC (and PR) scores when applying the DP-SGD algorithm. We iterate the DP-SGD until $\epsilon=10$ while fixing $\delta=10^{-5}$.
}
\centering
\resizebox{0.3\textwidth}{!}{
    \begin{tabular}{l|ccc}
    \hline
        Method & DeepSVDD & ODIM  \\ \hline
        AUC (PR) & 0.614 (0.152) & \textbf{0.710} (\textbf{0.234}) \\ \hline

\end{tabular}
}
\label{app_tab:dp_result}
\end{table}


\paragraph{Differentially private ODIM}
As the ODIM typically requires far fewer updates, combining our method with differentially private algorithms is expected to yield a synergic benefits. 
We impose privacy protection to our method by simply applying DP-SGD \citep{abadi2016deep}, instead of conventional SGD-based methods, when minimizing the loss function (\ref{eq:iwae_obj}). 
To guarantee differential privacy (DP), DP-SGD deforms gradient of each sample by clipping and adding noise, and parameters are trained based on this modified quantity. 

As a measure of DP, we adopt $(\epsilon,\delta)$-DP \citep{10.1007/11787006_1}, which is the de facto standard in this field. 
We iterate DP-SGD to train DGMs for obtaining the ODIM scores, until the privacy budget $\epsilon$ first exceeds a pre-specified value while fixing $\delta=10^{-5}$. 
Detailed explanations for $(\epsilon,\delta)$-DP and DP-SGD, and the calculation of privacy budget when utilizing DP-SGD is provided in Appendix D. 

Table \ref{app_tab:dp_result} shows the averaged AUC and PR scores of our method and DeepSVDD across several tabular dataset, both trained until $\epsilon=10$. 
Detailed results for each dataset can be found in Appendix D. 
We note that we exclude non-SGD-based methods such as OCSVM and IF  since they DP-SGD is not applicable to them. 
Large margins in AUC and PR between our method and DeepSVDD indicates the strong potential of the ODIM when publishing OD algorithms with privacy guarantee. 

\section{Concluding remarks}
\label{sec:6}

This paper proposed a powerful yet efficient UOD method called the ODIM. 
The ODIM is inspired by a new observation called the IM effect, that deep generative models tend to memorize inliers first during early training.
Combined with the technique to select the optimal number of training updates and the ensemble method, we showed that the ODIM provides consistently superior results in identifying outliers, regardless of data types, with significantly faster running times.
Exploring the optimal pre-processing technique for the ODIM and developing new methods discussed in Section \ref{sec:5} would be promising future works.

\section*{Impact Statement}
Our proposed method called ODIM enhances the accuracy and efficiency of anomaly detection across various domains, including healthcare, finance, and astronomy. 
ODIM's effectiveness and versatility can extend its applicability, benefiting a wide range of sectors and contributing to societal advancements in safety, security, and operational efficiency.
Additionally, its computational efficiency also makes it accessible for real-time applications, promoting its use in dynamic environments such as industrial monitoring. 

\section*{Acknowledgements}
DK was supported by the National Research Foundation of Korea(NRF) grant funded by the Korea government(MSIT) (No. NRF-2022R1G1A1010894 and No. RS-2023-00218231).
YK was supported by the National Research Foundation of Korea(NRF) grant funded by the Korea government(MSIT)
 (No. 2020R1A2C3A01003550) and Institute of Information \& communications Technology 
Planning \& Evaluation (IITP) grant funded by the Korea government(MSIT) [NO. RS-2021-II211343, 
Artificial Intelligence Graduate School Program (Seoul National University)].

\bibliography{references}

\begin{thebibliography}{72}
\providecommand{\natexlab}[1]{#1}
\providecommand{\url}[1]{\texttt{#1}}
\expandafter\ifx\csname urlstyle\endcsname\relax
  \providecommand{\doi}[1]{doi: #1}\else
  \providecommand{\doi}{doi: \begingroup \urlstyle{rm}\Url}\fi

\bibitem[Abadi et~al.(2016)Abadi, Chu, Goodfellow, McMahan, Mironov, Talwar,
  and Zhang]{abadi2016deep}
Abadi, M., Chu, A., Goodfellow, I., McMahan, H.~B., Mironov, I., Talwar, K.,
  and Zhang, L.
\newblock Deep learning with differential privacy.
\newblock In \emph{Proceedings of the 2016 ACM SIGSAC conference on computer
  and communications security}, pp.\  308--318, 2016.

\bibitem[Arpit et~al.(2017)Arpit, Jastrzebski, Ballas, Krueger, Bengio, Kanwal,
  Maharaj, Fischer, Courville, Bengio, et~al.]{arpit2017closer}
Arpit, D., Jastrzebski, S., Ballas, N., Krueger, D., Bengio, E., Kanwal, M.~S.,
  Maharaj, T., Fischer, A., Courville, A., Bengio, Y., et~al.
\newblock A closer look at memorization in deep networks.
\newblock In \emph{International Conference on Machine Learning}, pp.\
  233--242. PMLR, 2017.

\bibitem[Bergman \& Hoshen(2020)Bergman and Hoshen]{goad}
Bergman, L. and Hoshen, Y.
\newblock Classification-based anomaly detection for general data.
\newblock In \emph{International Conference on Learning Representations}, 2020.

\bibitem[Breunig et~al.(2000{\natexlab{a}})Breunig, Kriegel, Ng, and
  Sander]{10.1145/335191.335388}
Breunig, M.~M., Kriegel, H.-P., Ng, R.~T., and Sander, J.
\newblock Lof: Identifying density-based local outliers.
\newblock \emph{SIGMOD Rec.}, 29\penalty0 (2):\penalty0 93–104, may
  2000{\natexlab{a}}.
\newblock ISSN 0163-5808.
\newblock \doi{10.1145/335191.335388}.

\bibitem[Breunig et~al.(2000{\natexlab{b}})Breunig, Kriegel, Ng, and
  Sander]{breunig2000LOF}
Breunig, M.~M., Kriegel, H.-P., Ng, R.~T., and Sander, J.
\newblock Lof: identifying density-based local outliers.
\newblock In \emph{Proceedings of the 2000 ACM SIGMOD international conference
  on Management of data}, pp.\  93--104, 2000{\natexlab{b}}.

\bibitem[Burda et~al.(2016)Burda, Grosse, and
  Salakhutdinov]{DBLP:journals/corr/BurdaGS15}
Burda, Y., Grosse, R.~B., and Salakhutdinov, R.
\newblock Importance weighted autoencoders.
\newblock In Bengio, Y. and LeCun, Y. (eds.), \emph{4th International
  Conference on Learning Representations, {ICLR} 2016, San Juan, Puerto Rico,
  May 2-4, 2016, Conference Track Proceedings}, 2016.

\bibitem[Chen et~al.(2020{\natexlab{a}})Chen, Orekondy, and Fritz]{chen2020gs}
Chen, D., Orekondy, T., and Fritz, M.
\newblock Gs-wgan: A gradient-sanitized approach for learning differentially
  private generators.
\newblock \emph{Advances in Neural Information Processing Systems},
  33:\penalty0 12673--12684, 2020{\natexlab{a}}.

\bibitem[Chen et~al.(2020{\natexlab{b}})Chen, Kornblith, Norouzi, and
  Hinton]{simclr}
Chen, T., Kornblith, S., Norouzi, M., and Hinton, G.
\newblock A simple framework for contrastive learning of visual
  representations.
\newblock In III, H.~D. and Singh, A. (eds.), \emph{Proceedings of the 37th
  International Conference on Machine Learning}, volume 119 of
  \emph{Proceedings of Machine Learning Research}, pp.\  1597--1607. PMLR,
  13--18 Jul 2020{\natexlab{b}}.

\bibitem[Daniel et~al.(2019)Daniel, Kurutach, and Tamar]{daniel2019deep}
Daniel, T., Kurutach, T., and Tamar, A.
\newblock Deep variational semi-supervised novelty detection.
\newblock \emph{arXiv preprint arXiv:1911.04971}, 2019.

\bibitem[Devlin et~al.(2019)Devlin, Chang, Lee, and
  Toutanova]{devlin-etal-2019-bert}
Devlin, J., Chang, M.-W., Lee, K., and Toutanova, K.
\newblock {BERT}: Pre-training of deep bidirectional transformers for language
  understanding.
\newblock In \emph{Proceedings of the 2019 Conference of the North {A}merican
  Chapter of the Association for Computational Linguistics: Human Language
  Technologies, Volume 1 (Long and Short Papers)}, pp.\  4171--4186,
  Minneapolis, Minnesota, June 2019. Association for Computational Linguistics.
\newblock \doi{10.18653/v1/N19-1423}.

\bibitem[Dinh et~al.(2015)Dinh, Krueger, and
  Bengio]{DBLP:journals/corr/DinhKB14}
Dinh, L., Krueger, D., and Bengio, Y.
\newblock {NICE:} non-linear independent components estimation.
\newblock In Bengio, Y. and LeCun, Y. (eds.), \emph{3rd International
  Conference on Learning Representations, {ICLR} 2015, San Diego, CA, USA, May
  7-9, 2015, Workshop Track Proceedings}, 2015.

\bibitem[Dinh et~al.(2017)Dinh, Sohl{-}Dickstein, and
  Bengio]{DBLP:conf/iclr/DinhSB17}
Dinh, L., Sohl{-}Dickstein, J., and Bengio, S.
\newblock Density estimation using real {NVP}.
\newblock In \emph{5th International Conference on Learning Representations,
  {ICLR} 2017, Toulon, France, April 24-26, 2017, Conference Track
  Proceedings}. OpenReview.net, 2017.

\bibitem[Dong et~al.(2019)Dong, Roth, and Su]{dong2019gaussian}
Dong, J., Roth, A., and Su, W.~J.
\newblock Gaussian differential privacy.
\newblock \emph{arXiv preprint arXiv:1905.02383}, 2019.

\bibitem[Dosovitskiy et~al.(2021)Dosovitskiy, Beyer, Kolesnikov, Weissenborn,
  Zhai, Unterthiner, Dehghani, Minderer, Heigold, Gelly, Uszkoreit, and
  Houlsby]{DBLP:conf/iclr/DosovitskiyB0WZ21}
Dosovitskiy, A., Beyer, L., Kolesnikov, A., Weissenborn, D., Zhai, X.,
  Unterthiner, T., Dehghani, M., Minderer, M., Heigold, G., Gelly, S.,
  Uszkoreit, J., and Houlsby, N.
\newblock An image is worth 16x16 words: Transformers for image recognition at
  scale.
\newblock In \emph{9th International Conference on Learning Representations,
  {ICLR} 2021, Virtual Event, Austria, May 3-7, 2021}. OpenReview.net, 2021.

\bibitem[Dwork(2006)]{10.1007/11787006_1}
Dwork, C.
\newblock Differential privacy.
\newblock In Bugliesi, M., Preneel, B., Sassone, V., and Wegener, I. (eds.),
  \emph{Automata, Languages and Programming}, pp.\  1--12, Berlin, Heidelberg,
  2006. Springer Berlin Heidelberg.
\newblock ISBN 978-3-540-35908-1.

\bibitem[Fauconnier \& Haesbroeck(2009)Fauconnier and
  Haesbroeck]{fauconnier2009outliers}
Fauconnier, C. and Haesbroeck, G.
\newblock Outliers detection with the minimum covariance determinant estimator
  in practice.
\newblock \emph{Statistical Methodology}, 6\penalty0 (4):\penalty0 363--379,
  2009.

\bibitem[Golan \& El-Yaniv(2018)Golan and El-Yaniv]{geom}
Golan, I. and El-Yaniv, R.
\newblock Deep anomaly detection using geometric transformations.
\newblock In \emph{Proceedings of the 32nd International Conference on Neural
  Information Processing Systems}, NIPS'18, pp.\  9781–9791, Red Hook, NY,
  USA, 2018. Curran Associates Inc.

\bibitem[Goldstein \& Dengel(2012)Goldstein and Dengel]{goldstein2012histogram}
Goldstein, M. and Dengel, A.
\newblock Histogram-based outlier score (hbos): A fast unsupervised anomaly
  detection algorithm.
\newblock \emph{KI-2012: poster and demo track}, 1:\penalty0 59--63, 2012.

\bibitem[Gomes et~al.(2022)Gomes, Alberge, Duhamel, and
  Piantanida]{gomes2022igeood}
Gomes, E. D.~C., Alberge, F., Duhamel, P., and Piantanida, P.
\newblock Igeood: An information geometry approach to out-of-distribution
  detection.
\newblock In \emph{International Conference on Learning Representations}, 2022.

\bibitem[Goyal et~al.(2020)Goyal, Raghunathan, Jain, Simhadri, and
  Jain]{DBLP:conf/icml/GoyalRJS020}
Goyal, S., Raghunathan, A., Jain, M., Simhadri, H.~V., and Jain, P.
\newblock {DROCC:} deep robust one-class classification.
\newblock In \emph{Proceedings of the 37th International Conference on Machine
  Learning, {ICML} 2020, 13-18 July 2020, Virtual Event}, volume 119 of
  \emph{Proceedings of Machine Learning Research}, pp.\  3711--3721. {PMLR},
  2020.

\bibitem[Han et~al.(2022{\natexlab{a}})Han, Hu, Huang, Jiang, and
  Zhao]{DBLP:conf/nips/HanHHJ022}
Han, S., Hu, X., Huang, H., Jiang, M., and Zhao, Y.
\newblock Adbench: Anomaly detection benchmark.
\newblock In \emph{NeurIPS}, 2022{\natexlab{a}}.

\bibitem[Han et~al.(2022{\natexlab{b}})Han, Hu, Huang, Jiang, and
  Zhao]{han2022adbench}
Han, S., Hu, X., Huang, H., Jiang, M., and Zhao, Y.
\newblock Adbench: Anomaly detection benchmark.
\newblock In \emph{Neural Information Processing Systems (NeurIPS)},
  2022{\natexlab{b}}.

\bibitem[He et~al.(2003)He, Xu, and Deng]{he2003discovering}
He, Z., Xu, X., and Deng, S.
\newblock Discovering cluster-based local outliers.
\newblock \emph{Pattern recognition letters}, 24\penalty0 (9-10):\penalty0
  1641--1650, 2003.

\bibitem[Hendrycks \& Gimpel(2017)Hendrycks and Gimpel]{hendrycks17baseline}
Hendrycks, D. and Gimpel, K.
\newblock A baseline for detecting misclassified and out-of-distribution
  examples in neural networks.
\newblock \emph{Proceedings of International Conference on Learning
  Representations}, 2017.

\bibitem[Ho et~al.(2020)Ho, Jain, and Abbeel]{DBLP:conf/nips/HoJA20}
Ho, J., Jain, A., and Abbeel, P.
\newblock Denoising diffusion probabilistic models.
\newblock In Larochelle, H., Ranzato, M., Hadsell, R., Balcan, M., and Lin, H.
  (eds.), \emph{Advances in Neural Information Processing Systems 33: Annual
  Conference on Neural Information Processing Systems 2020, NeurIPS 2020,
  December 6-12, 2020, virtual}, 2020.

\bibitem[Jiang et~al.(2022)Jiang, Sun, and Yu]{jiang2022revisiting}
Jiang, D., Sun, S., and Yu, Y.
\newblock Revisiting flow generative models for out-of-distribution detection.
\newblock In \emph{International Conference on Learning Representations}, 2022.

\bibitem[Jiang et~al.(2018)Jiang, Zhou, Leung, Li, and
  Fei-Fei]{jiang2018mentornet}
Jiang, L., Zhou, Z., Leung, T., Li, L.-J., and Fei-Fei, L.
\newblock Mentornet: Learning data-driven curriculum for very deep neural
  networks on corrupted labels.
\newblock In \emph{International Conference on Machine Learning}, pp.\
  2304--2313. PMLR, 2018.

\bibitem[Kim et~al.(2020)Kim, Hwang, and Kim]{kim2020casting}
Kim, D., Hwang, J., and Kim, Y.
\newblock On casting importance weighted autoencoder to an em algorithm to
  learn deep generative models.
\newblock In \emph{International Conference on Artificial Intelligence and
  Statistics}, pp.\  2153--2163. PMLR, 2020.

\bibitem[Kingma \& Ba(2014)Kingma and Ba]{kingma2014adam}
Kingma, D.~P. and Ba, J.
\newblock Adam: A method for stochastic optimization.
\newblock \emph{arXiv preprint arXiv:1412.6980}, 2014.

\bibitem[Kingma \& Dhariwal(2018)Kingma and Dhariwal]{DBLP:conf/nips/KingmaD18}
Kingma, D.~P. and Dhariwal, P.
\newblock Glow: Generative flow with invertible 1x1 convolutions.
\newblock In Bengio, S., Wallach, H.~M., Larochelle, H., Grauman, K.,
  Cesa{-}Bianchi, N., and Garnett, R. (eds.), \emph{Advances in Neural
  Information Processing Systems 31: Annual Conference on Neural Information
  Processing Systems 2018, NeurIPS 2018, December 3-8, 2018, Montr{\'{e}}al,
  Canada}, pp.\  10236--10245, 2018.

\bibitem[Kingma \& Welling(2013)Kingma and Welling]{kingma2013auto}
Kingma, D.~P. and Welling, M.
\newblock Auto-encoding variational bayes.
\newblock \emph{arXiv preprint arXiv:1312.6114}, 2013.

\bibitem[Kobyzev et~al.(2020)Kobyzev, Prince, and
  Brubaker]{kobyzev2020normalizing}
Kobyzev, I., Prince, S.~J., and Brubaker, M.~A.
\newblock Normalizing flows: An introduction and review of current methods.
\newblock \emph{IEEE transactions on pattern analysis and machine
  intelligence}, 43\penalty0 (11):\penalty0 3964--3979, 2020.

\bibitem[Kriegel et~al.(2009)Kriegel, Kr{\"o}ger, Schubert, and
  Zimek]{kriegel2009outlier}
Kriegel, H.-P., Kr{\"o}ger, P., Schubert, E., and Zimek, A.
\newblock Outlier detection in axis-parallel subspaces of high dimensional
  data.
\newblock In \emph{Advances in Knowledge Discovery and Data Mining: 13th
  Pacific-Asia Conference, PAKDD 2009 Bangkok, Thailand, April 27-30, 2009
  Proceedings 13}, pp.\  831--838. Springer, 2009.

\bibitem[Lai et~al.(2020)Lai, Zou, and Lerman]{lai2020robust}
Lai, C.-H., Zou, D., and Lerman, G.
\newblock Robust subspace recovery layer for unsupervised anomaly detection.
\newblock In \emph{International Conference on Learning Representations}, 2020.

\bibitem[Lan \& Dinh(2021)Lan and Dinh]{DBLP:journals/entropy/LanD21}
Lan, C.~L. and Dinh, L.
\newblock Perfect density models cannot guarantee anomaly detection.
\newblock \emph{Entropy}, 23\penalty0 (12):\penalty0 1690, 2021.
\newblock \doi{10.3390/E23121690}.

\bibitem[Lazarevic \& Kumar(2005)Lazarevic and Kumar]{lazarevic2005feature}
Lazarevic, A. and Kumar, V.
\newblock Feature bagging for outlier detection.
\newblock In \emph{Proceedings of the eleventh ACM SIGKDD international
  conference on Knowledge discovery in data mining}, pp.\  157--166, 2005.

\bibitem[Li et~al.(2020)Li, Zhao, Botta, Ionescu, and Hu]{li2020copod}
Li, Z., Zhao, Y., Botta, N., Ionescu, C., and Hu, X.
\newblock Copod: copula-based outlier detection.
\newblock In \emph{2020 IEEE international conference on data mining (ICDM)},
  pp.\  1118--1123. IEEE, 2020.

\bibitem[Li et~al.(2022)Li, Zhao, Hu, Botta, Ionescu, and Chen]{li2022ecod}
Li, Z., Zhao, Y., Hu, X., Botta, N., Ionescu, C., and Chen, G.
\newblock Ecod: Unsupervised outlier detection using empirical cumulative
  distribution functions.
\newblock \emph{IEEE Transactions on Knowledge and Data Engineering}, 2022.

\bibitem[Liang et~al.(2018)Liang, Li, and Srikant]{liang2018enhancing}
Liang, S., Li, Y., and Srikant, R.
\newblock Enhancing the reliability of out-of-distribution image detection in
  neural networks.
\newblock In \emph{International Conference on Learning Representations}, 2018.

\bibitem[Liu et~al.(2008{\natexlab{a}})Liu, Ting, and Zhou]{4781136}
Liu, F.~T., Ting, K.~M., and Zhou, Z.-H.
\newblock Isolation forest.
\newblock In \emph{2008 Eighth IEEE International Conference on Data Mining},
  pp.\  413--422, 2008{\natexlab{a}}.
\newblock \doi{10.1109/ICDM.2008.17}.

\bibitem[Liu et~al.(2008{\natexlab{b}})Liu, Ting, and Zhou]{liu2008isolation}
Liu, F.~T., Ting, K.~M., and Zhou, Z.-H.
\newblock Isolation forest.
\newblock In \emph{2008 eighth ieee international conference on data mining},
  pp.\  413--422. IEEE, 2008{\natexlab{b}}.

\bibitem[Liu et~al.(2014)Liu, Hua, and Smith]{6909884}
Liu, W., Hua, G., and Smith, J.~R.
\newblock Unsupervised one-class learning for automatic outlier removal.
\newblock In \emph{2014 IEEE Conference on Computer Vision and Pattern
  Recognition}, pp.\  3826--3833, 2014.
\newblock \doi{10.1109/CVPR.2014.483}.

\bibitem[Liu et~al.(2019)Liu, Ott, Goyal, Du, Joshi, Chen, Levy, Lewis,
  Zettlemoyer, and Stoyanov]{DBLP:journals/corr/abs-1907-11692}
Liu, Y., Ott, M., Goyal, N., Du, J., Joshi, M., Chen, D., Levy, O., Lewis, M.,
  Zettlemoyer, L., and Stoyanov, V.
\newblock Roberta: {A} robustly optimized {BERT} pretraining approach.
\newblock \emph{CoRR}, abs/1907.11692, 2019.

\bibitem[Livernoche et~al.(2023)Livernoche, Jain, Hezaveh, and
  Ravanbakhsh]{DBLP:journals/corr/abs-2305-18593}
Livernoche, V., Jain, V., Hezaveh, Y., and Ravanbakhsh, S.
\newblock On diffusion modeling for anomaly detection.
\newblock \emph{CoRR}, abs/2305.18593, 2023.
\newblock \doi{10.48550/ARXIV.2305.18593}.

\bibitem[Mahmood et~al.(2021)Mahmood, Oliva, and Styner]{mahmood2021multiscale}
Mahmood, A., Oliva, J., and Styner, M.~A.
\newblock Multiscale score matching for out-of-distribution detection.
\newblock In \emph{International Conference on Learning Representations}, 2021.

\bibitem[Mironov(2017)]{mironov2017renyi}
Mironov, I.
\newblock R{\'e}nyi differential privacy.
\newblock In \emph{2017 IEEE 30th computer security foundations symposium
  (CSF)}, pp.\  263--275. IEEE, 2017.

\bibitem[Nalisnick et~al.(2019{\natexlab{a}})Nalisnick, Matsukawa, Teh, and
  Lakshminarayanan]{https://doi.org/10.48550/arxiv.1906.02994}
Nalisnick, E., Matsukawa, A., Teh, Y.~W., and Lakshminarayanan, B.
\newblock Detecting out-of-distribution inputs to deep generative models using
  typicality, 2019{\natexlab{a}}.

\bibitem[Nalisnick et~al.()Nalisnick, Matsukawa, Teh, and
  Lakshminarayanan]{DBLP:journals/corr/abs-1906-02994}
Nalisnick, E.~T., Matsukawa, A., Teh, Y.~W., and Lakshminarayanan, B.
\newblock Detecting out-of-distribution inputs to deep generative models using
  a test for typicality.
\newblock \emph{CoRR}, abs/1906.02994.

\bibitem[Nalisnick et~al.(2019{\natexlab{b}})Nalisnick, Matsukawa, Teh,
  G{\"{o}}r{\"{u}}r, and Lakshminarayanan]{DBLP:conf/iclr/NalisnickMTGL19}
Nalisnick, E.~T., Matsukawa, A., Teh, Y.~W., G{\"{o}}r{\"{u}}r, D., and
  Lakshminarayanan, B.
\newblock Do deep generative models know what they don't know?
\newblock In \emph{7th International Conference on Learning Representations,
  {ICLR} 2019, New Orleans, LA, USA, May 6-9, 2019}. OpenReview.net,
  2019{\natexlab{b}}.

\bibitem[Pevn{\`y}(2016)]{pevny2016loda}
Pevn{\`y}, T.
\newblock Loda: Lightweight on-line detector of anomalies.
\newblock \emph{Machine Learning}, 102:\penalty0 275--304, 2016.

\bibitem[Ramaswamy et~al.(2000)Ramaswamy, Rastogi, and
  Shim]{ramaswamy2000efficient}
Ramaswamy, S., Rastogi, R., and Shim, K.
\newblock Efficient algorithms for mining outliers from large data sets.
\newblock In \emph{Proceedings of the 2000 ACM SIGMOD international conference
  on Management of data}, pp.\  427--438, 2000.

\bibitem[Ruff et~al.(2018{\natexlab{a}})Ruff, Vandermeulen, Goernitz, Deecke,
  Siddiqui, Binder, M{\"u}ller, and Kloft]{deepsvdd}
Ruff, L., Vandermeulen, R., Goernitz, N., Deecke, L., Siddiqui, S.~A., Binder,
  A., M{\"u}ller, E., and Kloft, M.
\newblock Deep one-class classification.
\newblock In Dy, J. and Krause, A. (eds.), \emph{Proceedings of the 35th
  International Conference on Machine Learning}, volume~80 of \emph{Proceedings
  of Machine Learning Research}, pp.\  4393--4402. PMLR, 10--15 Jul
  2018{\natexlab{a}}.

\bibitem[Ruff et~al.(2018{\natexlab{b}})Ruff, Vandermeulen, Goernitz, Deecke,
  Siddiqui, Binder, M{\"u}ller, and Kloft]{ruff2018deep}
Ruff, L., Vandermeulen, R., Goernitz, N., Deecke, L., Siddiqui, S.~A., Binder,
  A., M{\"u}ller, E., and Kloft, M.
\newblock Deep one-class classification.
\newblock In \emph{International conference on machine learning}, pp.\
  4393--4402. PMLR, 2018{\natexlab{b}}.

\bibitem[Ruff et~al.(2020)Ruff, Vandermeulen, Görnitz, Binder, Müller,
  Müller, and Kloft]{deepsad}
Ruff, L., Vandermeulen, R.~A., Görnitz, N., Binder, A., Müller, E., Müller,
  K.-R., and Kloft, M.
\newblock Deep semi-supervised anomaly detection.
\newblock In \emph{International Conference on Learning Representations}, 2020.

\bibitem[Ryu et~al.(2018)Ryu, Koo, Yu, and Lee]{ryu-etal-2018-domain}
Ryu, S., Koo, S., Yu, H., and Lee, G.~G.
\newblock Out-of-domain detection based on generative adversarial network.
\newblock In \emph{Proceedings of the 2018 Conference on Empirical Methods in
  Natural Language Processing}, pp.\  714--718, Brussels, Belgium,
  October-November 2018. Association for Computational Linguistics.
\newblock \doi{10.18653/v1/D18-1077}.

\bibitem[Salimans et~al.(2017)Salimans, Karpathy, Chen, and
  Kingma]{DBLP:conf/iclr/SalimansK0K17}
Salimans, T., Karpathy, A., Chen, X., and Kingma, D.~P.
\newblock Pixelcnn++: Improving the pixelcnn with discretized logistic mixture
  likelihood and other modifications.
\newblock In \emph{5th International Conference on Learning Representations,
  {ICLR} 2017, Toulon, France, April 24-26, 2017, Conference Track
  Proceedings}. OpenReview.net, 2017.

\bibitem[Schölkopf et~al.(2001)Schölkopf, Platt, Shawe-Taylor, Smola, and
  Williamson]{ocsvm}
Schölkopf, B., Platt, J., Shawe-Taylor, J., Smola, A., and Williamson, R.
\newblock Estimating support of a high-dimensional distribution.
\newblock \emph{Neural Computation}, 13:\penalty0 1443--1471, 07 2001.
\newblock \doi{10.1162/089976601750264965}.

\bibitem[Sehwag et~al.(2021)Sehwag, Chiang, and Mittal]{ssd}
Sehwag, V., Chiang, M., and Mittal, P.
\newblock {\{}SSD{\}}: A unified framework for self-supervised outlier
  detection.
\newblock In \emph{International Conference on Learning Representations}, 2021.

\bibitem[Shenkar \& Wolf(2022)Shenkar and Wolf]{DBLP:conf/iclr/ShenkarW22}
Shenkar, T. and Wolf, L.
\newblock Anomaly detection for tabular data with internal contrastive
  learning.
\newblock In \emph{The Tenth International Conference on Learning
  Representations, {ICLR} 2022, Virtual Event, April 25-29, 2022}.
  OpenReview.net, 2022.

\bibitem[Shyu et~al.(2003)Shyu, Chen, Sarinnapakorn, and Chang]{shyu2003novel}
Shyu, M.-L., Chen, S.-C., Sarinnapakorn, K., and Chang, L.
\newblock A novel anomaly detection scheme based on principal component
  classifier.
\newblock In \emph{Proceedings of the IEEE foundations and new directions of
  data mining workshop}, pp.\  172--179. IEEE Press, 2003.

\bibitem[Tack et~al.(2020)Tack, Mo, Jeong, and Shin]{csi}
Tack, J., Mo, S., Jeong, J., and Shin, J.
\newblock Csi: Novelty detection via contrastive learning on distributionally
  shifted instances.
\newblock In Larochelle, H., Ranzato, M., Hadsell, R., Balcan, M.~F., and Lin,
  H. (eds.), \emph{Advances in Neural Information Processing Systems},
  volume~33, pp.\  11839--11852. Curran Associates, Inc., 2020.

\bibitem[Tang et~al.(2002)Tang, Chen, Fu, and Cheung]{tang2002enhancing}
Tang, J., Chen, Z., Fu, A. W.-C., and Cheung, D.~W.
\newblock Enhancing effectiveness of outlier detections for low density
  patterns.
\newblock In \emph{Advances in Knowledge Discovery and Data Mining: 6th
  Pacific-Asia Conference, PAKDD 2002 Taipei, Taiwan, May 6--8, 2002
  Proceedings 6}, pp.\  535--548. Springer, 2002.

\bibitem[Tax \& Duin(2004)Tax and Duin]{tax2004support}
Tax, D.~M. and Duin, R.~P.
\newblock Support vector data description.
\newblock \emph{Machine learning}, 54\penalty0 (1):\penalty0 45--66, 2004.

\bibitem[Tomczak \& Welling(2018)Tomczak and Welling]{tomczak2018vae}
Tomczak, J. and Welling, M.
\newblock Vae with a vampprior.
\newblock In \emph{International Conference on Artificial Intelligence and
  Statistics}, pp.\  1214--1223. PMLR, 2018.

\bibitem[van~den Oord et~al.(2016)van~den Oord, Kalchbrenner, and
  Kavukcuoglu]{DBLP:conf/icml/OordKK16}
van~den Oord, A., Kalchbrenner, N., and Kavukcuoglu, K.
\newblock Pixel recurrent neural networks.
\newblock In Balcan, M. and Weinberger, K.~Q. (eds.), \emph{Proceedings of the
  33nd International Conference on Machine Learning, {ICML} 2016, New York
  City, NY, USA, June 19-24, 2016}, volume~48 of \emph{{JMLR} Workshop and
  Conference Proceedings}, pp.\  1747--1756. JMLR.org, 2016.

\bibitem[Wang et~al.(2019)Wang, Zeng, Liu, Zhu, Yin, Xu, and
  Kloft]{NEURIPS2019_6c4bb406}
Wang, S., Zeng, Y., Liu, X., Zhu, E., Yin, J., Xu, C., and Kloft, M.
\newblock Effective end-to-end unsupervised outlier detection via inlier
  priority of discriminative network.
\newblock In Wallach, H., Larochelle, H., Beygelzimer, A., d\textquotesingle
  Alch\'{e}-Buc, F., Fox, E., and Garnett, R. (eds.), \emph{Advances in Neural
  Information Processing Systems}, volume~32. Curran Associates, Inc., 2019.

\bibitem[Xia et~al.(2015)Xia, Cao, Wen, Hua, and Sun]{7410534}
Xia, Y., Cao, X., Wen, F., Hua, G., and Sun, J.
\newblock Learning discriminative reconstructions for unsupervised outlier
  removal.
\newblock In \emph{2015 IEEE International Conference on Computer Vision
  (ICCV)}, pp.\  1511--1519, 2015.
\newblock \doi{10.1109/ICCV.2015.177}.

\bibitem[Zhai et~al.(2016)Zhai, Cheng, Lu, and Zhang]{10.5555/3045390.3045507}
Zhai, S., Cheng, Y., Lu, W., and Zhang, Z.
\newblock Deep structured energy based models for anomaly detection.
\newblock In \emph{Proceedings of the 33rd International Conference on
  International Conference on Machine Learning - Volume 48}, ICML'16, pp.\
  1100–1109. JMLR.org, 2016.

\bibitem[Zhao et~al.(2023)Zhao, Kunar, Birke, Van~der Scheer, and
  Chen]{zhao2023ctab}
Zhao, Z., Kunar, A., Birke, R., Van~der Scheer, H., and Chen, L.~Y.
\newblock Ctab-gan+: Enhancing tabular data synthesis.
\newblock \emph{Frontiers in big Data}, 6, 2023.

\bibitem[Zhou \& Paffenroth(2017)Zhou and Paffenroth]{rda}
Zhou, C. and Paffenroth, R.~C.
\newblock Anomaly detection with robust deep autoencoders.
\newblock KDD '17, pp.\  665–674, New York, NY, USA, 2017. Association for
  Computing Machinery.
\newblock ISBN 9781450348874.
\newblock \doi{10.1145/3097983.3098052}.

\bibitem[Zong et~al.(2018{\natexlab{a}})Zong, Song, Min, Cheng, Lumezanu, Cho,
  and Chen]{DBLP:conf/iclr/ZongSMCLCC18}
Zong, B., Song, Q., Min, M.~R., Cheng, W., Lumezanu, C., Cho, D., and Chen, H.
\newblock Deep autoencoding gaussian mixture model for unsupervised anomaly
  detection.
\newblock In \emph{6th International Conference on Learning Representations,
  {ICLR} 2018, Vancouver, BC, Canada, April 30 - May 3, 2018, Conference Track
  Proceedings}. OpenReview.net, 2018{\natexlab{a}}.

\bibitem[Zong et~al.(2018{\natexlab{b}})Zong, Song, Min, Cheng, Lumezanu, Cho,
  and Chen]{dagmm}
Zong, B., Song, Q., Min, M.~R., Cheng, W., Lumezanu, C., Cho, D., and Chen, H.
\newblock Deep autoencoding gaussian mixture model for unsupervised anomaly
  detection.
\newblock In \emph{International Conference on Learning Representations},
  2018{\natexlab{b}}.

\end{thebibliography}
\bibliographystyle{icml2024}

\newpage
\appendix
\onecolumn

\section*{A. Proof of Propositions}
\label{appendix:A}
\subsection*{A.1. Proof of Proposition 1}

Note that the objective function of the VAE is given as:
\bean
L^{\text{VAE}}(\theta,\phi;\bx):=\int_\bz \log \left( \frac{p(\bx|\bz;\theta)p(\bz)}{q(\bz|\bx;\phi)} \right) \cdot q(\bz|\bx;\phi)d\bz.
\eean
Thus, we have the equations:
{\normalsize
\bean
\frac{\partial}{\partial w_{ij}}L^{\text{VAE}}(\theta,\phi;\bx)=\int_\bz\frac{\partial}{\partial w_{ij}} \log \left( p(\bx|\bz;\theta)\right) \cdot q(\bz|\bx;\phi)d\bz,
\eean
}
and
{\normalsize
\bean
\frac{\partial}{\partial b_{i}}L^{\text{VAE}}(\theta,\phi;\bx)=\int_\bz\frac{\partial}{\partial b_{i}} \log \left( p(\bx|\bz;\theta)\right) \cdot q(\bz|\bx;\phi)d\bz,
\eean
}
where $w_{ij}$ and $b_i$ for $i\in [D]$ and $j\in [d]$ are the $(i,j)$ element of $W$ and the $i$-th element of $b$, respectively. 
Here, we define $[L]:=\{1,\ldots,L\}$ for $L\in\mathbb{N}$.
Note that 
{\small
\bean
\mathbb{E}_{\theta,\phi}\left\|\frac{\partial}{\partial \theta}L^{\text{VAE}}(\theta,\phi;\bx) \right\|_2^2&=&\sum_i\sum_j \mathbb{E}_{\theta,\phi}\left[\frac{\partial}{\partial w_{ij}}L^{\text{VAE}}(\theta,\phi;\bx)\right]^2
+\sum_i \mathbb{E}_{\theta,\phi}\left[\frac{\partial}{\partial b_{i}}L^{\text{VAE}}(\theta,\phi;\bx)\right]^2.
\eean
}
We are going to characterize the two terms,   $\mathbb{E}_{\theta,\phi}\left[\frac{\partial}{\partial w_{ij}}L^{\text{VAE}}(\theta,\phi;\bx)\right]^2$ and $\mathbb{E}_{\theta,\phi}\left[\frac{\partial}{\partial b_{i}}L^{\text{VAE}}(\theta,\phi;\bx)\right]^2$, 
and combine them to make the final conclusion. 

\textbf{w.r.t. $w_{ij}$}

Since $X|(Z=\bz)\sim \mathcal{N}\left( W\bz+b, \sigma^2\right)$ holds, we have
\bean
\log  p(\bx|\bz;\theta)
=-\frac{1}{2\sigma^2}\sum_{i=1}^D\left( x_i-\bw_i'\bz-b_i\right)^2+const
=-\frac{1}{2\sigma^2}\sum_{i=1}^D\left(x_i-\sum_{j=1}^d w_{ij}z_j-b_i\right)^2+const,
\eean
where $\bw_i$ is the $i$-th row of $W$ and $const$ is a constant not depending on $\theta$. 
Therefore, we can obtain the following result of the log-likelihood with respect to $w_{ij}$:
{\normalsize
\bean
&&\frac{\partial}{\partial w_{ij}} \log \left( p(\bx|\bz;\theta)\right)
=\frac{1}{\sigma^2}\left(x_i-\bw_i'\bz-b_i\right)\cdot z_j
=\frac{1}{\sigma^2}\left[x_iz_j-w_{ij}z_j^2-\sum_{j'\neq j}w_{ij'}z_jz_{j'}-b_iz_j \right].
\eean
}
Hence, the first derivative of the VAE objective function with respect to $w_{ij}$
becomes:
{\footnotesize
$$
\frac{\partial}{\partial w_{ij}}L^{\text{VAE}}(\theta,\phi;\bx) =\int_\bz \frac{1}{\sigma^2}\left[(x_i-b_i)z_j-w_{ij}z_j^2-\sum_{j'\neq j}w_{ij'}z_jz_{j'}-b_iz_j \right]\cdot q(\bz|\bx;\phi)d\bz
$$
\bean
&&=\frac{1}{\sigma^2}\left[ (x_i-b_i) (\bu_j'\bx+v_j)-w_{ij}\left(\left(\bu_j'\bx+v_j\right)^2+\eta^2\right)
-\left(\bu_j'\bx+v_j\right)\sum_{j'\neq j}w_{ij'}\left(\bu_{j'}'\bx+v_{j'}\right)\right]\\
&&=\frac{1}{\sigma^2}\left[ (x_i-b_i) (\bu_j'\bx+v_j)
- \left(\bu_j'\bx+v_j\right)\sum_{j'}w_{ij'}\left(\bu_{j'}'\bx+v_{j'}\right)-w_{ij}\eta^2 \right],
\eean
}
where $\bu_j$ is the $j$-th row of $U$. 
By squaring the above term, we have
{
\bean
\left[\frac{\partial}{\partial w_{ij}}L^{\text{VAE}}(\theta,\phi;\bx)\right]^2
=&&\frac{1}{\sigma^4}\left[{(x_i-b_i)^2\left(\bu_j'\bx+v_j\right)^2}
+\left(\bu_j'\bx+v_j\right)^2{\sum_{j'}w_{ij'}^2\left(\bu_{j'}'\bx+v_{j'}\right)^2}
+w_{ij}^4\eta^4\right.\\
&&+2\left(\bu_j'\bx+v_j\right)^2\sum_{j^{''}>j'}w_{ij'}w_{ij''}(\bu_{j'}'\bx+v_{j'})(\bu_{j''}'\bx+v_{j''})\\
&&+2w_{ij}\eta^2\left(\bu_j'\bx+v_j\right)\sum_{j'}w_{ij'}
\left(\bu_{j'}'\bx+v_{j'}\right)
-2w_{ij}\eta^2(x_i-b_i)\left(\bu_j'\bx+v_j\right)\\
&&\left.-2(x_i-b_i)\left(\bu_j'\bx+v_j\right)^2\sum_{j'}w_{ij'}
\left(\bu_{j'}'\bx+v_{j'}\right)\right].
\eean
}
Now, we will calculate the expected value of the above equation with respect to $\theta$ and $\phi$. 
To do this, we will take the expectation for each term in the RHS of the above equation. 
Note that, for a random variable $X\sim Unif[-1,1]$, $\mathbb{E}\left[X\right]=0$, $\mathbb{E}\left[X^2\right]=1/3$, and $\mathbb{E}\left[X^4\right]=1/5$. 
Thus, we have
\bean
&&\mathbb{E}_{\theta,\phi}\left[(x_i-b_i)^2\left(\bu_j'\bx+v_j\right)^2\right]
=\mathbb{E}_{\theta,\phi}\left[(x_i^2-2x_ib_i+b_i^2)\left((\bu_j'\bx)^2+v_j^2+2v_j\bu_j'\bx\right)\right]\\
&&=\mathbb{E}_{\theta,\phi}\left[(x_i^2+b_i^2)\left(\sum_{i'}u_{ji'}^2x_{i'}^2+v_j^2\right)\right]
=\left(x_i^2+\frac{1}{3}\right)\cdot\left(\frac{1}{3}\left\|\bx\right\|_2^2+\frac{1}{3}\right),
\eean
{\small
\bean
&&\mathbb{E}_{\theta,\phi}\left[\left(\bu_j'\bx+v_j\right)^2{\sum_{j'}w_{ij'}^2\left(\bu_{j'}'\bx+v_{j'}\right)^2}\right]
=\mathbb{E}_{\theta,\phi}\left[\left(\sum_{i'}u_{ji'}x_{i'}+v_j\right)^2{\sum_{j'}w_{ij'}^2\left(\sum_{i'}u_{j'i'}x_{i'}+v_{j'}\right)^2}\right]\\
&&=\frac{1}{3}\mathbb{E}_{\theta,\phi}\left[ \left(\sum_{i'}u_{ji'}x_{i'}+v_j\right)^4
+ \left(\sum_{i'}u_{ji'}x_{i'}+v_j\right)^2\sum_{j'\neq j}\left(\sum_{i'}u_{j'i'}x_{i'}+v_{j'}\right)^2 \right]\\
&&=\frac{1}{3}\mathbb{E}_{\theta,\phi}\left[ \left(\sum_{i'}u_{ji'}x_{i'} \right)^4+6\left(\sum_{i'}u_{ji'}x_{i'}\right)^2v_j^2+v_j^4
+\left(\sum_{i'}u_{ji'}x_{i'}+v_j\right)^2\sum_{j'\neq j}\left(\sum_{i'}u_{j'i'}x_{i'}+v_{j'}\right)^2 \right]\\
&&=\frac{1}{3}\mathbb{E}_{\theta,\phi}\left[ \sum_{i'}u_{ji'}^4x_{i'}^4+6\sum_{i''>i'}u_{ji'}^2u_{ji''}x_{i'}^2x_{i''}^2
+6\left(\sum_{i'}u_{ji'}^2x_{i'}^2\right)v_j^2+v_j^4
+\left(\sum_{i'}u_{ji'}^2x_{i'}^2+v_j^2\right)
\sum_{j'\neq j}\left(\sum_{i'}u_{j'i'}^2x_{i'}^2+v_{j'}^2\right)\right]\\
&&=\frac{1}{3}\left[ \frac{1}{5}\sum_{i'}x_i^4+\frac{2}{3}\sum_{i''>i'}x_{i'}^2x_{i''}^2+\frac{2}{3}\sum_{i'}x_{i'}^2+\frac{1}{5}
+ \left(\frac{1}{3}\sum_{i'}x_{i'}^2\right)\sum_{j'\neq j}\left(\frac{1}{3}\sum_{i'}x_{i'}^2\right) \right]\\
&&=\frac{1}{3}\left[ \frac{1}{3}\|\bx\|_2^4-\frac{2}{15}\|\bx\|_4^4+\frac{2}{3}\|\bx\|_2^2+\frac{1}{5}+(d-1)\left(\frac{1}{3}\|\bx\|_2^2+\frac{1}{3}\right) \right],
\eean
}
\bean
\mathbb{E}_{\theta,\phi}w_{ij}^2\eta^4=\eta^4\mathbb{E}_{\theta}w_{ij}^2=\frac{1}{3}\eta^4,
\eean
{\small
\bean
&&\mathbb{E}_{\theta,\phi}\left[2\left(\bu_j'\bx+v_j\right)^2\sum_{j^{''}>j'}w_{ij'}w_{ij''}(\bu_{j'}'\bx+v_{j'})(\bu_{j''}'\bx+v_{j''})\right]
=0,
\eean
}
\bean
&&\mathbb{E}_{\theta,\phi}\left[2w_{ij}\eta^2\sum_{j'}w_{ij'}
\left(\bu_j'\bx+v_j\right)\left(\bu_{j'}'\bx+v_{j'}\right)\right]
=2\mathbb{E}_{\theta,\phi}\left[ w_{ij}^2 \eta^2\left(\bu_j'\bx+v_j\right)^2 \right]
=\frac{2}{3}\eta^2\mathbb{E}_{\phi}\left[ \left( \sum_{i'}u_{ji'}x_{i'}+v_j \right)^2\right]\\
&=&\frac{2}{3}\eta^2\mathbb{E}_{\phi}\left[ \sum_{i'}u_{ji'}^2x_{i'}^2+v_j^2\right]
=\frac{2}{9}\eta^2\left(\left\|\bx\right\|_2^2+1\right),
\eean
\bean
\mathbb{E}_{\theta,\phi}\left[ -2w_{ij}\eta^2(x_i-b_i)\left(\bu_j'\bx+v_j\right)\right]=0,
\eean
and
\bean
\mathbb{E}_{\theta,\phi}\left[-2(x_i-b_i)\left(\bu_j'\bx+v_j\right)^2\sum_{j'}w_{ij'}
\left(\bu_{j'}'\bx+v_{j'}\right)\right]=0.
\eean
By integrating all the above expected values and using the property $\|\bx\|_2\ge\|\bx\|_4$, we arrive at the following result:
\bean
\mathbb{E}_{\theta,\phi}\left[\frac{\partial}{\partial w_{ij}}L^{\text{VAE}}(\theta,\phi;\bx)\right]^2=\Theta\left(\|\bx\|_2^4\right).
\eean


\textbf{w.r.t. $b_{i}$}

We have 
\bean
\frac{\partial}{\partial b_i}\log p(\bx|\bz;\theta)=\frac{1}{\sigma^2}\left[x_i-\sum_{j=1}^d w_{ij}z_j-b_i\right],
\eean
thus, 
\bean
&&\frac{\partial}{\partial b_i}L^{\text{VAE}}(\theta,\phi;\bx)
=\int_\bz \frac{1}{\sigma^2}\left[x_i-\sum_{j=1}^d w_{ij}z_j-b_i\right]\cdot q(\bz|\bx;\phi)d\bz
=\frac{1}{\sigma^2}\left[ (x_i-b_i)-\sum_j w_{ij}(\bu_j'\bx+v_j) \right].
\eean
By squaring the above term, 
\bean
&&\left[\frac{\partial}{\partial b_i}L^{\text{VAE}}(\theta,\phi;\bx)\right]^2\\
&&=\frac{1}{\sigma^4}\left[ (x_i-b_i)^2+\sum_{j}w_{ij}^2(\bu_j'\bx+v_j)^2 \right.
+2\sum_{j'>j}w_{ij}w_{ij'}(\bu_j'\bx+v_j)(\bu_{j'}'\bx+v_{j'})
-\left. 2(x_i-b_i)\sum_j w_{ij}(\bu_j'\bx+v_j) \right].
\eean
Here, we calculate the expected value of each term in the above RHS. We have 
\bean
\mathbb{E}_{\theta,\phi}\left[ (x_i-b_i)^2 \right]=x_i^2+\frac{1}{3},
\eean
\bean
\mathbb{E}_{\theta,\phi}\left[ \sum_{j}w_{ij}^2(\bu_j'\bx+v_j)^2 \right]&=&\frac{1}{3}\sum_{j}\mathbb{E}_{\theta,\phi}\left[ \sum_{i'} u_{ji'}^2x_{i'}^2+v_j^2 \right]
=\frac{d}{9}\|\bx\|_2^2+\frac{d}{9},
\eean
\bean
\mathbb{E}_{\theta,\phi}\left[ 2\sum_{j'>j}w_{ij}w_{ij'}(\bu_j'\bx+v_j)(\bu_{j'}'\bx+v_{j'}) \right]=0,
\eean
and 
\bean
\mathbb{E}_{\theta,\phi}\left[ 2(x_i-b_i)\sum_j w_{ij}(\bu_j'\bx+v_j) \right]=0.
\eean 
Combining the above expectations, we have
\bean
\mathbb{E}_{\theta,\phi}\left[\frac{\partial}{\partial b_i}L^{\text{VAE}}(\theta,\phi;\bx)\right]^2=\Theta\left(\|\bx\|_2^2\right).
\eean

\paragraph{Final conclusion}
Combining the above results, we have 
{
\bean
&&\mathbb{E}_{\theta,\phi}\left\|\frac{\partial}{\partial \theta}L^{\text{VAE}}(\theta,\phi;\bx) \right\|_2^2
=\sum_i\sum_j \mathbb{E}_{\theta,\phi}\left[\frac{\partial}{\partial w_{ij}}L^{\text{VAE}}(\theta,\phi;\bx)\right]^2
+\sum_i \mathbb{E}_{\theta,\phi}\left[\frac{\partial}{\partial b_{i}}L^{\text{VAE}}(\theta,\phi;\bx)\right]^2\\
&&=\sum_{i}\sum_j \Theta\left(\|\bx\|_2^4\right)+\sum_i \Theta\left(\|\bx\|_2^2\right)
=\Theta\left(\|\bx\|_2^4\right).
\eean
Thus, the proof is completed by using the inequality  
 $D^{-1/2}\|\bx\|_1\le\|\bx\|_2\le\|\bx\|_1$
 \text{\qed}}

\subsection*{A.2. Proof of Proposition 2}
Due to the second condition of the Proposition 2, there exist four real numbers $-\infty<a_j<b_j<c_j<d_j<\infty$ and a small positive number $\epsilon>0$ such that  $X_{j}^{in}\in [b_j,c_j]$ and $X_j^{out}\in [a_j,b_j-\epsilon]\cup[c_j+\epsilon,d_j]$ for all $j\in[D]$. 
Suppose that $\inf X_{j}^{in} = a_j$ and $\sup X_{j}^{in}=d_j$ for $j\in[D]$.  
And we define $X_{j,+}^{in}=\max\{0,X_j^{in}\}$ and $X_{j,-}^{in}=\max\{0,-X_j^{in}\}$ so that $X_{j}^{in}=X_{j,+}^{in}-X_{j,-}^{in}$. 
Similarly we define $X_{j,+}^{out}$ and $X_{j,-}^{out}$.

\paragraph{About min-max}
Since $\mathbb{E}X_j^{in}=\mathbb{E}X_j^{out}=0$ for $j\in[D]$, it is trivial to show that $\mathbb{E}|X_{mm,j}^{in}|=\mathbb{E}|X_{mm,j}^{out}|=-a_j/(d_j-a_j)$. 

\paragraph{About standardization}
It is sufficient to show that $\mathbb{E}|X_{j}^{in}|<\mathbb{E}|X_{j}^{out}|$ since $\mathbb{E}X_j^{in}=\mathbb{E}X_j^{out}=0$. 
Note that $0\le X_{j,+}^{in}\ge c_j<c_j+\epsilon\ge X_{j,+}^{in}<d_j$, so we can have $\mathbb{E}X_{j,+}^{in}<\mathbb{E}X_{j,+}^{out}$. 
In a similar manner, we can also derive $\mathbb{E}X_{j,-}^{in}<\mathbb{E}X_{j,-}^{out}$. 
By considering that $|X_{j}^{in}|=X_{j,-}^{in}+X_{j,+}^{in}$ and $|X_{j}^{out}|=X_{j,-}^{out}+X_{j,+}^{out}$, the proof is completed.
\qed

\clearpage

\clearpage



\section*{B. Why do we choose the IWAE method as a learning framework?}


In training DGMs with likelihood regimes, two primary approaches are commonly used: 1) employing the lower bound of the log-likelihood, often called ELBO,  \citep{kingma2013auto,DBLP:journals/corr/BurdaGS15,tomczak2018vae,kim2020casting}, and 2) normalizing flows which calculate the exact log-likelihood \citep{DBLP:journals/corr/DinhKB14,DBLP:conf/iclr/DinhSB17,DBLP:conf/nips/KingmaD18}.

\begin{figure*}[ht]
\renewcommand\thefigure{B.1}
\begin{center}
\centerline{
\includegraphics[width=0.2\linewidth]{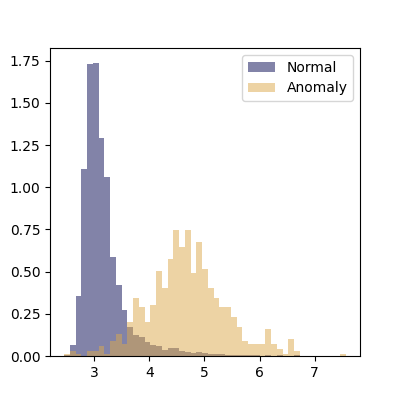}
\includegraphics[width=0.2\linewidth]{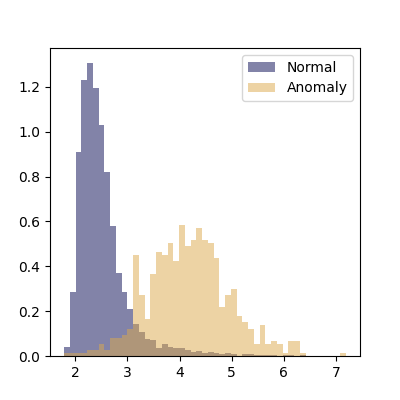}
\includegraphics[width=0.2\linewidth]{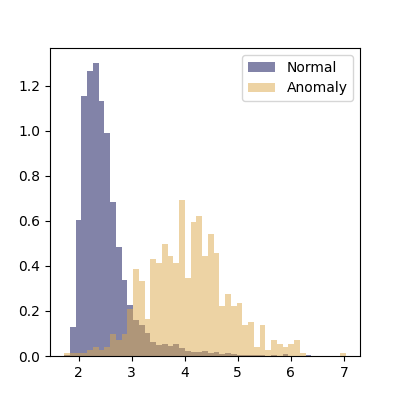}
\includegraphics[width=0.2\linewidth]{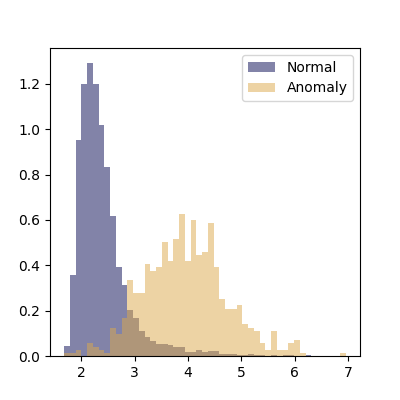}
}
\caption{
({\bf Left to Right}) The distributions of the per-sample (normalized) negative log-likelihood values of GLOW on {\texttt{FMNIST}} after 10, 20, 30, and 40 updates, respectively.
For each panel, we depict the histograms of inliers and outliers separately. 
}
\label{fig:glow}
\end{center}
\end{figure*}

Two things need to be considered when selecting a DGM framework for exploiting the IM effect: 1) the clarity of the IM effect during learning and 2) computational efficiency.
Among the normalizing flows methods, we considered the GLOW \citep{DBLP:conf/nips/KingmaD18}. 
We trained GLOW on the \texttt{FMNIST} dataset and monitored the per-sample loss values, i.e., negative log-likelihood, of inliers and outliers during the early learning phases. 
The results, illustrated in Figure \ref{fig:glow}, shows that the IM effect clearly appears during GLOW training.

Although the IM effect occurs when using GLOW, we ultimately chose the IWAE method, one of the ELBO methods, because the performance of ODIM using GLOW was insufficient compared to IWAE. 
Additionally, ELBO-based methods are relatively flexible in choosing the encoder and decoder, while normalizing flows are somewhat limited in this regard. 
And the advantage of IWAE in comparison with VAE is described in Section \ref{sec:ablation_study} of the main manuscript.

In fact, there is another line of work, called auto-regressive models \citep{DBLP:conf/icml/OordKK16,DBLP:conf/iclr/SalimansK0K17}, that utilizes an exact log-likelihood. 
We note that we excluded auto-regressive models from the exact likelihood approach since they are generally very computationally burdensome.

\clearpage
\section*{C. Numerical experiments}

\subsection*{C.1. Data description}


We examine a total of 46 tabular datasets, 6 image datasets, and 5 sequential datasets. 
All of these datasets are taken from a source called \texttt{ADBench} \citep{han2022adbench}. 
The basic information of all datasets we analyze is summarized in Table \ref{tab:tabular_description}.

\begin{table*}[h]
\renewcommand\thetable{C.1}
\caption{Description of \texttt{ADBench} datasets}
\scriptsize
\centering
    \resizebox{0.62\textwidth}{!}{
    \begin{tabular}{llccccc}
    \toprule
    \textbf{Number} & \textbf{Data} & \textbf{\# Samples} & \textbf{\# Features} & \textbf{\# Anomaly} & \textbf{\% Anomaly}  \\ \midrule
    \textbf{1} & \texttt{ALOI} & 49534 & 27 & 1508 & 3.04  \\
    \textbf{2} & \texttt{Annthyroid} & 7200 & 6 & 534 & 7.42  \\
    \textbf{3} & \texttt{Backdoor} & 95329 & 196 & 2329 & 2.44  \\
    \textbf{4} & \texttt{Breastw} & 683 & 9 & 239 & 34.99  \\
    \textbf{5} & \texttt{Campaign} & 41188 & 62 & 4640 & 11.27  \\
    \textbf{6} & \texttt{Cardio} & 1831 & 21 & 176 & 9.61  \\
    \textbf{7} & \texttt{Cardiotocography} & 2114 & 21 & 466 & 22.04  \\
    \textbf{8} & \texttt{Celeba} & 202599 & 39 & 4547 & 2.24  \\
    \textbf{9} & \texttt{Census} & 299285 & 500 & 18568 & 6.20  \\
    \textbf{10} & \texttt{Cover} & 286048 & 10 & 2747 & 0.96  \\
    \textbf{11} & \texttt{Donors} & 619326 & 10 & 36710 & 5.93  \\
    \textbf{12} & \texttt{Fault} & 1941 & 27 & 673 & 34.67  \\
    \textbf{13} & \texttt{Fraud} & 284807 & 29 & 492 & 0.17  \\
    \textbf{14} & \texttt{Glass} & 214 & 7 & 9 & 4.21  \\
    \textbf{15} & \texttt{Hepatitis} & 80 & 19 & 13 & 16.25  \\
    \textbf{16} & \texttt{Http} & 567498 & 3 & 2211 & 0.39  \\
    \textbf{17} & \texttt{InternetAds} & 1966 & 1555 & 368 & 18.72  \\
    \textbf{18} & \texttt{Ionosphere} & 351 & 32 & 126 & 35.90  \\
    \textbf{19} & \texttt{Landsat} & 6435 & 36 & 1333 & 20.71  \\
    \textbf{20} & \texttt{Letter} & 1600 & 32 & 100 & 6.25  \\
    \textbf{21} & \texttt{Lymphography} & 148 & 18 & 6 & 4.05  \\
    \textbf{22} & \texttt{Magic.gamma} & 19020 & 10 & 6688 & 35.16  \\
    \textbf{23} & \texttt{Mammography} & 11183 & 6 & 260 & 2.32  \\
    \textbf{24} & \texttt{MNIST} & 7603 & 100 & 700 & 9.21  \\
    \textbf{25} & \texttt{Musk} & 3062 & 166 & 97 & 3.17  \\
    \textbf{26} & \texttt{Optdigits} & 5216 & 64 & 150 & 2.88  \\
    \textbf{27} & \texttt{PageBlocks} & 5393 & 10 & 510 & 9.46  \\
    \textbf{28} & \texttt{Pendigits} & 6870 & 16 & 156 & 2.27  \\
    \textbf{29} & \texttt{Pima} & 768 & 8 & 268 & 34.90  \\
    \textbf{30} & \texttt{Satellite} & 6435 & 36 & 2036 & 31.64  \\
    \textbf{31} & \texttt{Satimage-2} & 5803 & 36 & 71 & 1.22  \\
    \textbf{32} & \texttt{Shuttle} & 49097 & 9 & 3511 & 7.15  \\
    \textbf{33} & \texttt{Skin} & 245057 & 3 & 50859 & 20.75  \\
    \textbf{34} & \texttt{Smtp} & 95156 & 3 & 30 & 0.03  \\
    \textbf{35} & \texttt{SpamBase} & 4207 & 57 & 1679 & 39.91  \\
    \textbf{36} & \texttt{Speech} & 3686 & 400 & 61 & 1.65  \\
    \textbf{37} & \texttt{Stamps} & 340 & 9 & 31 & 9.12  \\
    \textbf{38} & \texttt{Thyroid} & 3772 & 6 & 93 & 2.47  \\
    \textbf{39} & \texttt{Vertebral} & 240 & 6 & 30 & 12.50  \\
    \textbf{40} & \texttt{Vowels} & 1456 & 12 & 50 & 3.43  \\
    \textbf{41} & \texttt{Waveform} & 3443 & 21 & 100 & 2.90  \\
    \textbf{42} & \texttt{WBC} & 223 & 9 & 10 & 4.48  \\
    \textbf{43} & \texttt{WDBC} & 367 & 30 & 10 & 2.72  \\
    \textbf{44} & \texttt{Wilt} & 4819 & 5 & 257 & 5.33  \\
    \textbf{45} & \texttt{Wine} & 129 & 13 & 10 & 7.75  \\
    \textbf{46} & \texttt{WPBC} & 198 & 33 & 47 & 23.74  \\
    \textbf{47} & \texttt{Yeast} & 1484 & 8 & 507 & 34.16  \\
    \textbf{48} & \texttt{CIFAR10} & 5263 & 512 & 263 & 5.00  \\
    \textbf{49} & \texttt{FMNIST} & 6315 & 512 & 315 & 5.00  \\
    \textbf{50} & \texttt{MNIST-C} & 10000 & 512 & 500 & 5.00  \\
    \textbf{51} & \texttt{MVTec-AD} & 5354 & 512 & 1258 & 23.5  \\
    \textbf{52} & \texttt{SVHN} & 5208 & 512 & 260 & 5.00  \\
    \textbf{53} & \texttt{Agnews} & 10000 & 768 & 500 & 5.00  \\
    \textbf{54} & \texttt{Amazon} & 10000 & 768 & 500 & 5.00  \\
    \textbf{55} & \texttt{Imdb} & 10000 & 768 & 500 & 5.00  \\
    \textbf{56} & \texttt{Yelp} & 10000 & 768 & 500 & 5.00  \\
    \textbf{57} & \texttt{20news} & 10000 & 768 & 500 & 5.00  \\
    \bottomrule
    \end{tabular}
    }
    \label{tab:tabular_description}
\end{table*}

\clearpage

\subsection*{C.2. Detailed AUC and PR results over tabular datasets (training data)}

Tables \ref{tab:tabular_auc_result-1}-\ref{tab:tabular_auc_result} and \ref{tab:tabular_ap_result-1}-\ref{tab:tabular_ap_result} provide a comparison of inlier identification performance for each method and dataset in terms of AUC and PR.

\begin{table}[h!]
\renewcommand\thetable{C.2.1}
\scriptsize
\caption{Training AUC value comparisons on tabular datasets}
    \centering
    \resizebox{0.99\textwidth}{!}{
    \begin{tabular}{l|c|c|c|c|c|c|c|c|c|c|c|c|c}
    \hline
        \textbf{Data} & \textbf{CBLOF} & \textbf{FeatureBagging} & \textbf{HBOS} & \textbf{IForest} & \textbf{kNN} & \textbf{LODA} & \textbf{LOF} & \textbf{MCD} & \textbf{PCA} & \textbf{DAGMM} & \textbf{DROCC} & \textbf{GOAD} & \textbf{PlanarFlow} \\ 
        \hline
        \texttt{Aloi} & 0.556 & 0.792 & 0.531 & 0.542 & 0.613 & 0.495 & 0.767 & 0.520 & 0.549 & 0.517 & 0.500 & 0.497 & 0.520 \\ 
        \texttt{Annthyroid} & 0.676 & 0.788 & 0.608 & 0.816 & 0.761 & 0.453 & 0.710 & 0.918 & 0.676 & 0.548 & 0.631 & 0.453 & 0.966 \\ 
        \texttt{Backdoor} & 0.897 & 0.790 & 0.740 & 0.725 & 0.826 & 0.515 & 0.764 & 0.848 & 0.888 & 0.752 & 0.500 & 0.587 & 0.787 \\ 
        \texttt{Breastw} & 0.961 & 0.408 & 0.984 & 0.983 & 0.980 & 0.970 & 0.446 & 0.985 & 0.946 & 0.811 & 0.847 & 0.845 & 0.965 \\ 
        \texttt{Campaign} & 0.738 & 0.594 & 0.768 & 0.704 & 0.750 & 0.493 & 0.614 & 0.775 & 0.734 & 0.580 & 0.500 & 0.443 & 0.566 \\ 
        \texttt{Cardio} & 0.832 & 0.579 & 0.839 & 0.922 & 0.830 & 0.856 & 0.551 & 0.815 & 0.949 & 0.625 & 0.655 & 0.908 & 0.796 \\ 
        \texttt{Cardiotocography} & 0.561 & 0.538 & 0.595 & 0.681 & 0.503 & 0.708 & 0.527 & 0.500 & 0.747 & 0.546 & 0.449 & 0.624 & 0.643 \\ 
        \texttt{Celeba} & 0.753 & 0.514 & 0.754 & 0.707 & 0.736 & 0.600 & 0.432 & 0.803 & 0.792 & 0.627 & 0.726 & 0.432 & 0.703 \\ 
        \texttt{Census} & 0.664 & 0.538 & 0.611 & 0.607 & 0.671 & 0.454 & 0.562 & 0.731 & 0.662 & 0.491 & 0.443 & 0.488 & 0.604 \\ 
        \texttt{Cover} & 0.922 & 0.571 & 0.707 & 0.873 & 0.866 & 0.922 & 0.568 & 0.696 & 0.934 & 0.742 & 0.747 & 0.124 & 0.417 \\ 
        \texttt{Donors} & 0.808 & 0.691 & 0.743 & 0.771 & 0.829 & 0.566 & 0.629 & 0.765 & 0.825 & 0.558 & 0.747 & 0.225 & 0.899 \\ 
        \texttt{Fault} & 0.665 & 0.591 & 0.506 & 0.544 & 0.715 & 0.478 & 0.579 & 0.505 & 0.480 & 0.495 & 0.668 & 0.546 & 0.469 \\ 
        \texttt{Fraud} & 0.954 & 0.616 & 0.945 & 0.950 & 0.955 & 0.856 & 0.548 & 0.911 & 0.952 & 0.857 & 0.500 & 0.724 & 0.895 \\ 
        \texttt{Glass} & 0.855 & 0.659 & 0.820 & 0.790 & 0.870 & 0.624 & 0.618 & 0.795 & 0.715 & 0.630 & 0.743 & 0.545 & 0.766 \\ 
        \texttt{Hepatitis} & 0.635 & 0.469 & 0.768 & 0.683 & 0.669 & 0.557 & 0.468 & 0.721 & 0.748 & 0.600 & 0.582 & 0.637 & 0.654 \\ 
        \texttt{Http} & 0.996 & 0.288 & 0.991 & 0.999 & 0.051 & 0.060 & 0.338 & 0.999 & 0.997 & 0.838 & 0.500 & 0.996 & 0.994 \\ 
        \texttt{Internetads} & 0.616 & 0.494 & 0.696 & 0.686 & 0.616 & 0.541 & 0.587 & 0.660 & 0.609 & 0.515 & 0.500 & 0.614 & 0.608 \\ 
        \texttt{Ionosphere} & 0.892 & 0.876 & 0.544 & 0.833 & 0.922 & 0.788 & 0.864 & 0.951 & 0.777 & 0.641 & 0.766 & 0.829 & 0.884 \\ 
        \texttt{Landsat} & 0.548 & 0.540 & 0.575 & 0.474 & 0.614 & 0.382 & 0.549 & 0.607 & 0.366 & 0.533 & 0.626 & 0.506 & 0.464 \\ 
        \texttt{Letter} & 0.763 & 0.886 & 0.589 & 0.616 & 0.812 & 0.537 & 0.878 & 0.804 & 0.524 & 0.503 & 0.780 & 0.598 & 0.689 \\ 
        \texttt{Lymphography} & 0.994 & 0.523 & 0.995 & 0.999 & 0.995 & 0.900 & 0.636 & 0.989 & 0.997 & 0.840 & 0.878 & 0.995 & 0.940 \\ 
        \texttt{Magic.gamma} & 0.725 & 0.700 & 0.709 & 0.721 & 0.795 & 0.655 & 0.678 & 0.699 & 0.667 & 0.584 & 0.728 & 0.442 & 0.742 \\ 
        \texttt{Mammography} & 0.795 & 0.726 & 0.838 & 0.860 & 0.852 & 0.867 & 0.702 & 0.690 & 0.888 & 0.719 & 0.779 & 0.414 & 0.782 \\ 
        \texttt{Musk} & 1.000 & 0.575 & 1.000 & 0.998 & 0.964 & 0.993 & 0.581 & 0.999 & 1.000 & 0.912 & 0.575 & 1.000 & 0.748 \\ 
        \texttt{Optdigits} & 0.785 & 0.539 & 0.868 & 0.696 & 0.395 & 0.493 & 0.538 & 0.413 & 0.518 & 0.408 & 0.565 & 0.657 & 0.492 \\ 
        \texttt{Pageblocks} & 0.893 & 0.758 & 0.779 & 0.897 & 0.919 & 0.712 & 0.703 & 0.923 & 0.907 & 0.753 & 0.914 & 0.609 & 0.908 \\ 
        \texttt{Pendigits} & 0.864 & 0.518 & 0.925 & 0.947 & 0.828 & 0.895 & 0.534 & 0.834 & 0.936 & 0.548 & 0.520 & 0.592 & 0.780 \\ 
        \texttt{Pima} & 0.655 & 0.573 & 0.704 & 0.674 & 0.723 & 0.595 & 0.563 & 0.686 & 0.651 & 0.522 & 0.542 & 0.606 & 0.615 \\ 
        \texttt{Satellite} & 0.742 & 0.545 & 0.762 & 0.695 & 0.721 & 0.614 & 0.550 & 0.804 & 0.601 & 0.675 & 0.608 & 0.702 & 0.671 \\ 
        \texttt{Satimage-2} & 0.999 & 0.526 & 0.976 & 0.993 & 0.992 & 0.981 & 0.539 & 0.995 & 0.977 & 0.911 & 0.579 & 0.996 & 0.970 \\ 
        \texttt{Shuttle} & 0.621 & 0.493 & 0.986 & 0.997 & 0.732 & 0.389 & 0.526 & 0.990 & 0.990 & 0.898 & 0.500 & 0.208 & 0.852 \\ 
        \texttt{Skin} & 0.675 & 0.534 & 0.588 & 0.670 & 0.720 & 0.442 & 0.550 & 0.892 & 0.447 & 0.554 & 0.708 & 0.579 & 0.773 \\ 
        \texttt{Smtp} & 0.863 & 0.794 & 0.809 & 0.905 & 0.933 & 0.819 & 0.899 & 0.948 & 0.856 & 0.868 & 0.500 & 0.915 & 0.784 \\ 
        \texttt{Spambase} & 0.541 & 0.424 & 0.664 & 0.637 & 0.566 & 0.480 & 0.453 & 0.446 & 0.548 & 0.488 & 0.490 & 0.496 & 0.528 \\ 
        \texttt{Speech} & 0.471 & 0.509 & 0.473 & 0.476 & 0.480 & 0.466 & 0.512 & 0.494 & 0.469 & 0.522 & 0.483 & 0.458 & 0.496 \\ 
        \texttt{Stamps} & 0.660 & 0.502 & 0.904 & 0.907 & 0.870 & 0.831 & 0.512 & 0.838 & 0.909 & 0.719 & 0.760 & 0.774 & 0.838 \\ 
        \texttt{Thyroid} & 0.909 & 0.707 & 0.948 & 0.979 & 0.965 & 0.819 & 0.657 & 0.986 & 0.955 & 0.719 & 0.889 & 0.574 & 0.992 \\ 
        \texttt{Vertebral} & 0.463 & 0.473 & 0.317 & 0.362 & 0.379 & 0.294 & 0.487 & 0.389 & 0.378 & 0.470 & 0.425 & 0.468 & 0.409 \\ 
        \texttt{Vowels} & 0.884 & 0.933 & 0.679 & 0.763 & 0.951 & 0.705 & 0.932 & 0.732 & 0.604 & 0.464 & 0.738 & 0.791 & 0.888 \\ 
        \texttt{Waveform} & 0.701 & 0.715 & 0.694 & 0.707 & 0.750 & 0.594 & 0.693 & 0.572 & 0.635 & 0.523 & 0.674 & 0.592 & 0.640 \\ 
        \texttt{Wbc} & 0.977 & 0.388 & 0.987 & 0.996 & 0.982 & 0.992 & 0.607 & 0.988 & 0.993 & 0.821 & 0.821 & 0.949 & 0.934 \\ 
        \texttt{Wdbc} & 0.990 & 0.867 & 0.989 & 0.988 & 0.980 & 0.980 & 0.849 & 0.969 & 0.988 & 0.715 & 0.347 & 0.983 & 0.985 \\ 
        \texttt{Wilt} & 0.396 & 0.666 & 0.348 & 0.451 & 0.511 & 0.313 & 0.678 & 0.859 & 0.239 & 0.432 & 0.400 & 0.555 & 0.794 \\         
        \texttt{Wine} & 0.453 & 0.323 & 0.907 & 0.786 & 0.470 & 0.822 & 0.330 & 0.975 & 0.819 & 0.513 & 0.621 & 0.734 & 0.390 \\ 
        \texttt{Wpbc} & 0.487 & 0.436 & 0.548 & 0.516 & 0.512 & 0.501 & 0.447 & 0.534 & 0.486 & 0.449 & 0.483 & 0.467 & 0.483 \\ 
        \texttt{Yeast} & 0.461 & 0.465 & 0.402 & 0.394 & 0.396 & 0.461 & 0.453 & 0.406 & 0.418 & 0.503 & 0.396 & 0.503 & 0.442 \\ 
        \hline
        \textbf{Average} & 0.746 & 0.596 & 0.742 & 0.759 & 0.738 & 0.641 & 0.600 & 0.769 & 0.734 & 0.629 & 0.616 & 0.623 & 0.721 \\
        \hline
    \end{tabular}
    }
    \label{tab:tabular_auc_result-1}
\end{table}

\begin{table}[h!]
\renewcommand\thetable{C.2.2}
\scriptsize
\caption{Training AUC value comparisons on tabular datasets}
    \centering
    \resizebox{0.8\textwidth}{!}{
    \begin{tabular}{l|c|c|c|c|c|c|c|c}
    \hline
        \textbf{Data} &  \textbf{OCSVM} & \textbf{COPOD} & \textbf{ECOD} & \textbf{DeepSVDD} & \textbf{ICL} & \textbf{DDPM} & \textbf{DTE} & \textbf{ODIM} \\ 
        \hline
        \texttt{ALOI}  & 0.549  & 0.515  & 0.531  & 0.514  & 0.548  & 0.532  & 0.645  & 0.531  \\ 
        \texttt{Annthyroid} & 0.682  & 0.777  & 0.789  & 0.739  & 0.599  & 0.814  & 0.781  & 0.588  \\ 
        \texttt{Backdoor}  & 0.889  & 0.500  & 0.500  & 0.735  & 0.936  & 0.892 & 0.806 & 0.885  \\ 
        \texttt{Breastw} & 0.935  & 0.994  & 0.990  & 0.625  & 0.807  & 0.766  & 0.976  & 0.991  \\ 
        \texttt{Campaign}  & 0.737  & 0.783  & 0.769  & 0.508  & 0.766  & 0.724  & 0.746 & 0.727  \\ 
        \texttt{Cardio}  & 0.934  & 0.921  & 0.935  & 0.498  & 0.461  & 0.723  & 0.777 & 0.907  \\ 
        \texttt{Cardiotocography}  & 0.691  & 0.664  & 0.784  & 0.488  & 0.372  & 0.579  & 0.493  & 0.610  \\ 
        \texttt{Celeba}  & 0.781  & 0.757  & 0.763  & 0.491  & 0.684  & 0.796  & 0.699 & 0.842  \\ 
        \texttt{Census}  & 0.655  & 0.500  & 0.500  & 0.527  & 0.668  & 0.659  & 0.672  & 0.662  \\ 
        \texttt{Cover} & 0.952  & 0.882  & 0.919  & 0.580  & 0.681  & 0.808  & 0.838  & 0.899  \\ 
        \texttt{Donors}  & 0.770  & 0.815  & 0.888  & 0.511  & 0.739  & 0.806  & 0.832 & 0.808  \\ 
        \texttt{Fault}  & 0.537  & 0.455  & 0.468  & 0.522  & 0.661  & 0.562  & 0.726 & 0.567  \\ 
        \texttt{Fraud}  & 0.954  & 0.943  & 0.949  & 0.769  & 0.931  & 0.924  & 0.956 & 0.944  \\ 
        \texttt{Glass}  & 0.661  & 0.760  & 0.710  & 0.517  & 0.729  & 0.560  & 0.881 & 0.785  \\ 
        \texttt{Hepatitis}  & 0.704  & 0.807  & 0.737  & 0.361  & 0.616  & 0.461  & 0.631  & 0.764  \\ 
        \texttt{Http}  & 0.994  & 0.991  & 0.980  & 0.249  & 0.921  & 0.998  & 0.051  & 0.995  \\ 
        \texttt{Internetads}  & 0.615  & 0.676  & 0.677  & 0.583  & 0.592  & 0.614  & 0.634 & 0.625  \\ 
        \texttt{Ionosphere}  & 0.838  & 0.783  & 0.717  & 0.514  & 0.629  & 0.758  & 0.924  & 0.848  \\ 
        \texttt{Landsat}  & 0.423  & 0.422  & 0.368  & 0.631  & 0.649  & 0.496  & 0.602  & 0.462  \\ 
        \texttt{Letter}  & 0.598  & 0.560  & 0.573  & 0.517  & 0.737  & 0.847  & 0.850  & 0.636  \\ 
        \texttt{Lymphography}  & 0.996  & 0.996  & 0.995  & 0.681  & 0.884  & 0.958  & 0.989  & 0.998  \\ 
        \texttt{Magic.gamma}  & 0.673  & 0.681  & 0.638  & 0.604  & 0.676  & 0.763  & 0.801 & 0.745  \\ 
        \texttt{Mammography}  & 0.871  & 0.905  & 0.906  & 0.451  & 0.658  & 0.749  & 0.849 & 0.846  \\ 
        \texttt{Musk}  & 1.000  & 0.948  & 0.953  & 0.538  & 0.790  & 0.999  & 0.882  & 1.000  \\ 
        \texttt{Optdigits}  & 0.507  & 0.500  & 0.500  & 0.519  & 0.533  & 0.402  & 0.386  & 0.586  \\ 
        \texttt{Pageblocks}  & 0.914  & 0.875  & 0.914  & 0.592  & 0.768  & 0.820  & 0.906 & 0.882  \\ 
        \texttt{Pendigits}  & 0.929  & 0.906  & 0.927  & 0.383  & 0.650  & 0.700  & 0.786  & 0.951  \\ 
        \texttt{Pima}  & 0.631  & 0.662  & 0.604  & 0.510  & 0.524  & 0.537  & 0.707 & 0.719  \\ 
        \texttt{Satellite}  & 0.662  & 0.633  & 0.583  & 0.562  & 0.627  & 0.715  & 0.702 & 0.698  \\ 
        \texttt{Satimage-2}  & 0.997  & 0.975  & 0.965  & 0.551  & 0.898  & 0.996  & 0.980 & 0.998  \\ 
        \texttt{Shuttle}  & 0.992  & 0.995  & 0.993  & 0.576  & 0.642  & 0.975  & 0.698 & 0.986  \\        
        \texttt{Skin}      & 0.547 & 0.471 & 0.490 & 0.548 & 0.265 & 0.461 & 0.718 & 0.662 \\ 
        \texttt{Smtp}      & 0.845 & 0.912 & 0.882 & 0.895 & 0.656 & 0.956 & 0.930 & 0.822 \\ 
        \texttt{Spambase}  & 0.534 & 0.688 & 0.656 & 0.584 & 0.459 & 0.510 & 0.545 & 0.550 \\ 
        \texttt{Speech}    & 0.466 & 0.489 & 0.470 & 0.512 & 0.512 & 0.466 & 0.487 & 0.475 \\ 
        \texttt{Stamps}    & 0.882 & 0.929 & 0.877 & 0.465 & 0.505 & 0.556 & 0.820 & 0.897 \\ 
        \texttt{Thyroid}   & 0.958 & 0.939 & 0.977 & 0.505 & 0.693 & 0.871 & 0.964 & 0.916 \\ 
        \texttt{Vertebral} & 0.426 & 0.263 & 0.417 & 0.394 & 0.449 & 0.563 & 0.400 & 0.345 \\ 
        \texttt{Vowels}    & 0.779 & 0.496 & 0.593 & 0.514 & 0.784 & 0.903 & 0.964 & 0.823 \\ 
        \texttt{Waveform}  & 0.669 & 0.739 & 0.603 & 0.609 & 0.661 & 0.617 & 0.729 & 0.722 \\ 
        \texttt{Wbc}       & 0.987 & 0.994 & 0.994 & 0.503 & 0.853 & 0.948 & 0.979 & 0.992 \\ 
        \texttt{Wdbc}      & 0.984 & 0.993 & 0.971 & 0.602 & 0.738 & 0.965 & 0.975 & 0.977 \\ 
        \texttt{Wilt}      & 0.317 & 0.345 & 0.394 & 0.465 & 0.649 & 0.659 & 0.552 & 0.329 \\ 
        \texttt{Wine}      & 0.671 & 0.865 & 0.738 & 0.507 & 0.455 & 0.374 & 0.425 & 0.906 \\ 
        \texttt{Wpbc}      & 0.485 & 0.519 & 0.489 & 0.493 & 0.488 & 0.493 & 0.502 & 0.510 \\ 
        \texttt{Yeast}     & 0.420 & 0.380 & 0.443 & 0.520 & 0.466 & 0.463 & 0.400 & 0.394 \\ 
        \hline
        {\textbf{Average}}    & { 0.740}          & { 0.730    }    & { 0.729   }    & { 0.543    }      & { 0.652    }     & { 0.712      }       & { 0.730  }        & {\textbf{0.757}} \\
        \hline
    \end{tabular}
    }
    \label{tab:tabular_auc_result}
\end{table}

\begin{table}[h!]
\renewcommand\thetable{C.3.1}
\scriptsize
\caption{Training PR value comparisons on tabular datasets}
    \centering
    \resizebox{0.99\textwidth}{!}{
    \begin{tabular}{l|c|c|c|c|c|c|c|c|c|c|c|c|c}
    \hline
        \textbf{Data} & \textbf{CBLOF} & \textbf{FeatureBagging} & \textbf{HBOS} & \textbf{IForest} & \textbf{kNN} & \textbf{LODA} & \textbf{LOF} & \textbf{MCD} & \textbf{PCA} & \textbf{DAGMM} & \textbf{DROCC} & \textbf{GOAD} & \textbf{PlanarFlow} \\ 
        \hline
        \texttt{Aloi} & 0.037 & 0.104 & 0.034 & 0.034 & 0.048 & 0.033 & 0.097 & 0.032 & 0.037 & 0.033 & 0.030 & 0.033 & 0.032 \\
        \texttt{Annthyroid} & 0.169 & 0.206 & 0.228 & 0.312 & 0.224 & 0.098 & 0.163 & 0.503 & 0.196 & 0.109 & 0.185 & 0.131 & 0.654 \\
        \texttt{Backdoor} & 0.546 & 0.217 & 0.052 & 0.045 & 0.479 & 0.101 & 0.358 & 0.121 & 0.531 & 0.250 & 0.025 & 0.347 & 0.336 \\
        \texttt{Breastw} & 0.890 & 0.284 & 0.954 & 0.956 & 0.932 & 0.955 & 0.296 & 0.962 & 0.946 & 0.660 & 0.776 & 0.826 & 0.908 \\
        \texttt{Campaign} & 0.287 & 0.145 & 0.352 & 0.279 & 0.289 & 0.131 & 0.158 & 0.325 & 0.284 & 0.163 & 0.113 & 0.105 & 0.191 \\
        \texttt{Cardio} & 0.482 & 0.161 & 0.458 & 0.559 & 0.402 & 0.428 & 0.159 & 0.364 & 0.609 & 0.193 & 0.272 & 0.540 & 0.471 \\
        \texttt{Cardiotocography} & 0.335 & 0.276 & 0.361 & 0.436 & 0.324 & 0.463 & 0.271 & 0.311 & 0.462 & 0.271 & 0.258 & 0.403 & 0.348 \\
        \texttt{Celeba} & 0.069 & 0.024 & 0.089 & 0.063 & 0.061 & 0.047 & 0.018 & 0.092 & 0.112 & 0.044 & 0.047 & 0.021 & 0.066 \\
        \texttt{Census} & 0.087 & 0.061 & 0.073 & 0.073 & 0.088 & 0.065 & 0.069 & 0.153 & 0.087 & 0.062 & 0.058 & 0.072 & 0.073 \\
        \texttt{Cover} & 0.070 & 0.019 & 0.026 & 0.052 & 0.054 & 0.090 & 0.019 & 0.016 & 0.075 & 0.044 & 0.056 & 0.005 & 0.010 \\
        \texttt{Donors} & 0.148 & 0.120 & 0.135 & 0.124 & 0.182 & 0.255 & 0.109 & 0.141 & 0.166 & 0.086 & 0.123 & 0.040 & 0.241 \\
        \texttt{Fault} & 0.473 & 0.396 & 0.360 & 0.395 & 0.522 & 0.336 & 0.388 & 0.334 & 0.332 & 0.361 & 0.496 & 0.381 & 0.329 \\
        \texttt{Fraud} & 0.145 & 0.003 & 0.209 & 0.145 & 0.169 & 0.146 & 0.003 & 0.488 & 0.149 & 0.084 & 0.002 & 0.257 & 0.447 \\
        \texttt{Glass} & 0.144 & 0.151 & 0.161 & 0.144 & 0.167 & 0.090 & 0.144 & 0.113 & 0.112 & 0.111 & 0.159 & 0.075 & 0.113 \\
        \texttt{Hepatitis} & 0.304 & 0.225 & 0.328 & 0.243 & 0.252 & 0.275 & 0.214 & 0.363 & 0.339 & 0.253 & 0.221 & 0.291 & 0.317 \\
        \texttt{Http} & 0.464 & 0.047 & 0.302 & 0.886 & 0.010 & 0.004 & 0.050 & 0.865 & 0.500 & 0.368 & 0.004 & 0.441 & 0.363 \\
        \texttt{Internetads} & 0.296 & 0.182 & 0.523 & 0.486 & 0.296 & 0.242 & 0.232 & 0.344 & 0.276 & 0.207 & 0.197 & 0.288 & 0.262 \\
        \texttt{Ionosphere} & 0.881 & 0.821 & 0.353 & 0.779 & 0.911 & 0.741 & 0.807 & 0.947 & 0.721 & 0.473 & 0.728 & 0.781 & 0.824 \\
        \texttt{Landsat} & 0.212 & 0.246 & 0.231 & 0.194 & 0.258 & 0.183 & 0.250 & 0.253 & 0.163 & 0.230 & 0.272 & 0.198 & 0.186 \\
        \texttt{Letter} & 0.166 & 0.445 & 0.078 & 0.086 & 0.203 & 0.083 & 0.433 & 0.174 & 0.076 & 0.083 & 0.252 & 0.098 & 0.153 \\
        \texttt{Lymphography} & 0.915 & 0.090 & 0.919 & 0.972 & 0.894 & 0.490 & 0.135 & 0.767 & 0.935 & 0.454 & 0.463 & 0.897 & 0.417 \\
        \texttt{Magic.gamma} & 0.666 & 0.539 & 0.617 & 0.638 & 0.723 & 0.579 & 0.520 & 0.631 & 0.589 & 0.450 & 0.627 & 0.326 & 0.692 \\
        \texttt{Mammography} & 0.139 & 0.070 & 0.132 & 0.218 & 0.181 & 0.218 & 0.085 & 0.036 & 0.204 & 0.111 & 0.114 & 0.046 & 0.074 \\
        \texttt{Musk} & 1.000 & 0.139 & 0.999 & 0.945 & 0.708 & 0.842 & 0.118 & 0.992 & 1.000 & 0.500 & 0.196 & 1.000 & 0.391 \\
        \texttt{Optdigits} & 0.059 & 0.036 & 0.192 & 0.046 & 0.022 & 0.029 & 0.035 & 0.022 & 0.027 & 0.026 & 0.032 & 0.039 & 0.027 \\
        \texttt{Pageblocks} & 0.547 & 0.341 & 0.319 & 0.464 & 0.556 & 0.410 & 0.292 & 0.617 & 0.525 & 0.255 & 0.632 & 0.373 & 0.538 \\
        \texttt{Pendigits} & 0.192 & 0.048 & 0.247 & 0.260 & 0.099 & 0.186 & 0.040 & 0.069 & 0.219 & 0.056 & 0.027 & 0.075 & 0.060 \\
        \texttt{Pima} & 0.484 & 0.412 & 0.577 & 0.510 & 0.530 & 0.404 & 0.406 & 0.498 & 0.492 & 0.372 & 0.413 & 0.476 & 0.476 \\
        \texttt{Satellite} & 0.656 & 0.378 & 0.688 & 0.649 & 0.582 & 0.613 & 0.381 & 0.768 & 0.606 & 0.527 & 0.465 & 0.658 & 0.595 \\
        \texttt{Satimage-2} & 0.972 & 0.042 & 0.760 & 0.917 & 0.690 & 0.857 & 0.041 & 0.682 & 0.872 & 0.289 & 0.076 & 0.949 & 0.484 \\
        \texttt{Shuttle} & 0.184 & 0.081 & 0.965 & 0.976 & 0.193 & 0.168 & 0.109 & 0.841 & 0.913 & 0.438 & 0.072 & 0.136 & 0.346 \\
        \texttt{Skin} & 0.289 & 0.207 & 0.232 & 0.254 & 0.290 & 0.180 & 0.221 & 0.490 & 0.172 & 0.226 & 0.285 & 0.232 & 0.335 \\
        \texttt{Smtp} & 0.403 & 0.001 & 0.005 & 0.005 & 0.415 & 0.312 & 0.022 & 0.006 & 0.382 & 0.179 & 0.000 & 0.358 & 0.006 \\
        \texttt{Spambase} & 0.402 & 0.344 & 0.518 & 0.487 & 0.415 & 0.387 & 0.360 & 0.349 & 0.409 & 0.389 & 0.383 & 0.387 & 0.433 \\
        \texttt{Speech} & 0.019 & 0.022 & 0.023 & 0.020 & 0.019 & 0.016 & 0.022 & 0.019 & 0.018 & 0.022 & 0.020 & 0.019 & 0.018 \\
        \texttt{Stamps} & 0.211 & 0.143 & 0.332 & 0.347 & 0.317 & 0.280 & 0.153 & 0.257 & 0.364 & 0.198 & 0.241 & 0.285 & 0.284 \\
        \texttt{Thyroid} & 0.272 & 0.069 & 0.501 & 0.562 & 0.392 & 0.189 & 0.077 & 0.702 & 0.356 & 0.126 & 0.338 & 0.318 & 0.734 \\
        \texttt{Vertebral} & 0.123 & 0.124 & 0.091 & 0.097 & 0.095 & 0.089 & 0.130 & 0.101 & 0.099 & 0.134 & 0.117 & 0.123 & 0.111 \\
        \texttt{Vowels} & 0.166 & 0.314 & 0.078 & 0.162 & 0.443 & 0.127 & 0.326 & 0.085 & 0.069 & 0.041 & 0.178 & 0.154 & 0.295 \\
        \texttt{Waveform} & 0.122 & 0.078 & 0.048 & 0.056 & 0.133 & 0.040 & 0.071 & 0.040 & 0.044 & 0.032 & 0.150 & 0.042 & 0.150 \\
        \texttt{Wbc} & 0.691 & 0.037 & 0.728 & 0.948 & 0.743 & 0.898 & 0.077 & 0.839 & 0.913 & 0.327 & 0.358 & 0.736 & 0.431 \\
        \texttt{Wdbc} & 0.688 & 0.154 & 0.761 & 0.702 & 0.521 & 0.527 & 0.128 & 0.395 & 0.613 & 0.152 & 0.039 & 0.589 & 0.568 \\
        \texttt{Wilt} & 0.040 & 0.081 & 0.039 & 0.044 & 0.049 & 0.036 & 0.083 & 0.153 & 0.032 & 0.047 & 0.041 & 0.065 & 0.115 \\
        \texttt{Wine} & 0.170 & 0.061 & 0.412 & 0.207 & 0.081 & 0.250 & 0.064 & 0.737 & 0.264 & 0.120 & 0.126 & 0.229 & 0.086 \\
        \texttt{Wpbc} & 0.227 & 0.206 & 0.241 & 0.237 & 0.234 & 0.226 & 0.210 & 0.257 & 0.229 & 0.214 & 0.234 & 0.214 & 0.236 \\
        \texttt{Yeast} & 0.314 & 0.325 & 0.328 & 0.304 & 0.294 & 0.330 & 0.315 & 0.298 & 0.302 & 0.353 & 0.284 & 0.332 & 0.309 \\
        \hline
        \textbf{Average} & 0.351 & 0.184 & 0.349 & 0.377 & 0.337 & 0.292 & 0.188 & 0.382 & 0.366 & 0.220 & 0.221 & 0.313 & 0.316 \\
        \hline
    \end{tabular}
    }
    \label{tab:tabular_ap_result-1}
\end{table}

\begin{table}[h!]
\renewcommand\thetable{C.3.2}
\scriptsize
\caption{Training PR value comparisons on tabular datasets}
    \centering
    \resizebox{0.8\textwidth}{!}{
    \begin{tabular}{l|c|c|c|c|c|c|c|c}
    \hline
        \textbf{Data} &  \textbf{OCSVM} & \textbf{COPOD} & \textbf{ECOD} & \textbf{DeepSVDD} & \textbf{ICL} & \textbf{DDPM} & \textbf{DTE} & \textbf{ODIM} \\ 
        \hline
        \texttt{Aloi}            & 0.039 & 0.031 & 0.033 & 0.034 & 0.046 & 0.036 & 0.056 & 0.041 \\ 
        \texttt{Annthyroid}      & 0.188 & 0.174 & 0.272 & 0.192 & 0.123 & 0.297 & 0.228 & 0.161 \\ 
        \texttt{Backdoor}        & 0.534 & 0.025 & 0.025 & 0.372 & 0.717 & 0.520 & 0.473 & 0.272 \\ 
        \texttt{Breastw}         & 0.897 & 0.989 & 0.982 & 0.482 & 0.635 & 0.537 & 0.921 & 0.988 \\ 
        \texttt{Campaign}        & 0.283 & 0.368 & 0.354 & 0.149 & 0.267 & 0.299 & 0.281 & 0.297 \\ 
        \texttt{Cardio}          & 0.536 & 0.576 & 0.567 & 0.177 & 0.108 & 0.278 & 0.376 & 0.564 \\ 
        \texttt{Cardiotocography} & 0.408 & 0.403 & 0.502 & 0.252 & 0.188 & 0.338 & 0.312 & 0.389 \\ 
        \texttt{Celeba}          & 0.103 & 0.093 & 0.095 & 0.031 & 0.045 & 0.093 & 0.052 & 0.118 \\ 
        \texttt{Census}          & 0.085 & 0.062 & 0.062 & 0.075 & 0.095 & 0.086 & 0.090 & 0.084 \\ 
        \texttt{Cover}           & 0.099 & 0.068 & 0.113 & 0.048 & 0.022 & 0.046 & 0.048 & 0.032 \\ 
        \texttt{Donors}          & 0.139 & 0.209 & 0.265 & 0.112 & 0.119 & 0.143 & 0.188 & 0.124 \\ 
        \texttt{Fault}           & 0.401 & 0.313 & 0.326 & 0.375 & 0.473 & 0.392 & 0.532 & 0.444 \\ 
        \texttt{Fraud}           & 0.110 & 0.252 & 0.215 & 0.250 & 0.127 & 0.146 & 0.137 & 0.257 \\ 
        \texttt{Glass}           & 0.130 & 0.111 & 0.183 & 0.090 & 0.122 & 0.073 & 0.206 & 0.119 \\ 
        \texttt{Hepatitis}       & 0.278 & 0.389 & 0.295 & 0.170 & 0.231 & 0.165 & 0.238 & 0.315 \\ 
        \texttt{Http}            & 0.356 & 0.280 & 0.145 & 0.093 & 0.091 & 0.642 & 0.024 & 0.280 \\ 
        \texttt{Internetads}     & 0.291 & 0.505 & 0.505 & 0.252 & 0.237 & 0.295 & 0.290 & 0.234 \\ 
        \texttt{Ionosphere}      & 0.829 & 0.663 & 0.633 & 0.392 & 0.472 & 0.633 & 0.920 & 0.867 \\ 
        \texttt{Landsat}         & 0.175 & 0.176 & 0.164 & 0.362 & 0.329 & 0.200 & 0.255 & 0.179 \\ 
        \texttt{Letter}          & 0.113 & 0.068 & 0.077 & 0.099 & 0.208 & 0.367 & 0.256 & 0.121 \\ 
        \texttt{Lymphography}    & 0.885 & 0.907 & 0.894 & 0.254 & 0.264 & 0.731 & 0.805 & 0.967 \\ 
        \texttt{Magic.gamma}     & 0.625 & 0.588 & 0.533 & 0.499 & 0.548 & 0.651 & 0.730 & 0.693 \\ 
        \texttt{Mammography}     & 0.187 & 0.430 & 0.435 & 0.025 & 0.046 & 0.099 & 0.175 & 0.098  \\ 
        \texttt{Musk}            & 1.000 & 0.369 & 0.475 & 0.107 & 0.128 & 0.984 & 0.434 & 1.000 \\ 
        \texttt{Optdigits}       & 0.027 & 0.029 & 0.029 & 0.039 & 0.030 & 0.022 & 0.021 & 0.030 \\ 
        \texttt{Pageblocks}      & 0.531 & 0.370 & 0.520 & 0.288 & 0.285 & 0.493 & 0.530 & 0.509 \\ 
        \texttt{Pendigits}       & 0.226 & 0.177 & 0.270 & 0.022 & 0.045 & 0.056 & 0.089 & 0.302 \\ 
        \texttt{Pima}            & 0.477 & 0.536 & 0.484 & 0.366 & 0.385 & 0.400 & 0.528 & 0.491 \\ 
        \texttt{Satellite}       & 0.654 & 0.570 & 0.526 & 0.406 & 0.451 & 0.662 & 0.563 & 0.652 \\ 
        \texttt{Satimage-2}      & 0.965 & 0.797 & 0.666 & 0.052 & 0.102 & 0.783 & 0.507 & 0.949 \\ 
        \texttt{Shuttle}         & 0.907 & 0.962 & 0.905 & 0.149 & 0.135 & 0.779 & 0.187 & 0.947 \\ 
        \texttt{Skin}            & 0.220 & 0.179 & 0.183 & 0.221 & 0.173 & 0.175 & 0.290 & 0.239 \\ 
        \texttt{Smtp}            & 0.383 & 0.005 & 0.589 & 0.240 & 0.004 & 0.502 & 0.411 & 0.261 \\ 
        \texttt{Spambase}        & 0.402 & 0.544 & 0.518 & 0.456 & 0.370 & 0.384 & 0.407 & 0.410 \\ 
        \texttt{Speech}          & 0.019 & 0.019 & 0.020 & 0.018 & 0.020 & 0.020 & 0.019 & 0.018 \\ 
        \texttt{Stamps}          & 0.318 & 0.398 & 0.324 & 0.099 & 0.117 & 0.143 & 0.273 & 0.336 \\ 
        \texttt{Thyroid}         & 0.329 & 0.179 & 0.472 & 0.024 & 0.066 & 0.325 & 0.360 & 0.327 \\ 
        \texttt{Vertebral}       & 0.107 & 0.085 & 0.110 & 0.107 & 0.115 & 0.150 & 0.098 & 0.109 \\ 
        \texttt{Vowels}          & 0.196 & 0.034 & 0.083 & 0.037 & 0.219 & 0.311 & 0.504 & 0.375 \\ 
        \texttt{Waveform}        & 0.052 & 0.057 & 0.040 & 0.061 & 0.063 & 0.050 & 0.109 & 0.053 \\ 
        \texttt{Wbc}             & 0.813 & 0.883 & 0.882 & 0.069 & 0.211 & 0.758 & 0.722 & 0.710 \\ 
        \texttt{Wdbc}            & 0.539 & 0.760 & 0.493 & 0.063 & 0.065 & 0.482 & 0.465 & 0.612 \\ 
        \texttt{Wilt}            & 0.035 & 0.037 & 0.042 & 0.046 & 0.109 & 0.076 & 0.054 & 0.036 \\ 
        \texttt{Wine}            & 0.135 & 0.364 & 0.195 & 0.116 & 0.087 & 0.074 & 0.074 & 0.239 \\ 
        \texttt{Wpbc}            & 0.222 & 0.234 & 0.217 & 0.240 & 0.234 & 0.238 & 0.227 & 0.236 \\ 
        \texttt{Yeast}           & 0.303 & 0.308 & 0.332 & 0.350 & 0.318 & 0.320 & 0.294 & 0.287 \\ 
        \hline
        \textbf{Average}           & 0.360 & 0.339 & 0.349 & 0.182 & 0.201 & 0.332 & 0.321 & \textbf{0.366} \\ 
        \hline
    \end{tabular}
    }
    \label{tab:tabular_ap_result}
\end{table}

 \clearpage

\clearpage

\subsection*{C.3. Detailed AUC and PR results over image datasets (training data)}

Tables \ref{tab:image_auc_result-1}-\ref{tab:image_auc_result} and \ref{tab:image_ap_result-1}-\ref{tab:image_ap_result} present the AUC and PR results on the image datasets.

\begin{table}[h!]
\renewcommand\thetable{C.4.1}
\scriptsize
\caption{Training AUC value comparisons on image datasets}
    \centering
    \resizebox{0.95\textwidth}{!}{
    \begin{tabular}{l|c|c|c|c|c|c|c|c|c|c|c|c|c}
    \hline
        \textbf{Data} & \textbf{CBLOF} & \textbf{FeatureBagging} & \textbf{HBOS} & \textbf{IForest} & \textbf{kNN} & \textbf{LODA} & \textbf{LOF} & \textbf{MCD} & \textbf{PCA} & \textbf{DAGMM} & \textbf{DROCC} & \textbf{GOAD} & \textbf{PlanarFlow} \\ 
        \hline
        \texttt{MNIST} & 0.843 & 0.664 & 0.574 & 0.811 & 0.867 & 0.564 & 0.658 & 0.856 & 0.848 & 0.631 & 0.615 & 0.698 & 0.645 \\
        \texttt{MNIST-C} & 0.757 & 0.702 & 0.689 & 0.733 & 0.786 & 0.591 & 0.699 & 0.739 & 0.741 & 0.581 & 0.594 & 0.752 & 0.705 \\
        \texttt{FMNIST} & 0.871 & 0.748 & 0.748 & 0.831 & 0.875 & 0.672 & 0.738 & 0.840 & 0.853 & 0.664 & 0.564 & 0.860 & 0.819 \\
        \texttt{CIFAR10} & 0.663 & 0.687 & 0.572 & 0.629 & 0.659 & 0.591 & 0.686 & 0.639 & 0.659 & 0.530 & 0.503 & 0.659 & 0.621 \\
        \texttt{SVHN} & 0.601 & 0.629 & 0.542 & 0.580 & 0.604 & 0.534 & 0.628 & 0.583 & 0.599 & 0.528 & 0.521 & 0.597 & 0.580 \\
        \texttt{MVTec-AD} & 0.754 & 0.745 & 0.732 & 0.747 & 0.763 & 0.644 & 0.742 & 0.618 & 0.724 & 0.596 & 0.544 & 0.730 & 0.637 \\
        \hline
        \textbf{Average} & 0.748 & 0.696 & 0.643 & 0.722 & 0.759 & 0.599 & 0.692 & 0.712 & 0.737 & 0.588 & 0.557 & 0.716 & 0.668 \\ 
        \hline
    \end{tabular}
    }
    \label{tab:image_auc_result-1}
\end{table}

\begin{table}[h!]
\renewcommand\thetable{C.4.2}
\scriptsize
\caption{Training AUC value comparisons on image datasets}
    \centering
    \resizebox{0.7\textwidth}{!}{
    \begin{tabular}{l|c|c|c|c|c|c|c|c}
    \hline
        \textbf{Data} &  \textbf{OCSVM} & \textbf{COPOD} & \textbf{ECOD} & \textbf{DeepSVDD} & \textbf{ICL} & \textbf{DDPM} & \textbf{DTE} & \textbf{ODIM} \\ 
        \hline
        \texttt{MNIST}           & 0.849 & 0.500 & 0.500 & 0.605 & 0.691 & 0.816 & 0.853 & 0.836 \\ 
        \texttt{MNIST-C}         & 0.751 & 0.500 & 0.500 & 0.552 & 0.670 & 0.751 & 0.788 & 0.736 \\ 
        \texttt{FMNIST}    & 0.860 & 0.500 & 0.500 & 0.647 & 0.758 & 0.861 & 0.873 & 0.909 \\ 
        \texttt{CIFAR10}         & 0.663 & 0.548 & 0.567 & 0.555 & 0.557 & 0.663 & 0.660 & 0.922 \\ 
        \texttt{SVHN}            & 0.604 & 0.500 & 0.500 & 0.521 & 0.571 & 0.605 & 0.607 & 0.568 \\ 
        \texttt{MVTec-AD}        & 0.735 & 0.500 & 0.500 & 0.603 & 0.683 & 0.732 & 0.761 & 0.906 \\ 
        \hline
        \textbf{Average}           & 0.744 & 0.508 & 0.511 & 0.580 & 0.655 & 0.738 & 0.757 & \textbf{0.813} \\ 
        \hline
    \end{tabular}
    }
    \label{tab:image_auc_result}
\end{table}

\begin{table}[h!]
\renewcommand\thetable{C.5.1}
\scriptsize
\caption{Training PR value comparisons on image datasets}
    \centering
    \resizebox{0.95\textwidth}{!}{
    \begin{tabular}{l|c|c|c|c|c|c|c|c|c|c|c|c|c}
    \hline
        \textbf{Data} & \textbf{CBLOF} & \textbf{FeatureBagging} & \textbf{HBOS} & \textbf{IForest} & \textbf{kNN} & \textbf{LODA} & \textbf{LOF} & \textbf{MCD} & \textbf{PCA} & \textbf{DAGMM} & \textbf{DROCC} & \textbf{GOAD} & \textbf{PlanarFlow} \\ 
        \hline
        \texttt{MNIST} & 0.386 & 0.241 & 0.109 & 0.290 & 0.409 & 0.170 & 0.233 & 0.308 & 0.381 & 0.215 & 0.237 & 0.297 & 0.259 \\
        \texttt{MNIST-C} & 0.173 & 0.128 & 0.126 & 0.178 & 0.191 & 0.101 & 0.127 & 0.166 & 0.170 & 0.092 & 0.096 & 0.177 & 0.154 \\
        \texttt{FMNIST} & 0.329 & 0.194 & 0.269 & 0.320 & 0.346 & 0.180 & 0.188 & 0.245 & 0.319 & 0.138 & 0.106 & 0.328 & 0.297 \\
        \texttt{CIFAR10} & 0.103 & 0.115 & 0.075 & 0.089 & 0.102 & 0.086 & 0.114 & 0.084 & 0.101 & 0.062 & 0.060 & 0.102 & 0.085 \\
        \texttt{SVHN} & 0.079 & 0.083 & 0.063 & 0.073 & 0.079 & 0.064 & 0.083 & 0.068 & 0.078 & 0.059 & 0.060 & 0.078 & 0.074 \\
        \texttt{MVTec-AD} & 0.570 & 0.536 & 0.546 & 0.570 & 0.580 & 0.464 & 0.532 & 0.451 & 0.540 & 0.362 & 0.317 & 0.546 & 0.454 \\
        \hline
        \textbf{Average} & 0.273 & 0.216 & 0.198 & 0.253 & 0.285 & 0.177 & 0.213 & 0.221 & 0.265 & 0.155 & 0.146 & 0.255 & 0.221 \\ 
        \hline
    \end{tabular}
    }
    \label{tab:image_ap_result-1}
\end{table}

\begin{table}[h!]
\renewcommand\thetable{C.5.2}
\scriptsize
\caption{Training PR value comparisons on image datasets}
    \centering
    \resizebox{0.7\textwidth}{!}{
    \begin{tabular}{l|c|c|c|c|c|c|c|c}
    \hline
        \textbf{Data} &  \textbf{OCSVM} & \textbf{COPOD} & \textbf{ECOD} & \textbf{DeepSVDD} & \textbf{ICL} & \textbf{DDPM} & \textbf{DTE} & \textbf{ODIM} \\ 
        \hline
        \texttt{MNIST}           & 0.385 & 0.092 & 0.092 & 0.253 & 0.232 & 0.374 & 0.400 & 0.462 \\ 
        \texttt{MNIST-C}         & 0.179 & 0.050 & 0.050 & 0.097 & 0.098 & 0.178 & 0.192 & 0.245 \\ 
        \texttt{FMNIST}    & 0.329 & 0.050 & 0.050 & 0.181 & 0.158 & 0.325 & 0.339 & 0.650 \\ 
        \texttt{CIFAR10}         & 0.102 & 0.065 & 0.067 & 0.073 & 0.070 & 0.102 & 0.104 & 0.530 \\ 
        \texttt{SVHN}            & 0.078 & 0.050 & 0.050 & 0.063 & 0.068 & 0.078 & 0.080 & 0.078 \\ 
        \texttt{MVTec-AD}        & 0.555 & 0.236 & 0.236 & 0.387 & 0.404 & 0.546 & 0.578 & 0.611 \\ 
        \hline
        \textbf{Average}           & 0.271 & 0.090 & 0.091 & 0.176 & 0.172 & 0.267 & 0.282 & \textbf{0.429} \\ 
        \hline
    \end{tabular}
    }
    \label{tab:image_ap_result}
\end{table}

\clearpage

\subsection*{C.4. Detailed AUC and PR results over text datasets (training data)}

Tables \ref{tab:text_auc_result-1}-\ref{tab:text_auc_result} and \ref{tab:text_ap_result-1}-\ref{tab:text_ap_result} present the AUC and PR results on the text datasets.

\begin{table}[h!]
\renewcommand\thetable{C.6.1}
\scriptsize
\caption{Training AUC value comparisons on text datasets}
    \centering
    \resizebox{0.99\textwidth}{!}{
    \begin{tabular}{l|c|c|c|c|c|c|c|c|c|c|c|c|c}
    \hline
        \textbf{Data} & \textbf{CBLOF} & \textbf{FeatureBagging} & \textbf{HBOS} & \textbf{IForest} & \textbf{kNN} & \textbf{LODA} & \textbf{LOF} & \textbf{MCD} & \textbf{PCA} & \textbf{DAGMM} & \textbf{DROCC} & \textbf{GOAD} & \textbf{PlanarFlow} \\ 
        \hline
        \texttt{Amazon} & 0.579 & 0.572 & 0.563 & 0.558 & 0.603 & 0.526 & 0.571 & 0.597 & 0.550 & 0.501 & 0.500 & 0.560 & 0.495 \\
        \texttt{20news} & 0.564 & 0.610 & 0.537 & 0.550 & 0.567 & 0.539 & 0.610 & 0.583 & 0.545 & 0.518 & 0.496 & 0.553 & 0.513 \\
        \texttt{Agnews} & 0.619 & 0.715 & 0.554 & 0.584 & 0.647 & 0.568 & 0.714 & 0.665 & 0.566 & 0.508 & 0.500 & 0.592 & 0.497 \\
        \texttt{Imdb} & 0.496 & 0.499 & 0.499 & 0.489 & 0.494 & 0.466 & 0.500 & 0.504 & 0.478 & 0.487 & 0.500 & 0.486 & 0.493 \\
        \texttt{Yelp} & 0.635 & 0.661 & 0.600 & 0.602 & 0.670 & 0.581 & 0.661 & 0.655 & 0.592 & 0.498 & 0.504 & 0.590 & 0.527 \\
        \hline
        \textbf{Average} & 0.579 & 0.611 & 0.550 & 0.557 & 0.596 & 0.536 & 0.611 & 0.601 & 0.546 & 0.502 & 0.500 & 0.556 & 0.505 \\ 
        \hline
    \end{tabular}
    }
    \label{tab:text_auc_result-1}
\end{table}

\begin{table}[h!]
\renewcommand\thetable{C.6.2}
\scriptsize
\caption{Training AUC value comparisons on text datasets}
    \centering
    \resizebox{0.7\textwidth}{!}{
    \begin{tabular}{l|c|c|c|c|c|c|c|c}
    \hline
        \textbf{Data} &  \textbf{OCSVM} & \textbf{COPOD} & \textbf{ECOD} & \textbf{DeepSVDD} & \textbf{ICL} & \textbf{DDPM} & \textbf{DTE} & \textbf{ODIM} \\ 
        \hline
        \texttt{Amazon}          & 0.565 & 0.571 & 0.541 & 0.464 & 0.528 & 0.551 & 0.603 & 0.597 \\ 
        \texttt{20news}          & 0.559 & 0.533 & 0.544 & 0.515 & 0.547 & 0.547 & 0.570 & 0.687 \\ 
        \texttt{Agnews}          & 0.601 & 0.551 & 0.552 & 0.494 & 0.591 & 0.571 & 0.652 & 0.805 \\ 
        \texttt{Imdb}            & 0.484 & 0.512 & 0.471 & 0.526 & 0.521 & 0.478 & 0.495 & 0.522 \\ 
        \texttt{Yelp}            & 0.621 & 0.605 & 0.578 & 0.524 & 0.545 & 0.594 & 0.671 & 0.682 \\ 
        \hline
        \textbf{Average}           & 0.566 & 0.554 & 0.537 & 0.504 & 0.546 & 0.548 & 0.598 & \textbf{0.659} \\ 
        \hline
    \end{tabular}
    }
    \label{tab:text_auc_result}
\end{table}

\begin{table}[h!]
\renewcommand\thetable{C.7.1}
\scriptsize
\caption{Training PR value comparisons on text datasets}
    \centering
    \resizebox{0.99\textwidth}{!}{
    \begin{tabular}{l|c|c|c|c|c|c|c|c|c|c|c|c|c}
    \hline
        \textbf{Data} & \textbf{CBLOF} & \textbf{FeatureBagging} & \textbf{HBOS} & \textbf{IForest} & \textbf{kNN} & \textbf{LODA} & \textbf{LOF} & \textbf{MCD} & \textbf{PCA} & \textbf{DAGMM} & \textbf{DROCC} & \textbf{GOAD} & \textbf{PlanarFlow} \\ 
        \hline
        \texttt{Amazon} & 0.067 & 0.087 & 0.061 & 0.062 & 0.069 & 0.062 & 0.088 & 0.072 & 0.062 & 0.054 & 0.055 & 0.063 & 0.056 \\
        \texttt{20news} & 0.072 & 0.125 & 0.059 & 0.064 & 0.082 & 0.064 & 0.125 & 0.077 & 0.061 & 0.053 & 0.051 & 0.066 & 0.050 \\
        \texttt{Agnews} & 0.061 & 0.058 & 0.059 & 0.058 & 0.062 & 0.054 & 0.058 & 0.062 & 0.057 & 0.049 & 0.050 & 0.058 & 0.050 \\
        \texttt{Imdb} & 0.047 & 0.049 & 0.047 & 0.047 & 0.047 & 0.046 & 0.049 & 0.049 & 0.046 & 0.049 & 0.050 & 0.047 & 0.051 \\
        \texttt{Yelp} & 0.073 & 0.085 & 0.070 & 0.070 & 0.083 & 0.067 & 0.085 & 0.075 & 0.069 & 0.049 & 0.051 & 0.068 & 0.056 \\
        \hline
        \textbf{Average} & 0.064 & 0.081 & 0.059 & 0.060 & 0.068 & 0.059 & 0.081 & 0.067 & 0.059 & 0.051 & 0.051 & 0.060 & 0.053 \\ 
        \hline
    \end{tabular}
    }
    \label{tab:text_ap_result-1}
\end{table}

\begin{table}[h!]
\renewcommand\thetable{C.7.2}
\scriptsize
\caption{Training PR value comparisons on text datasets}
    \centering
    \resizebox{0.7\textwidth}{!}{
    \begin{tabular}{l|c|c|c|c|c|c|c|c}
    \hline
        \textbf{Data} &  \textbf{OCSVM} & \textbf{COPOD} & \textbf{ECOD} & \textbf{DeepSVDD} & \textbf{ICL} & \textbf{DDPM} & \textbf{DTE} & \textbf{ODIM} \\ 
        \hline
        \texttt{Amazon}          & 0.064 & 0.061 & 0.062 & 0.058 & 0.063 & 0.063 & 0.072 & 0.063 \\ 
        \texttt{20news}          & 0.068 & 0.059 & 0.058 & 0.053 & 0.069 & 0.062 & 0.085 & 0.111 \\ 
        \texttt{Agnews}          & 0.059 & 0.060 & 0.055 & 0.046 & 0.052 & 0.057 & 0.062 & 0.176 \\ 
        \texttt{Imdb}            & 0.047 & 0.050 & 0.045 & 0.053 & 0.054 & 0.046 & 0.047 & 0.049 \\ 
        \texttt{Yelp}            & 0.073 & 0.072 & 0.065 & 0.058 & 0.054 & 0.069 & 0.085 & 0.086 \\ 
        \hline
        \textbf{Average}           & 0.062 & 0.060 & 0.057 & 0.054 & 0.058 & 0.059 & 0.070 & \textbf{0.097} \\ 
        \hline
    \end{tabular}
    }
    \label{tab:text_ap_result}
\end{table}

 \clearpage

\subsection*{C.8. Ablation studies}
\label{app:ablation_studies}

\paragraph{Number of samples used in the IWAE} 
Table \ref{tab:abl_k} summarizes the AUC values on several tabular datasets with various $K$s from 1 to 100. 
Note that the IWAE with $K=1$ equals the original VAE. 
As expected, a lower bound closer to the log-likelihood tends to provide a more obvious IM effect, leading to better identification performances. 
We can also observe that when the value of $K$ becomes larger than 50, the enhancement seems saturated. 
For this reason, we set $K=50$ in our experiments.  

\begin{table*}[h!]
\renewcommand\thetable{C.40}
\caption{AUC results of the ODIM with various values of $K$.}
\centering
\resizebox{0.74\textwidth}{!}{
\begin{tabular}{l|ccccccccc}
\hline
\textbf{Data}           & $\mathbf{K=1}$       & $\mathbf{K=2}$       & $\mathbf{K=5}$       & $\mathbf{K=10}$      & $\mathbf{K=20}$      & $\mathbf{K=50}$      & $\mathbf{K=70}$      & $\mathbf{K=100}$      \\
\hline
\texttt{Glass}       & 0.525 & 0.676 & 0.636 & 0.689 & 0.702 & 0.704 & 0.704 & 0.727  \\
\texttt{Mammography} & 0.565 & 0.794 & 0.807 & 0.835 & 0.841 & 0.852 & 0.852 & 0.853  \\
\texttt{Pendigits}   & 0.847 & 0.930 & 0.922 & 0.955 & 0.959 & 0.950 & 0.950 & 0.950  \\
\texttt{Pima}        & 0.626 & 0.663 & 0.682 & 0.706 & 0.696 & 0.706 & 0.706 & 0.705  \\
\texttt{Vowels}      & 0.528 & 0.655 & 0.549 & 0.583 & 0.875 & 0.900 & 0.900 & 0.896  \\
\texttt{Wbc}         & 0.866 & 0.929 & 0.915 & 0.919 & 0.937 & 0.935 & 0.935 & 0.939  \\
\texttt{Cardio}      & 0.794 & 0.919 & 0.913 & 0.932 & 0.925 & 0.914 & 0.914 & 0.919  \\
\texttt{Thyroid}     & 0.636 & 0.860 & 0.868 & 0.887 & 0.908 & 0.924 & 0.924 & 0.926  \\ \hline
\textbf{Average} & 0.673 &	0.803 &	0.787 &	0.813 &	0.855 &	0.861 &	0.861 &	0.864 \\ \hline
\end{tabular}
}
\label{tab:abl_k}
\end{table*}

\paragraph{Number of models to ensemble}
We vary the number of models used in the ensemble, i.e., $B$, from one to twenty and compare the performances on several tabular datasets, whose results are presented in Table \ref{tab:abl_nens}.
There is a general tendency that using more models helps improve the identifying performance. 
The optimal number of models varies according to datasets, but the performance is not sensitive to the number of models used in the ensemble unless it is too small.

\begin{table*}[h!]
\renewcommand\thetable{C.41}
\caption{AUC results of the ODIM with various numbers of generative models to take an ensemble.}
\centering
\resizebox{0.65\textwidth}{!}{
\begin{tabular}{l|ccccccc}
\hline
\textbf{Data}            & $B=1$    & $B=2$    & $B=5$    & $B=10$   & $B=12$   & $B=15$   & $B=20$    \\
\hline
\texttt{Glass}       & 0.707 & 0.748 & 0.750 & 0.712 & 0.706 & 0.700 & 0.707  \\
\texttt{Mammography} & 0.825 & 0.836 & 0.844 & 0.856 & 0.857 & 0.858 & 0.859  \\
\texttt{Pendigits}   & 0.885 & 0.900 & 0.945 & 0.955 & 0.954 & 0.960 & 0.962  \\
\texttt{Pima}        & 0.686 & 0.690 & 0.699 & 0.703 & 0.704 & 0.705 & 0.705  \\
\texttt{Vowels}      & 0.877 & 0.882 & 0.903 & 0.904 & 0.896 & 0.895 & 0.899  \\
\texttt{Wbc}         & 0.922 & 0.922 & 0.931 & 0.936 & 0.935 & 0.938 & 0.937  \\
\texttt{Cardio}      & 0.898 & 0.896 & 0.915 & 0.918 & 0.911 & 0.924 & 0.920  \\
\texttt{Thyroid}     & 0.876 & 0.913 & 0.924 & 0.927 & 0.927 & 0.926 & 0.926  \\ \hline
\textbf{Average} & 0.835 &	0.848 &	0.864 &	0.864 &	0.861 &	0.863 &	0.864  \\ \hline
\end{tabular}
}
\label{tab:abl_nens}
\end{table*}

\clearpage

\paragraph{Number of patience}
We also examine the behavior of the ODIM with various values of the number of patience, $N_\text{pat}$, whose results are provided in Table \ref{tab:abl_npat}. 
Similar to the ensemble scenario, increasing the value of this hyper-parameter generally leads to enhanced performance. 
However, it appears that the extent of improvement levels off when $N_{\text{pat}}\ge 10$. 
A largeer value of $N_{\text{pat}}$ requires more computing time, therefore, we set $N_{\text{pat}}=10$ throughout all experiments in our manuscript.   

\begin{table*}[h!]
\renewcommand\thetable{C.42}
    \centering
    \caption{AUC results of the ODIM with various values of $N_\text{pat}.$}
    \resizebox{0.6\textwidth}{!}{
    \begin{tabular}{l|c|c|c|c|c}
        \hline
        \textbf{Data} & $\mathbf{N_{pat}=1}$ & $\mathbf{N_{pat}=2}$ & $\mathbf{N_{pat}=5}$ & $\mathbf{N_{pat}=10}$ & $\mathbf{N_{pat}=15}$ \\ \hline
        \texttt{Glass} & 0.526 & 0.652 & 0.706 & 0.726 & 0.744 \\ 
        \texttt{Mammography} & 0.587 & 0.815 & 0.853 & 0.851 & 0.846 \\ 
        \texttt{Pendigits} & 0.914 & 0.940 & 0.954 & 0.946 & 0.934 \\ 
        \texttt{Pima} & 0.482 & 0.617 & 0.707 & 0.705 & 0.706 \\ 
        \texttt{Vowels} & 0.502 & 0.517 & 0.802 & 0.907 & 0.921 \\ 
        \texttt{Wbc} & 0.528 & 0.740 & 0.869 & 0.941 & 0.941 \\ 
        \texttt{Cardio} & 0.819 & 0.925 & 0.924 & 0.882 & 0.879 \\ 
        \texttt{Thyroid} & 0.726 & 0.863 & 0.921 & 0.925 & 0.930 \\ \hline
        \textbf{Average} & 0.636 &	0.759 	&0.842 &	0.860 &	0.863 \\ \hline
    \end{tabular}
    }
    \label{tab:abl_npat}
\end{table*}


\paragraph{Learning schedule}
We evaluate the robustness of the ODIM to the learning schedule.  
We consider the Adam optimizer with various learning rates from 1e-4 to 1e-1, whose results on \texttt{FMNIST} are summarized in Table \ref{tab:abl_lr}.
We present the results of the 10 classes separately, where each class is regarded as the inlier class. 
Note that the identifying performances rarely change until we use a learning rate larger than 1e-2. 
As we usually do not consider a learning rate much larger than 1e-3 when we apply the Adam optimizer, we can conclude that our method is stable with respect to the learning schedule, which
implies that our method can be used in practice without delicate tunnings.

\begin{table*}[h]
\renewcommand\thetable{C.43}
    \centering
    \caption{(Train) AUC results of the ODIM with various values of learning rates on \texttt{FMNIST}.}
    \resizebox{0.72\textwidth}{!}{
    \begin{tabular}{l|c|c|c|c|c|c|c|c|c|c}
        \hline
        & \multicolumn{10}{c}{\textbf{Learning rate}} \\ \hline
        \textbf{Class} & \textbf{1e-4} & \textbf{2.5e-4} & \textbf{5e-4} & \textbf{1e-3} & \textbf{2.5e-3} & \textbf{5e-3} & \textbf{1e-2} & \textbf{2.5e-2} & \textbf{5e-2} & \textbf{1e-1} \\ \hline
        \texttt{0} & 0.902 & 0.915 & 0.913 & 0.911 & 0.907 & 0.902 & 0.881 & 0.561 & 0.564 & 0.481 \\ 
        \texttt{1} & 0.977 & 0.978 & 0.978 & 0.978 & 0.977 & 0.977 & 0.979 & 0.977 & 0.979 & 0.712 \\ 
        \texttt{2} & 0.814 & 0.857 & 0.861 & 0.864 & 0.869 & 0.860 & 0.845 & 0.500 & 0.415 & 0.392 \\ 
        \texttt{3} & 0.950 & 0.951 & 0.950 & 0.950 & 0.948 & 0.947 & 0.944 & 0.877 & 0.680 & 0.608 \\ 
        \texttt{4} & 0.917 & 0.912 & 0.913 & 0.904 & 0.902 & 0.913 & 0.791 & 0.447 & 0.453 & 0.314 \\ 
        \texttt{5} & 0.900 & 0.902 & 0.903 & 0.903 & 0.906 & 0.904 & 0.916 & 0.921 & 0.887 & 0.882 \\ 
        \texttt{6} & 0.808 & 0.809 & 0.809 & 0.807 & 0.803 & 0.796 & 0.735 & 0.566 & 0.522 & 0.494 \\ 
        \texttt{7} & 0.977 & 0.976 & 0.977 & 0.977 & 0.976 & 0.975 & 0.972 & 0.973 & 0.967 & 0.959 \\ 
        \texttt{8} & 0.863 & 0.845 & 0.843 & 0.841 & 0.821 & 0.807 & 0.661 & 0.409 & 0.397 & 0.423 \\ 
        \texttt{9} & 0.960 & 0.962 & 0.960 & 0.959 & 0.956 & 0.960 & 0.935 & 0.753 & 0.741 & 0.732 \\ \hline
        \textbf{Average} & 0.907 & 0.911 & 0.911 & 0.909 & 0.907 & 0.904 & 0.866 & 0.698 & 0.661 & 0.660 \\ \hline
    \end{tabular}
    }
    \label{tab:abl_lr}
\end{table*}

\clearpage


\paragraph{Min-max scaling v.s. Standardization}
We investigate which pre-processing technique is more suitable for our method. 
The two pre-processing techniques are considered, min-max scaling and standardization. 
We compare their corresponding results for ODIMs on 30 tabular dataset, which are provided in Table \ref{tab:auc_result_data_preprocessing}. 
We can see that using the min-max scaling usually gives better results than standardization. 
This supports our theoretical claim in Proposition 2. 


\begin{table}[h]
\renewcommand\thetable{C.44}
\caption{Comparison of the two data pre-processing methods on 30 tabular datasets: 1) min-max and 2) standardization. We report the AUCs.}
    \centering
    \resizebox{0.38\textwidth}{!}{
    \begin{tabular}{l|c|c}
    \hline
        \textbf{Data} & \textbf{Min-max} & \textbf{Standardization} \\ \hline
        \texttt{Annthyroid} & 0.591 
  & 0.700  \\ 
        \texttt{Breastw} & 0.992  & 0.772  \\ 
        \texttt{Cover} & 0.899  & 0.957  \\ 
        \texttt{Glass} & 0.758  & 0.687  \\ 
        \texttt{Ionosphere} & 0.926  & 0.862  \\ 
        \texttt{Letter} & 0.706  & 0.619  \\ 
        \texttt{Mammography} & 0.850  & 0.765  \\ 
        \texttt{Musk} & 1.000  & 0.963  \\ 
        \texttt{Pendigits} & 0.950  & 0.941  \\ 
        \texttt{Pima} & 0.718  & 0.452  \\ 
        \texttt{Speech} & 0.477  & 0.450  \\ 
        \texttt{Vertebral} & 0.385  & 0.547  \\ 
        \texttt{Vowels} & 0.896  & 0.596  \\ 
        \texttt{Wbc} & 0.937  & 0.840  \\ 
        \texttt{Arrhythmia} & 0.782  & 0.768  \\ 
        \texttt{Cardio} & 0.862  & 0.948  \\ 
        \texttt{Satellite} & 0.695  & 0.739  \\ 
        \texttt{Satimage-2} & 0.999  & 0.971  \\ 
        \texttt{Shuttle} & 0.986  & 0.994  \\ 
        \texttt{Thyroid} & 0.935  & 0.966  \\ 
        \texttt{ALOI} & 0.528  & 0.545  \\ 
        \texttt{Backdoor} & 0.877  & 0.895  \\ 
        \texttt{Campaign} & 0.732  & 0.753  \\ 
        \texttt{Celeba} & 0.832  & 0.769  \\ 
        \texttt{Fraud} & 0.921  & 0.956  \\ 
        \texttt{Landsat} & 0.461  & 0.540  \\ 
        \texttt{Magic.gamma} & 0.774  & 0.638  \\ 
        \texttt{PageBlocks} & 0.879  & 0.919  \\ 
        \texttt{Skin} & 0.648  & 0.597  \\ 
        \texttt{Waveform} & 0.692  & 0.641 \\ \hline
        \textbf{Average} & 0.790 & 0.760 \\ \hline
    \end{tabular}
    }
    \label{tab:auc_result_data_preprocessing}
\end{table}


\clearpage
\section*{D. Further discussions}
\subsection*{D.1. ODIM with partially labeled outliers}

Some studies have utilized the availability of outlier labels to enhance the efficiency of outlier detection tasks \citep{deepsad,daniel2019deep}.
But as far as we know, these existing works require that all outliers should be labeled, which is equivalent to the SOD setting. 
The only difference of these studies compared to the conventional SOD solvers is that they cast the problem into an one-class classification problem rather than a two-class one.

While it is very costly to obtain perfectly labeled data, partially labeled data are frequently met in practice.
In this section, we claim that the ODIM can be modified easily for such a situation by leveraging these labeled data. 
Assume that besides $\cU^{tr}$, a few labeled outlier dataset  $\cL^{tr}=\{(\bx_1^l,1),\ldots,(\bx_m^l,1)\}$ is also available. 
Again note that outliers are still present in $\cU^{tr}$

We simply adopt the idea of \citet{daniel2019deep}, which encourages the log-likelihood of known outliers to decrease with the variational \textit{upper} bound. 
For $u>1$, the upper bound, called $\chi$ upper bound (CUBO), is given as:
\bean
\label{eq:cubo}
    \begin{aligned}
L^{\text{CUBO}}&(\theta,\phi;\bx)\\
:=&\frac{1}{u}\log\mathbb{E}_{\bz\sim q(\bz|\bx;\phi)}\left[ \left(\frac{p(\bx|\bz;\theta)p(\bz)}{q(\bz|\bx;\phi)}\right)^u \right].
\end{aligned}
\eean
With the above CUBO term, we modify the loss function of the ODIM by adding the expected CUBO on $\cL^{tr}$ from the original IWAE loss function to have:
\bean
-\mathbb{E}_{\bx\sim\cU^{tr}}{L}^{\text{IWAE}}(\theta,\phi;\bx)+\gamma \cdot
\mathbb{E}_{\bx\sim\cL^{tr}}{L}^{\text{CUBO}}(\theta,\phi;\bx),
\eean
where $\gamma>0$ is a tuning parameter controlling the degree of the CUBO loss. 
The CUBO loss generally increases the IWAE per-sample loss values of outliers,
encouraging the IM effect to occur more clearly. 
In our paper, we set $u=2$ and $\gamma=1$. 

Table \ref{app_tab:auc_result_label_re} summarizes the averaged results of training AUC and PR scores across six tabular datasets for different proportions of labeled outliers, representing the ratio of labeled outliers to the total number of outliers. 
It is clearly seen that using label information helps to enhance identifying performance by a large margin. 

Note that the proposed modification can be improved further. 
For instance, we could use the labeled information to select the optimal number of updates. 
There would be other rooms to improve the ODIM with partially labeled data, which we leave as future research directions.

\begin{table}[h]
\renewcommand\thetable{D.1}
\caption{Training AUC (and PR) scores with various values of $l$. 
We consider three values for $l$, $l=0.0$, $0.3$, $0.5$.
}
\centering
\resizebox{0.5\textwidth}{!}{
    \begin{tabular}{l|ccc}
    \hline
        $l$ & $0.0$ & $0.3$ & $0.5$ \\ \hline
        \texttt{Arrhythmia} & 0.800 (0.443) & 0.837 (0.767) & 0.888 (0.772) \\ 
        \texttt{Cardio} & 0.916 (0.564) & 0.991 (0.934) & 0.993 (0.943) \\ 
        \texttt{Satellite} & 0.690 (0.652) & 0.868 (0.849) & 0.881 (0.849) \\ 
        \texttt{Satimage-2} & 0.997 (0.949) & 0.998 (0.954) & 0.999 (0.958) \\ 
        \texttt{Shuttle} & 0.981 (0.947) & 0.990 (0.977) & 0.990 (0.979) \\ 
        \texttt{Thyroid} & 0.928 (0.327) & 0.995 (0.844) & 0.995 (0.845) \\ \hline
        \textbf{Average} &0.885 (0.647) & 0.947 (0.871) & 0.958 (0.891)  \\ \hline
\end{tabular}
}
\label{app_tab:auc_result_label_re}
\end{table}

\clearpage

\subsection*{D.2. Differentially private ODIM}
\paragraph{DP-SGD}
DP-SGD is a variant of SGD applied when updating parameters while imposing DP guarantee to the model. 
For a per-sample loss for a given sample $\bx$, i.e., $l(\theta,\phi;\bx)$, we calculate gradient vector $\nabla_{\theta,\phi}l(\theta,\phi;\bx)$. 
We conduct a clipping operation with a given positive number $C>0$, then add a noise from the Gaussian distribution $\mathcal{N}(0,\sigma^2C^2 I)$ to have a deformed gradient $\bar{\nabla}_{\theta,\phi}l(\theta,\phi;\bx)={\nabla}_{\theta,\phi}l(\theta,\phi;\bx)/\max\left(1,\frac{\|{\nabla}_{\theta,\phi}l(\theta,\phi;\bx)\|_2}{C}\right)+\mathcal{N}(0,\sigma^2C^2 I)$, where $\sigma>0$ is also a pre-specified number. 
Then we update parameters $(\theta,\phi)$ using this update information $\bar{\nabla}_{\theta,\phi}l(\theta,\phi;\bx)$ using a conventional SGD or variants such as Adam and RMSProp. 

Here, the two hyper-parameters, $C$ and $\sigma$ need to be specified before implementing DP-SGD. 
In our experiment, we set $(C,\sigma)=(20,1.02)$. 
The algorithm of DP-SGD is summarized in the following. 

\begin{algorithm}[t]
\end{algorithm}

\begin{algorithm}[ht]
 \caption{ \textbf{Differentially private SGD} (We set $(C,\sigma)=(20,1.02)$.) }
    \begin{algorithmic}[1]
        \REQUIRE: Training dataset $\mathcal{U}^{tr}=\{\bx_1,...,\bx_n\}$, parameters: $(\theta,\phi)$, loss function of a given sample: $l(\theta,\phi;\bx)$, optimizer: $Opt(\theta,\phi,\nabla)$, group size: $L$, number of updates: $T$
        

        Initialize $(\theta_0,\phi_0)$
        \FOR{t in $1:T$}
            \STATE Drawn a random sample indices $A$ from $[n]$ with sampling probability of $L/n$
            \STATE For each $i\in A$, compute the gradient \\$g_{t-1}(\bx_i)\leftarrow\nabla_{\theta,\phi}l(\theta_{t-1},\phi_{t-1};\bx)$
            \STATE Clip the gradient and add Gaussian noise\\
            $\bar{g}_{t-1}(\bx_i)\leftarrow{g}_{t-1}(\bx_i)/\max\left(1,\frac{\|{g}_{t-1}(\bx_i)\|_2}{C}\right)+\mathcal{N}(0,\sigma^2C^2 I)$
            \STATE Aggregate the gradients\\
            $\bar{g}_{t-1}\leftarrow \frac{1}{L}\sum_{i\in A} \bar{g}_{t-1}(\bx_i)$
            \STATE Update parameters\\
            $(\theta_t,\phi_t)\leftarrow Opt(\theta_{t-1},\phi_{t-1},\bar{g}_{t-1})$
        \ENDFOR
        \\
    \end{algorithmic}
\textbf{Output}: $(\theta_T,\phi_T)$ 
\end{algorithm}

\paragraph{$(\epsilon,\delta)$-DP} 
$(\epsilon,\delta$-DP is one of the widely used measures to examine the amount of privacy protection for a given random mechanism. 
The definition of $(\epsilon,\delta)$-DP is as follows:

\paragraph{Definition D.1.} A randomized mechanism $\mathcal{M}$ with range $\mathcal{R}$ is $(\epsilon,\delta)$-DP, if 
\bean
\Pr[\mathcal{M}(\mathcal{S})\in\mathcal{O}]\le \exp(\epsilon)\cdot \Pr[\mathcal{M}(\mathcal{S'})\in\mathcal{O}]+\delta
\eean
holds for any subset of outputs $\mathcal{O}\subseteq \mathcal{R}$ and for any neighboring datasets $\mathcal{S}$ and $\mathcal{S'}$, i.e., $|\mathcal{S}-\mathcal{S'}|=1$.

It can be easily inferred that a random mechanism with small values of $\epsilon$ and $\delta$ is believed to provide strict privacy protection. 

\paragraph{Calculation of $(\epsilon,\delta)$-DP when applying DP-SGD}
It is well known that a single iteration of DP-SGD satisfies $(\epsilon,\delta)$-DP for certain values of $\epsilon$ and $\delta$. 
However, to train a given model, or parameters, we have to iterate DP-SGD multiple times. 
At this point, the key lies in calculating $\epsilon$ and $\delta$ properly of \textit{composition} of DP-SGD operations. 

There have been numerous techniques to calculate tight $\epsilon$ when random mechanisms are sequentially composed \citep{abadi2016deep,mironov2017renyi,dong2019gaussian}.  
Among them, we adopt the method of \citet{mironov2017renyi} as other recent methods followed this approach \citep{chen2020gs,zhao2023ctab}. 

\clearpage

\paragraph{Detailed experimental results}

We compare our method to DeepSVDD when DP-SGD is applied. 
As mentioned earlier, we set $\delta=10^{-5}$ and iterate DP-SGD to train each model and stopping the learning process when the privacy budget first exceed 10, i.e., $\epsilon=10$. 
We note that applying DP-SGD to non-deep-learning-based methods is not possible, as they are not trained using SGD-based optimizers. 

We analyze four tabular dataset, whose results are presented in Table \ref{app_tab:dp_results}. 

\begin{table}[h!]
\renewcommand\thetable{D.2}
\caption{Train AUC (and PR) value comparisons when DP-SGD is applied. We analyze four tabular datasets and set $(\epsilon,\delta)=(10,10^{-5})$.}
    \centering
    \resizebox{0.38\textwidth}{!}{
    \begin{tabular}{l|c|c}
    \hline
        \textbf{Class} & \textbf{DeepSVDD} & \textbf{ODIM} \\ \hline
        \texttt{Arrhythmia} & 0.585 (0.325) & 0.626 (0.362)  \\ 
        \texttt{Cardio} & 0.543 (0.132)& 0.743 (0.321) \\ 
        \texttt{Thyroid} & 0.745 (0.109)  & 0.805 (0.155)  \\ 
        \texttt{Vowels} & 0.582 (0.040) & 0.666 (0.096)  \\ \hline
        \textbf{Average} & 0.614 (0.152) & 0.738 (0.234) \\ \hline
    \end{tabular}
    }
\label{app_tab:dp_results}
\end{table}

\clearpage

\subsection*{D.3. Likelihood based methods have poor performances in UOD tasks}

In the experimental section of the main manuscript, we have mentioned that the existing likelihood-based SSOD solvers often struggle to distinguish inliers with outliers in UOD tasks. 
In this section, we evaluate the performance of a likelihood-based SSOD method \citep{DBLP:conf/iclr/NalisnickMTGL19} and compare its effectiveness to our method on \texttt{MNIST} and \texttt{FMNIST} datasets. 
It is worth noting that we attempted to implement other methods, such as \citet{DBLP:journals/corr/abs-1906-02994,DBLP:journals/entropy/LanD21}, but one of them does not have an official GitHub code, and the other is based on an old version of Tensorflow, which is also not available.
Therefore, we have excluded them in our experiment. 

Tables \ref{app_tab:auc_ap_result_likelihood_mnist} and \ref{app_tab:auc_ap_result_likelihood_fmnist} provide a summary of the comparison results. 
It is evident that our method outperforms the competitor across all considered scenarios. 
Notably, the AUC scores of \citet{DBLP:journals/corr/abs-1906-02994,DBLP:journals/entropy/LanD21} often fall below 0.5, indicating a failure to identifying inliers from a given dataset. 
Consequently, we can conclude that likelihood-based SSOD methods are sub-optimal as UOD solvers. 

\begin{table}[h!]
\renewcommand\thetable{D.3}
\caption{Train AUC (and PR) value comparisons on \texttt{MNIST} (image data)}
    \centering
    \resizebox{0.5\textwidth}{!}{
    \begin{tabular}{l|c|c}
    \hline
        \textbf{Class} & \textbf{\citet{DBLP:conf/iclr/NalisnickMTGL19}+GLOW} & \textbf{ODIM} \\ \hline
        \texttt{0} & 0.645 (0.951) & 0.937 (0.990)  \\ 
        \texttt{1} & 0.710 (0.723)& 0.997 (0.999) \\ 
        \texttt{2} & 0.555 (0.928)  & 0.732 (0.957)  \\ 
        \texttt{3} & 0.502 (0.916) & 0.800 (0.969)  \\ 
        \texttt{4} & 0.322 (0.882)  & 0.854 (0.978)  \\ 
        \texttt{5} & 0.484 (0.924)  & 0.729 (0.964)  \\ 
        \texttt{6} & 0.445 (0.910) & 0.859 (0.976)  \\ 
        \texttt{7} & 0.204 (0.840) & 0.928 (0.990)   \\ 
        \texttt{8} & 0.671 (0.954)  & 0.709 (0.953)  \\ 
        \texttt{9} & 0.333 (0.881) & 0.889 (0.983)  \\ \hline
        \textbf{Average} & 0.387 (0.869) & 0.843 (0.976) \\ \hline
    \end{tabular}
    }
\label{app_tab:auc_ap_result_likelihood_mnist}
\end{table}

\begin{table}[h!]
\renewcommand\thetable{D.4}
\caption{Train AUC(and PR) value comparisons on \texttt{FMNIST} (image data)}
    \centering
    \resizebox{0.5\textwidth}{!}{
    \begin{tabular}{l|c|c}
    \hline
        \textbf{Class} & \textbf{\citet{DBLP:conf/iclr/NalisnickMTGL19}+GLOW} & \textbf{ODIM} \\ \hline
        \texttt{0} & 0.460 (0.904) & 0.905 (0.986)  \\ 
        \texttt{1} & 0.059 (0.756)& 0.976 (0.997)   \\ 
        \texttt{2} & 0.589 (0.938)  & 0.858 (0.981)   \\ 
        \texttt{3} & 0.265 (0.821) &  0.943 (0.993)  \\ 
        \texttt{4} & 0.593 (0.938)  & 0.890 (0.985) \\ 
        \texttt{5} & 0.222 (0.807)  & 0.899 (0.988)  \\ 
        \texttt{6} & 0.645 (0.944) & 0.802 (0.971)  \\ 
        \texttt{7} & 0.101 (0.781) & 0.980 (0.998) \\ 
        \texttt{8} & 0.633 (0.941)  & 0.839 (0.973)  \\ 
        \texttt{9} & 0.361 (0.887) & 0.958 (0.995) \\ \hline
        \textbf{Average} & 0.398 (0.886) & 0.905 (0.987) \\ \hline
    \end{tabular}
    }
\label{app_tab:auc_ap_result_likelihood_fmnist}
\end{table}





\end{document}